\documentclass{article}

\PassOptionsToPackage{numbers,sort,compress}{natbib}

\usepackage[main, final]{neurips_2026}

\usepackage[utf8]{inputenc} %
\usepackage[T1]{fontenc}    %
\usepackage[hidelinks]{hyperref}       %
\usepackage{url}            %
\usepackage{booktabs}       %
\usepackage{amsfonts}       %
\usepackage{nicefrac}       %
\usepackage{microtype}      %
\usepackage{xcolor}         %

\usepackage{amsmath}
\usepackage{amssymb}
\usepackage{mathtools}
\usepackage{amsthm}
\usepackage{bbm}
\usepackage[capitalize,noabbrev]{cleveref}
\usepackage{algorithm}
\usepackage[noend]{algpseudocode}
\usepackage{threeparttable}
\usepackage{subcaption}
\usepackage{mathrsfs}
\usepackage{wrapfig}
\usepackage{caption}
\usepackage{tabularx}
\newlength{\authorboxwidth}
\usepackage{makecell}
\usepackage{tikz}
\usetikzlibrary{arrows.meta,positioning,fit,calc}

\usepackage[most]{tcolorbox}
\newlength{\capshift}
\newcommand{\shiftedcaption}[1]{\par\nointerlineskip\noindent\hspace*{\capshift}\makebox[0pt][l]{\begin{minipage}[t]{\linewidth}\caption{#1}\end{minipage}}}
\algrenewcommand\algorithmicrequire{\textbf{Require:}}
\algrenewcommand\algorithmicensure{\textbf{Return:}}
\algrenewcommand\textproc{\textsf}
\algrenewcommand\alglinenumber[1]{\footnotesize #1:}
\makeatletter
\renewcommand{\ALG@beginalgorithmic}{\small}
\tcbset{
  highlightbox/.style={
    colback=black!7, colframe=black!7, boxrule=0pt,
    arc=0pt, left=2pt, right=2pt, top=3pt, bottom=3pt,
    enlarge left by=0mm, enlarge right by=0mm
  }
}

\title{A Trust Region Approach\\for Learning Schrödinger Bridges}

\author{%
  \makebox[\authorboxwidth]{Takeshi Koshizuka\thanks{Equal contribution.}} \\
  RIKEN AIP \\
  Kyushu University \\
  \texttt{takeshi.koshizuka@riken.jp} \\
  \And
  \makebox[\authorboxwidth]{Denis Blessing} \\
  Karlsruhe Institute of Technology \\
  \texttt{denis.blessing@kit.edu} \\
  \AND
  \makebox[\authorboxwidth]{Max Zimmer, Sebastian Pokutta} \\
  Zuse Institute Berlin \\
  Technische Universität Berlin \\
  \texttt{\{zimmer, pokutta\}@zib.de} \\
  \And
  \makebox[\authorboxwidth]{Lorenz Richter\footnotemark[1]} \\
  dida \\
  Zuse Institute Berlin \\
  \texttt{richter@zib.de} \\
}

\usepackage{aliascnt}
\newcommand{\sharedtheorem}[2]{\newaliascnt{#1}{theorem}\newtheorem{#1}[#1]{#2}\aliascntresetthe{#1}\crefname{#1}{#2}{#2s}}
\theoremstyle{plain}

\sharedtheorem{proposition}{Proposition}
\sharedtheorem{lemma}{Lemma}
\sharedtheorem{corollary}{Corollary}
\theoremstyle{definition}
\sharedtheorem{definition}{Definition}
\sharedtheorem{assumption}{Assumption}
\theoremstyle{remark}
\sharedtheorem{remark}{Remark}

\renewcommand{\P}{\mathbbm{P}}
\newcommand{\KL}{D_{\mathrm{KL}}}
\newcommand{\E}{\mathbb{E}}
\newcommand{\R}{\mathbb{R}}
\newcommand{\U}{\mathcal{U}}
\newcommand{\pderiv}[2]{\frac{\partial #1}{\partial #2}}

\providecommand{\widebar}[1]{\mkern 1.5mu\overline{\mkern-1.5mu#1\mkern-1.5mu}\mkern 1.5mu}

\DeclareMathOperator*{\argmax}{arg\,max}
\DeclareMathOperator*{\argmin}{arg\,min}

\makeatletter
\DeclareRobustCommand{\cev}[1]{%
  \mathpalette\do@cev{#1}%
}
\newcommand{\do@cev}[2]{%
  \fix@cev{#1}{+}%
  \reflectbox{$\m@th#1\vec{\reflectbox{$\fix@cev{#1}{-}\m@th#1#2\fix@cev{#1}{+}$}}$}%
  \fix@cev{#1}{-}%
}
\newcommand{\fix@cev}[2]{%
  \ifx#1\displaystyle
    \mkern#23mu
  \else
    \ifx#1\textstyle
      \mkern#23mu
    \else
      \ifx#1\scriptstyle
        \mkern#22mu
      \else
        \mkern#22mu
      \fi
    \fi
  \fi
}

\makeatother

\begin{document}

\maketitle
\vspace{-14pt}
\addtocontents{toc}{\protect\setcounter{tocdepth}{-1}}

\begin{abstract}
  We study the numerical approximation of Schrödinger Bridges in the setting where samples from the target distribution are unavailable and only an unnormalized target density is given. In this regime, the widely used iterative proportional fitting (IPF) procedure (and its variants) often exhibits severe numerical instabilities, stemming from ill-conditioned underlying control problems and from the fact that the optimal solution may lie far from the initialization. To address these issues, we propose a trust region variant of IPF that constrains successive updates to remain at a fixed Kullback-Leibler divergence from the current iterate. This restriction significantly improves numerical stability. Moreover, we show that the resulting method admits an interpretation as a mirror descent scheme with an adaptively chosen step size. We demonstrate the effectiveness of our approach on several challenging sampling tasks.
\end{abstract}

\section{Introduction}

We study the problem of sampling from a target distribution 
\begin{equation}
    p_\mathrm{target} = \frac{\exp\left(-\mathcal{E}\right)}{\mathcal Z},
\end{equation}
where $\mathcal{E} \in C(\R^d, \R)$ is an energy function that can be evaluated pointwise and $\mathcal Z = \int_{\R^d} \exp\left(-\mathcal{E}(x)\right) \mathrm dx$ is the normalizing constant. Our approach relies on transporting samples from a simple distribution to the target via a stochastic process. In particular, we study Schrödinger bridges, i.e., stochastic measure transport with minimal control costs, mathematically equivalent to entropy-regularized optimal transport \cite{leonard2013survey}. While Schrödinger bridges have been extensively studied in data-driven settings, where samples from the target distribution are available, their application to pure sampling tasks remains less explored. In the sampling regime, where only the energy $\mathcal{E}$ is available, the problem becomes significantly more challenging, e.g., due to unknown locations of the modes of the distribution.

\begin{figure}[t]
    \refstepcounter{figure}\label{fig: illustration}%
    \sbox0{\includegraphics[width=0.6\linewidth]{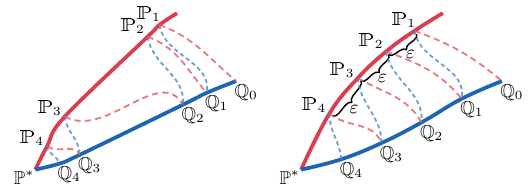}}%
    \noindent\hangindent=\dimexpr\wd0+\columnsep\relax\hangafter=-9%
    \llap{\raisebox{\dimexpr\ht\strutbox-\ht0\relax}[0pt][0pt]{\usebox0}\hspace{\columnsep}}%
    \textbf{Figure \thefigure:} The classical iterative proportional fitting (IPF) algorithm for solving Schrödinger bridges proceeds by alternating projections, producing sequences of path measures $(\P_i)_{i\geq 0}$ and $(\mathbbm{Q}_i)_{i\geq 0}$ that are guaranteed to converge to the optimal measure $\mathbbm{P}^*$. However, as illustrated on the left, successive measures $\P_i$ may be separated by arbitrarily large distances, which can complicate optimization and lead to numerical instabilities. To address this issue, we propose the \emph{Trust Region Schrödinger Bridge Sampler}, which enforces equidistant updates in terms of the Kullback-Leibler divergence.\par
\end{figure}

Existing approaches for solving Schrödinger bridges often rely on iterative schemes, such as iterative proportional fitting (IPF) or Markovian projections \cite{fortet1940resolution,shi2023diffusion}. However, applying these to sampling introduces difficult optimization landscapes characterized by a delicate exploration-exploitation tradeoff and a notorious tendency toward mode-collapse, often implying severe numerical instabilities. This stems from two primary sources: the ill-conditioning of the underlying control problems and the substantial distance often found between the initialization and the optimal solution. 
The classical IPF algorithm can be viewed as an alternating KL projection between boundary constraints. 
Given an intermediate path measure $\mathbb{Q}_i$, %
the next iterate is obtained by projecting onto a constraint $\mathcal{C}$: %
\begin{equation}
    \label{eq:kl_proj}
    \mathbb{P}_{i+1}=\argmin_{\mathbb{P}\in\mathcal{C}} \KL(\mathbb{P}|\mathbb{Q}_i).
\end{equation}

It can be shown that this corresponds to a mirror descent scheme with a fixed step size of $1$ \citep{karimi2023sinkhorn}. By relaxing this parameter, one can consider a generalized update rule with step size $\gamma$, akin to the discretization of continuous Sinkhorn flows,
\begin{equation}
\P_{i+1} \approx \P_i^{1-\gamma} (\operatorname{Proj}_{\mathcal{C}}\mathbb{Q}_i)^\gamma,
\end{equation}
where $\operatorname{Proj}_{\mathcal{C}}$ denotes the KL projection in \eqref{eq:kl_proj}. While this generalized view suggests that the step size $\gamma$ could technically be varied to stabilize training, practical guidelines for selecting this parameter are non-existent.

To address these limitations, we extend the recently introduced Adjoint Schrödinger Bridge Sampler (ASBS) \cite{liu2025adjoint} along two complementary axes. First, we generalize its unit-step forward matching update to arbitrary mirror-descent step sizes and select the step size adaptively through a path space KL trust region. Second, we incorporate the importance weight that the forward KL objective requires but ASBS omits. Combining these extensions yields the \emph{Trust Region Schrödinger Bridge Sampler} (TR-SBS), which mitigates mode collapse while retaining the simulation-efficient matching structure of ASBS. Our main contributions are:
\begin{enumerate}
\setlength{\itemsep}{0pt}\setlength{\parskip}{0pt}
    \item We propose TR-SBS, a $\gamma$-IPF scheme whose step size is selected by a path space KL trust region, adapted to the moving target of IPF (\Cref{prop: geometric interpretation of update}).
    \item We derive a generalized adjoint-matching objective that accommodates arbitrary step sizes and incorporates the importance weight that the cross-entropy objective requires (\Cref{eq: CE as matching objective}).
    \item We establish a convergence guarantee for the resulting trust region SB iteration (\Cref{thm:convergence_rate}) and empirically demonstrate that adaptive step size selection and importance weighting play complementary roles in mitigating mode collapse (\Cref{sec: experiments}).
\end{enumerate}

\subsection{Related work}

The Schrödinger bridge problem dates back to the seminal work of \cite{schrodinger1931umkehrung} and has since been studied extensively in probability theory and stochastic control; see, e.g., \cite{fortet1940resolution,dai1991stochastic,ruschendorf1995convergence,leonard2013survey}. More recently, the iterative proportional fitting (IPF) algorithm for numerical approximation has been revisited from an optimization perspective, and its equivalence to a mirror descent scheme has been analyzed in \cite{aubin2022mirror,karimi2023sinkhorn}.
In the past few years, Schrödinger bridges have emerged as a powerful framework for generative modeling, providing a unifying viewpoint that connects stochastic processes, optimal transport, and diffusion-based methods. Representative contributions include \cite{wang2021deep,vargas2021machine,vargas2021solving,de2021diffusion,chen2021likelihood,gottwald2024localized}, all of which focus on data-driven settings where samples from the target distribution are available. Subsequent research has extended these foundations in several directions. For instance, \cite{shi2023diffusion} proposed diffusion Schrödinger bridge matching, while \cite{peluchetti2023diffusion} investigated mixture-based transport constructions. A further unifying perspective was introduced by \cite{liu2023generalized}.

Schrödinger bridges have also been explored in data-free settings, where the goal is sampling from a target distribution; see, e.g., \cite{bernton2019schr,huang2021schrodinger}. In parallel, diffusion-based sampling methods employing alternative transport maps have led to substantial performance improvements, as demonstrated in \cite{zhang2022path,berner2022optimal,vargas2023denoising,chen2024sequential, berner2025discrete,sun2024dynamical,gruhlke2026tensor}. These developments were further unified and generalized in \cite{richter2023improved,vargas2024transport}, with extensions to the underdamped setting studied in \cite{blessing2025underdamped}. Related control-matching approaches based on fixed-point iterations have been investigated in \cite{domingoenrich2025adjoint,choi2025non,liu2025adjoint,havens2025adjoint,nam2025enhancing}.
Among these works, ASBS \cite{liu2025adjoint} is most closely related to ours. ASBS implements the unit-step IPF update through an unweighted adjoint-matching fixed-point objective. We generalize this construction to arbitrary step sizes, incorporate the importance weight that the forward KL objective requires, and select the step size adaptively through a KL trust region. ASBS has itself been generalized by the Bridge Matching Sampler~\cite{blessing2026bridge}, which casts such matching objectives as fixed-point iterations based on Nelson's relation and yields a single, stable objective.

Finally, we want to mention that trust region methods have a long history as robust optimization algorithms; see~\cite{conn2000trust} for an overview. These methods have been extended to probability distributions, particularly in reinforcement learning, see, e.g., \cite{peters2010relative,abdolmaleki2015model,schulman2015trust,thalmeier2020adaptive,otto2021differentiable}. A treatment of trust regions for stochastic optimal control and path space measures has recently been suggested in \cite{blessing2025trust}, where the target path measure is fixed. In contrast, the target of our forward update changes after every backward IPF step, which calls for a moving-target formulation and an SB-specific reformulation; we detail the differences in \Cref{app: relation to blessing}.

\section{Schrödinger bridges}

A Schrödinger bridge between $p_\mathrm{prior}$ and $p_\mathrm{target}$ is defined as a stochastic process
\begin{equation}
\label{eq: forward SDE}
    \mathrm d X^u_t = \sigma(t) u(X^u_t,t)\mathrm d t + \sigma(t) \vec{\mathrm d} W_t, \qquad X^u_0 \sim p_{\mathrm{prior}}
\end{equation}
with minimal control costs in the sense of
\begin{equation}
\label{eq: SB problem}
   \! \min_{u \in \mathcal{U}} \KL(\vec{\P}^u|\vec{\P}^\mathrm{ref})  \! = \!\min_{u \in \mathcal{U}}  \E\left[\frac{1}{2}\int_0^T \! \|u(X^u_t,t)\|^2 \mathrm d t\right]
\end{equation}
such that $X^u_T \sim p_{\mathrm{target}}$ (see \Cref{app: proofs} for remarks on the notation). This problem can be approached by additionally considering the reverse-time SDE
\begin{equation}
\label{eq: backward SDE}
    \mathrm d Y^v_t = -\sigma(t) v(Y^v_t,t)\mathrm d t + \sigma(t) \cev{\mathrm d} W_t, \qquad Y^v_T \sim p_{\mathrm{target}}.
\end{equation}
As for instance shown in \cite{leonard2013survey}, the solution is connected to the \textit{Schrödinger system}
\begin{align}
\label{eq: Schrödinger system boundary conditions}
    \varphi_t(x)=\int_{\R^d} p_{T | t}^{\mathrm{ref}}(y | x) \varphi_T(y) \mathrm{d} y, \qquad \widehat{\varphi}_t(x)=\int_{\R^d} p_{t | 0}^{\mathrm{ref}}(x | y) \widehat{\varphi}_0(y) \mathrm{d} y
\end{align}
with coupled boundary conditions $\varphi_0(x) \widehat{\varphi}_0(x)=p_\mathrm{prior}(x)$ and $\varphi_T(x) \widehat{\varphi}_T(x)=p_\mathrm{target}(x),$ where $p_{T | t}^{\mathrm{ref}}$ is the transition density of the reference process. The functions $\varphi$ and $\widehat{\varphi}$ are called \textit{Schrödinger potentials} and
the optimal controls solving the Schrödinger bridge problem (and its time-reversal) are then given by
\vspace{-0.2cm}
\begin{equation}
\label{eq: optimal controls}
    u^*=\sigma^\top \nabla \log \varphi, \qquad v^*=\sigma^\top \nabla \log \widehat{\varphi}.
\end{equation}
In this case, $Y^{v^*}\!$ is the time reversal of $X^{u^*}\!$, i.e., $\vec{\P}^{u^*} \!= \cev{\P}^{v^*}$.
\vspace{-0.2cm}
\subsection{Iterative proportional fitting}
\vspace{-0.1cm}
Solving for the optimal control functions \eqref{eq: optimal controls} is non-trivial, in particular due to the coupled boundary conditions \eqref{eq: Schrödinger system boundary conditions}. A classical approach is based on iteratively solving so-called \emph{Schrödinger half-bridges}, in which only one of the time marginals of the process is fixed. The resulting \emph{iterative proportional fitting} (IPF) procedure alternates updates of the forward and backward controls as
\begin{align}
\label{eq: IPF iterations}
\begin{split}
    u_{i+1} &= \argmin_{u \in \mathcal{U}} \Big\{ \KL(\vec{\P}^{u} | \cev{\P}^{v_i}) : \vec{\P}^u_0 = p_\mathrm{prior} \Big\}, \\
    v_{i+1} &= \argmin_{v \in \mathcal{U}} \Big\{ \KL(\cev{\P}^{v} | \vec{\P}^{u_{i+1}}) : \cev{\P}^v_T = p_\mathrm{target} \Big\}.
\end{split}
\end{align}
Since in our setting we do not have samples from $p_\mathrm{target}$, it is convenient to replace the reverse KL in the second equation by a forward KL, yielding
\vspace{-0.1cm}
\begin{align}
   \! v_{i+1} \! = \! \argmin_{v \in \mathcal{U}} \Big\{ \KL(\vec{\P}^{u_{i+1}} | \cev{\P}^{v}) : \cev{\P}^v_T = p_\mathrm{target} \Big\}.
\end{align}
As shown in Corollary E.2 of \cite{vargas2024transport}, replacing the reverse KL with the forward KL in the backward update does not alter the minimizing sequence $(u_i, v_i)$. Moreover, under regularity assumptions, the IPF iterates converge to the Schr\"odinger bridge in total variation as $i \to \infty$ \cite{de2021diffusion}.

\vspace{-0.1cm}
\subsection{IPF as mirror descent}
\label{sec:ipf_as_mirror_descent}
\vspace{-0.1cm}
\label{sec:mirror_descent}
In \cite{karimi2023sinkhorn} it is shown that the IPF iterates stated in \eqref{eq: IPF iterations} can be equivalently formulated by a mirror descent scheme on path space,
\vspace{-0.2cm}
\begin{equation}
u_{i+1} \!=\! \operatorname*{arg\,min}_{u \in \mathcal{U}} \left\{\! \langle \delta F(\vec{\P}^{u_i}), \vec{\P}^u \!-\! \vec{\P}^{u_i} \rangle \!+\! \frac{D_{\mathrm{KL}}(\vec{\P}^u | \vec{\P}^{u_i})}{\gamma_i}\! \right\}\!,
\end{equation}
when choosing $F(\vec{\P}^{u_i}) = \KL(\vec{\P}^{u_i}_T | p_\mathrm{target})$ and the step size $\gamma_i = 1$. Here, $\delta F(\vec{\P}^{u_i})$ denotes the first variation of the terminal loss functional $F$, evaluated at the current iterate $\vec{\P}^{u_i}$. Thus, each update balances a linearized descent step that drives the terminal marginal toward the target distribution $p_{\mathrm{target}}$ against a regularization term that penalizes deviations from the current path space dynamics $\vec{\P}^{u_i}$ via the  KL divergence. Further, as also shown in \cite{karimi2023sinkhorn}, we can integrate the step sizes $\gamma_i \in (0, 1]$ into the classical IPF formulation \eqref{eq: IPF iterations}, yielding
\begin{align}
\begin{split}
\label{eq: gamma IPF}
    u_{i+1} \!=\! \operatorname*{arg\,min}_{u \in \mathcal{U}}\! \left\{ \KL(\vec{\P}^{u} | \cev{\P}^{v_i}) \!+\! \frac{1{-}\gamma_i}{\gamma_i} \KL(\vec{\P}^{u} | \vec{\P}^{u_i}) \right\}
\end{split}
\end{align}
under the constraint $\vec{\P}^u_0  = p_\mathrm{prior}$ for the $u$-updates, while the $v$-updates remain the same. In practice, it is natural to ask which step size to choose in order to stabilize the procedure. We shall answer this question in the following section.

\section{Trust region based Schrödinger bridges}
\label{sec: trust region SB}
Inspired by \cite{blessing2025trust}, we propose to stabilize the $u$-updates in the IPF procedure \eqref{eq: IPF iterations} by incorporating a trust region constraint. The purpose of this constraint is to control the magnitude of each update and to ensure that successive control iterates remain within a prescribed neighborhood of the previous iterate. Concretely, we restrict deviations, measured in terms of the KL divergence, from the previous control $u_i$ by a radius $\varepsilon > 0$, and define
\begin{equation}
\label{eq: TR objective}
u_{i+1}
= \argmin_{u \in \mathcal{U}} \KL(\vec{\P}^{u} | \cev{\P}^{v_i}) \,\,\,\text{s.t.}\,\,\,
\KL(\vec{\P}^{u} | \vec{\P}^{u_i}) \le \varepsilon,
\end{equation}
again subject to the half-bridge constraint $\vec{\P}^{u}_0 = p_{\mathrm{prior}}$.
We call the iteration that alternates the backward IPF half-step with the forward update \eqref{eq: TR objective} \emph{trust region IPF} (TR-IPF) and state it in path space in \Cref{alg: TR-IPF} (\Cref{app: convergence}). TR-SBS, which we describe in \Cref{sec: learning constrained optimal control}, learns the steps of TR-IPF from samples.
Its convergence, speed, and stability are analyzed in \Cref{sec: convergence}.

In practice, under suitable regularity assumptions, the constrained optimization problem \eqref{eq: TR objective} can be addressed via a relaxed Lagrangian formulation. Introducing a Lagrange multiplier $\lambda \ge 0$, we define the augmented objective
\begin{equation}
\label{eq: Langrangian optimization}
\mathcal{L}^{\mathrm{TR}}_i(u, \lambda)
:= \KL(\vec{\P}^{u} | \cev{\P}^{v_i})
+ \lambda \bigl(\KL(\vec{\P}^{u} | \vec{\P}^{u_i}) - \varepsilon\bigr),
\end{equation}
which leads to the saddle-point problem
\begin{equation}
\label{eq: saddle-point problem}
\max_{\lambda \ge 0} \min_{u \in \mathcal{U}} \mathcal{L}^{\mathrm{TR}}_i(u, \lambda).
\end{equation}
We note that the KL divergence is convex in the path measures. Moreover, the value function $\lambda \mapsto \min_{u \in \mathcal{U}} \mathcal{L}^{\mathrm{TR}}_i(u,\lambda)$ is concave in $\lambda$, since it is given as the pointwise infimum of affine functions of $\lambda$; see, e.g., \cite{blessing2025trust}. Under standard regularity assumptions, the saddle-point problem \eqref{eq: saddle-point problem} thus admits a unique solution, which we denote by $(u_{i+1}, \lambda_i)$. For later analysis, we introduce the cost-to-go functional of the trust region control (TRC) problem, where it is crucial to restrict the evaluation to the future path segments. Let $\vec{\P}^{u}_{[t, T]}$ and $\cev{\P}^{v}_{[t, T]}$ denote the path measures restricted to the time interval $[t, T]$. We define\footnote{Note that the normalization constant $\mathcal{Z}$ from $p_\mathrm{target}$ in the Radon-Nikodym derivative $\mathrm{d}\vec{\P}^{u}_{[t, T]} / \mathrm{d}\cev{\P}^{v_i}_{[t, T]}$ is merely an additive constant independent of $u$ and $\gamma$ and can therefore be ignored in optimization.}
\begin{equation}
\label{eq: J_TRC_pointwise}
    \mathcal{L}^{\mathrm{TRC}}_i(u, \lambda; x, t) 
    \!:= \!\mathbb{E}_{\vec{\P}^{u}}\!\left[ \log\!\frac{\mathrm{d}\vec{\P}^{u}_{[t, T]}}{\mathrm{d}\cev{\P}^{v_i}_{[t, T]}} \!+\! \lambda \log\!\frac{\mathrm{d}\vec{\P}^{u}_{[t, T]}}{\mathrm{d}\vec{\P}^{u_i}_{[t, T]}} \Bigg| X_t \!= \! x \right]\!.
\end{equation}
 By definition, this functional accumulates the running and terminal costs only from time $t$ to $T$. Observe that taking the expectation at $t=0$ recovers the original Lagrangian up to a constant shift:
\begin{equation}
\label{eq: L_TRC_relation}
    \mathbb{E}_{X_0 \sim p_{\mathrm{prior}}} \left[\mathcal{L}^{\mathrm{TRC}}_i(u,\lambda; X_0, 0)\right] 
    = \mathcal{L}^{\mathrm{TR}}_i(u,\lambda) + \lambda \varepsilon.
\end{equation}
Since the term $\lambda \varepsilon$ is constant with respect to $u$, minimizing $\mathcal{L}^{\mathrm{TR}}_i(u, \lambda)$ is equivalent to minimizing the expected costs defined by \eqref{eq: L_TRC_relation}. 

Comparing $\mathcal{L}^{\mathrm{TRC}}_i$ with the generalized Sinkhorn update \eqref{eq: gamma IPF}, we observe that the Lagrange multiplier can be identified with the IPF step size via $\lambda = \frac{1-\gamma}{\gamma}$. Enforcing the constraint $\KL(\vec{\P}^{u} | \vec{\P}^{u_i}) \le \varepsilon$ induces an adaptive choice of the effective step size $\gamma_i$ at each iteration $i$. Further, we note that $\mathcal{L}^{\mathrm{TRC}}_i$ is again a cost functional of a classical unconstrained optimal control problem, see the proof of \Cref{prop: optimality for TR-SB} in \Cref{app: proofs}.

The following result shows that the new iterates $u_{i+1} = u_{i+1}^{\lambda_i}$, obtained by solving \eqref{eq: saddle-point problem} for the optimal Lagrange multiplier $\lambda_i$, can be derived from the previous iterates $u_i$ by reweighting with respect to the current target $\cev{\P}^{v_i}$. We refer to \Cref{app: proofs} for the proof.

\begin{proposition}[Geometric interpretation of the trust region update]
\label{prop: geometric interpretation of update}
Let $u_{i+1}^{\lambda} \coloneqq \argmin_{u \in \mathcal{U}} \mathcal{L}^{\mathrm{TR}}_i(u, \lambda)$ be the solution at iteration $i$. For any $\lambda \geq 0$, the path measure $\vec{\P}^{u_{i+1}^{\lambda}}$ can be interpreted as the projection of a geometric average of measures. In particular, it holds that
\begin{align}
\label{eq:annealing}
\frac{\mathrm{d}\vec{\P}^{u^{\lambda}_{i+1}}}{\mathrm{d}\vec{\P}^{u_{i}}}(X)
\propto \left(\frac{\mathrm{d}\cev{\P}^{v_i}}{\mathrm{d}\vec{\P}^{u_i}}(X) \right)^{\frac{1}{1+\lambda}}
= \left(\frac{p_\mathrm{target}}{\vec{\P}^{u_i}_T}(X_T) \right)^{\frac{1}{1+\lambda}} \eqqcolon w_i(X)^{\frac{1}{1+\lambda}}.
\end{align}
The proportionality constant depends only on $X_0$ and is fixed by the constraint $\vec{\P}^{u^\lambda_{i+1}}_0 = p_\mathrm{prior}$; explicitly, $\frac{\mathrm{d}\vec{\P}^{u^{\lambda}_{i+1}}}{\mathrm{d}\vec{\P}^{u_{i}}}(X) = w_i(X)^{\frac{1}{1+\lambda}} \big/ \E_{\vec{\P}^{u_i}}\big[w_i^{\frac{1}{1+\lambda}} \,\big|\, X_0\big]$.
\end{proposition}

\vspace{-0.1cm}
We can further derive an explicit formula for the value functions associated with the iterations in the trust region Schrödinger bridge problem. The proof is again stated in \Cref{app: proofs}.
\begin{proposition}[Optimality for trust region Schrödinger bridge problem]
\label{prop: optimality for TR-SB}
Let the value function be defined as the minimal cost-to-go conditioned on $X^u_t = x$:
\begin{equation}
    \label{eq: minimization prob}
    V_{i+1}^\lambda(x, t) := \inf_{u \in \mathcal{U}} \mathcal{L}^\mathrm{TRC}_i(u, \lambda; x, t).
\end{equation}
Then, the value function satisfies
\begin{align}
\begin{split}
\label{eq: value function lambda}
    &V_{i+1}^\lambda(x, t) = -(1 + \lambda) \log \E_{\vec{\P}^{u_i}}\left[\left(\frac{\mathrm d \vec{\P}^{u_i}_{[t, T]}}{\mathrm d \cev{\P}^{v_i}_{[t, T]}}(X) \right)^{-\frac{1}{1 + \lambda}}\Bigg| X_t=x \right],
\end{split}
\end{align}
where the Radon-Nikodym derivative restricted to $[t, T]$ is given explicitly in \Cref{lem: path likelihood ratio}.
Moreover, the optimal control admits the representation
    $u_{i+1}^{\lambda} = \frac{\lambda}{1+\lambda}u_i - \frac{1}{1+\lambda}v_i - \frac{1}{1+\lambda}\sigma^\top \nabla \big(V_{i+1}^\lambda - \log \vec{\P}^{u_i}_t\big).$
\end{proposition}

A key insight from \Cref{prop: optimality for TR-SB} is that it provides an explicit representation of the value function $V_{i+1}^\lambda$ for arbitrary $\lambda$, up to expectation terms that can be efficiently approximated via Monte Carlo sampling. 
This representation allows us to identify the optimal dual variable $\lambda_i$ by maximizing the associated Lagrangian dual function. Specifically, substituting the optimal control $u_{i+1}^\lambda$ into the objective yields the dual function
\begin{align}
    \label{eq: dual function}
    \mathcal{L}^{\mathrm{Dual}}_i(\lambda) 
    &:= \mathcal{L}^{\mathrm{TR}}_i(u_{i+1}^{\lambda}, \lambda) = \mathbb{E}_{X_0 \sim p_\mathrm{prior}}[V^{\lambda}_{i+1}(X_0, 0)] - \lambda \varepsilon.
\end{align}
Its first term depends on the trajectories only through the weights $w_i$ in \eqref{eq:annealing}, which also enter the matching loss \eqref{eq: CE matching}. A Monte Carlo estimate of $\mathcal{L}^{\mathrm{Dual}}_i(\lambda)$ only requires trajectories of the current forward process $X^{u_i}$ and energy evaluations $\mathcal{E}(X_T)$, but neither gradients of $\mathcal{E}$ nor differentiation through the simulation (\Cref{alg:tr-sbs-detail}). The optimal dual
\begin{equation}
    \lambda_i = \argmax_{\lambda \ge 0} \mathcal{L}^{\mathrm{Dual}}_i(\lambda)
\end{equation}
is the maximizer of a concave one-dimensional function and can therefore be computed with standard scalar optimization routines.

In principle, one could then define the next control as $u_{i+1} = u_{i+1}^{\lambda_i}$ using the characterization in \Cref{prop: optimality for TR-SB}. However, this requires Monte Carlo estimates of the value function and its gradient $\nabla V^{\lambda_i}_{i+1}$ at every state $x$, which is infeasible in high-dimensional state spaces. We therefore learn a parameterized approximation of $u_{i+1}$, as described in the next section.

\subsection{Learning the constrained optimal control}
\label{sec: learning constrained optimal control}

In this section, we describe how the optimal control is learned at each iteration. To this end, we introduce the cross-entropy loss defined via the forward KL divergence~\cite{nusken2021solving}
\begin{equation}
\label{eq: cross-entropy loss}
    \mathcal{L}_i^\mathrm{CE}(u) := \KL(\vec{\P}^{u_{i+1}} | \vec{\P}^{u}).
\end{equation}
This objective becomes tractable by means of importance sampling, since trajectories $X^{u_i}$ can be generated from the previous iterate:
\begin{equation}
    \mathcal{L}_i^\mathrm{CE}(u)
    =
    \E_{\vec{\P}^{u_i}}\!\left[
        \left(\log \frac{\mathrm d \vec{\P}^{u_{i+1}}}{\mathrm d \vec{\P}^{u}}(X^{u_i})\right)
        \frac{\mathrm d \vec{\P}^{u_{i+1}}}{\mathrm d \vec{\P}^{u_i}}(X^{u_i})
    \right].
\end{equation}

In general, importance reweighting introduces exponential terms that may lead to large variance and numerical instabilities, particularly in high-dimensional settings; see, e.g., \cite{hartmann2024nonasymptotic}. Omitting the importance weights leads to a fixed-point type iteration, as in ASBS \cite{liu2025adjoint}, whose objective no longer corresponds to a proper divergence and only uses the gradient of the target rather than the target itself. Gradient information alone does not determine the relative weights of well-separated modes (mode blindness), which is problematic for multimodal targets \cite{ouyang2026diffusive}.

In our setting, however, the trust region keeps the variance of the log-weights at about $2\varepsilon$ for small $\varepsilon$ (\Cref{sec: convergence}; see also \Cref{app: variance of importance weights} and, for the weights, \cite[Appendix I.3]{blessing2025trust}), which allows us to retain the principled importance reweighting. At the same time, the cross-entropy loss admits a least-squares matching representation \citep{domingo2024stochastic}.

\begin{proposition}[Generalized adjoint matching representation of the cross-entropy loss]
\label{eq: CE as matching objective}
Up to a positive factor and an additive constant, the cross-entropy loss \eqref{eq: cross-entropy loss} can be equivalently expressed as
\begin{align}
\label{eq: CE matching}
    \begin{split}
    \mathcal{L}_i^\mathrm{F}(u) \! = \mathbb{E}_{\vec{\P}^{u_i}_{0, T}\vec{\P}^{\mathrm{ref}}_{t|0,T}} \! \Bigg[ \! \Big\| u(X_t, t) \! + \! \frac{v_i(X_T, T) \! + \! \sigma^{\top} \nabla \mathcal{E}(X_T)}{1+\lambda_i} \! - \! \frac{\lambda_i}{1+\lambda_i} u_i(X_T, T) \Big\|^2 \, \frac{\mathrm{d}\vec{\P}^{u_{i+1}}}{\mathrm d \vec{\P}^{u_i}} (X) \Bigg],%
    \end{split}
\end{align}
where the importance weight $\frac{\mathrm{d}\vec{\P}^{u_{i+1}}}{\mathrm d \vec{\P}^{u_i}} \propto \big(\frac{\mathrm{d}\cev{\P}^{v_i}}{\mathrm d \vec{\P}^{u_i}}\big)^{\frac{1}{1+\lambda_i}}$ is given in \Cref{prop: geometric interpretation of update}, $\mathcal{E}(x) = -\log p_{\mathrm{target}}(x) - \log \mathcal{Z}$ denotes the energy function, the expectation is over endpoints $(X_0, X_T) \sim \vec{\P}^{u_i}_{0, T}$, $t \sim \mathcal{U}(0,T)$, and $X_t \sim \vec{\P}^{\mathrm{ref}}_{t|0,T}(\cdot \,|\, X_0, X_T)$, the reference process conditioned on these endpoints. For a time-dependent $\sigma$, the terms after $u(X_t, t)$ in \eqref{eq: CE matching} are evaluated with $\sigma = \sigma(T)$ and multiplied by $\sigma(t)^\top \sigma(T)^{-\top}$ (\Cref{app: proof CE matching}).
\end{proposition}

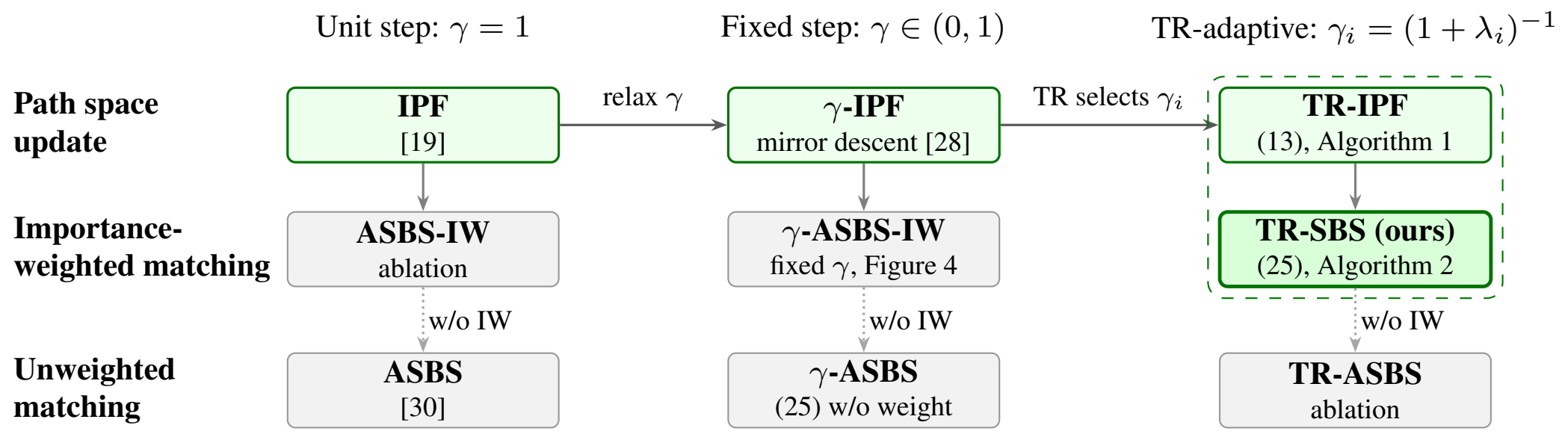
\begin{figure}[t]
\centering
\begin{tikzpicture}[
  font=\footnotesize,
  alg/.style={draw=green!45!black, fill=green!7, rounded corners=2pt, minimum width=2.4cm, minimum height=0.66cm, align=center, line width=0.6pt, inner sep=1.5pt},
  met/.style={draw=black!40, fill=black!5, rounded corners=2pt, minimum width=2.4cm, minimum height=0.66cm, align=center, inner sep=1.5pt},
  arr/.style={-{Stealth[length=4pt]}, line width=0.6pt, draw=black!65},
  lab/.style={anchor=west, align=left, font=\footnotesize\bfseries},
  ]
  \def\xa{3.55} \def\xb{7.45} \def\xc{11.8}
  \def\ya{0} \def\yb{-1.1} \def\yc{-2.35}
  \node[font=\footnotesize] at (\xa,0.85) {Unit step: $\gamma = 1$};
  \node[font=\footnotesize] at (\xb,0.85) {Fixed step: $\gamma \in (0,1)$};
  \node[font=\footnotesize] at (\xc,0.85) {TR-adaptive: $\gamma_i = (1+\lambda_i)^{-1}$};
  \node[lab, text=black] at (-0.2,\ya) {Path space\\update};
  \node[lab, text=black] at (-0.2,\yb) {Importance-\\weighted matching};
  \node[lab, text=black] at (-0.2,\yc) {Unweighted\\matching};
  \node[alg] (ipf) at (\xa,\ya) {\textbf{IPF}\\[-1pt]{\scriptsize\citep{fortet1940resolution}}};
  \node[alg] (gipf) at (\xb,\ya) {\textbf{$\gamma$-IPF}\\[-1pt]{\scriptsize mirror descent \citep{karimi2023sinkhorn}}};
  \node[alg] (tripf) at (\xc,\ya) {\textbf{TR-IPF}\\[-1pt]{\scriptsize \eqref{eq: TR objective}, \Cref{alg: TR-IPF}}};
  \draw[arr] (ipf) -- node[above, font=\scriptsize] {relax $\gamma$} (gipf);
  \draw[arr] (gipf) -- node[above, font=\scriptsize] {TR selects $\gamma_i$} (tripf);
  \node[met] (asbsiw) at (\xa,\yb) {\textbf{ASBS-IW}\\[-1pt]{\scriptsize ablation}};
  \node[met] (gasbsiw) at (\xb,\yb) {\textbf{$\gamma$-ASBS-IW}\\[-1pt]{\scriptsize fixed $\gamma$, \Cref{fig:ablation_gamma_mw}}};
  \node[alg, fill=green!15, line width=0.9pt] (trsbs) at (\xc,\yb) {\textbf{TR-SBS (ours)}\\[-1pt]{\scriptsize \eqref{eq: CE matching}, \Cref{alg: TR-SBS high level}}};
  \node[met] (asbs) at (\xa,\yc) {\textbf{ASBS}\\[-1pt]{\scriptsize\citep{liu2025adjoint}}};
  \node[met] (gasbs) at (\xb,\yc) {\textbf{$\gamma$-ASBS}\\[-1pt]{\scriptsize \eqref{eq: CE matching} w/o weight}};
  \node[met] (asbstr) at (\xc,\yc) {\textbf{TR-ASBS}\\[-1pt]{\scriptsize ablation}};
  \foreach \pa/\pb/\pc in {ipf/asbsiw/asbs, gipf/gasbsiw/gasbs, tripf/trsbs/asbstr} {
    \draw[arr, draw=black!50] (\pa.south) -- (\pb.north);
    \draw[arr, draw=black!35, densely dotted] (\pb.south) -- node[right, pos=0.5, inner sep=1pt, font=\scriptsize, text=black] {w/o IW} (\pc.north);
  }
  \node[draw=green!45!black, dashed, rounded corners=4pt, inner sep=2.5pt, fit=(tripf) (trsbs)] (ours) {};
\end{tikzpicture}
\caption{\textbf{Overview of IPF-based samplers.} The columns differ in how the mirror descent step size of the forward update is chosen. The rows show the path space update, which reweights the current iterate $\vec{\P}^{u_i}$ by $w_i^{\gamma}$ (\Cref{prop: geometric interpretation of update}), its implementation by matching objectives learned from samples of $\vec{\P}^{u_i}$ with this importance weight (IW), and the variants without it.}
\label{fig: overview}
\end{figure}

The objective \eqref{eq: CE matching} thus retains the simulation-efficient matching structure of \cite{liu2025adjoint} while remaining a proper divergence that incorporates target information through the importance weight; without this weight, one obtains fixed-point iterations such as ASBS, and with an inactive KL constraint ($\lambda_i = 0$, $\gamma_i = 1$), TR-SBS reduces to ASBS-IW, i.e., ASBS with the importance weight (\Cref{fig: overview} and \Cref{app: relation to ASBS}).

To complete the IPF scheme \eqref{eq: IPF iterations}, it remains to perform the $v$-updates. We employ a generalized matching objective based on \cite{liu2025adjoint}. A key distinction from standard bridge matching approaches \citep{peluchetti2023non,peluchetti2023diffusion,shi2023diffusion}, which rely on ground-truth target samples $X_T \sim p_{\mathrm{target}}$, is that our formulation utilizes samples $X_T \sim \vec{\P}^{u_{i+1}}_{T}$ generated by the current forward iterate. Specifically, we define the loss as
\begin{equation}
\label{eq: loss for v}
    \mathcal{L}^\mathrm{B}_i(v)  = \mathbb{E}_{ \vec{\P}^{u_{i+1}}_{0, T} \vec{\P}^{\mathrm{ref}}_{t|0,T}} \left[  \zeta_t  \left\| v(X_t, t)  - \sigma^\top \nabla \log p^{\mathrm{ref}}_{t|0}(X_t | X_0) \right\|^2 \right],
\end{equation}
where $\zeta_t = (1+\sigma(t)^2 (T-t)/t)^{-1}$ is a time-dependent weighting factor, cf. \cite{shi2023diffusion}, and the expectation is taken as in \eqref{eq: CE matching} but with endpoints from $\vec{\P}^{u_{i+1}}_{0,T}$ and without importance weight. Its minimizer is the backward IPF half-step (\Cref{lem: backward matching}). The overall training procedure of our Trust Region Schrödinger Bridge Sampler (TR-SBS) is summarized in \Cref{alg: TR-SBS high level} in \Cref{app:implementation}. The diffusion scale $\sigma_{\max}$ and the computational cost are discussed in \Cref{app: sigma ablation,app: computational cost}.

\subsection{Convergence of TR-IPF: speed and stability}
\label{sec: convergence}
We now analyze how the trust region affects convergence in the idealized setting in which each step of TR-SBS is computed exactly, i.e., without sampling and optimization errors, so that TR-SBS reduces to TR-IPF; we discuss deviations from this setting in \Cref{sec: limitations}. To this end, let $\P^*$ minimize $F(\P) \coloneqq \KL(\P_T | p_\mathrm{target})$ over path measures with $\P_0 = p_\mathrm{prior}$. As stated in \Cref{sec:mirror_descent}, IPF is mirror descent on $F$ with unit step size, so TR-IPF is mirror descent with the adaptive step size $\gamma_i$. This yields the following result, which we prove in \Cref{app: convergence}.
\begin{proposition}[Convergence of TR-IPF]
\label{thm:convergence_rate}
Let $(\vec{\P}^{u_i})_{i \ge 0}$ be generated by TR-IPF (\Cref{alg: TR-IPF}) and $\ell_i \coloneqq \log \frac{\mathrm d \cev{\P}^{v_i}}{\mathrm d \vec{\P}^{u_i}}$. Assume that the constraint is active, $\KL(\vec{\P}^{u_{i+1}} | \vec{\P}^{u_i}) = \varepsilon$, and that $\operatorname{Var}_{\vec{\P}^{u_i}}(\ell_i) \le M$ for all $i$. Then $F(\vec{\P}^{u_{i+1}}) \le F(\vec{\P}^{u_i})$ for all $i$, and, at leading order for small $\varepsilon$,
\begin{equation}
\label{eq: convergence rate}
\gamma_i \gtrsim \sqrt{2\varepsilon / M}, \qquad
F(\vec{\P}^{u_I}) - F(\P^*) \lesssim \frac{\KL(\P^* | \vec{\P}^{u_0}) \sqrt{M}}{I \sqrt{2\varepsilon}}.
\end{equation}
\end{proposition}
Both inequalities in \eqref{eq: convergence rate} rest on the small-step expansion $\varepsilon = \KL(\vec{\P}^{u_{i+1}} | \vec{\P}^{u_i}) \approx \gamma_i^2 \widebar{V}_i / 2$, where $\widebar{V}_i \coloneqq \E_{X_0 \sim p_\mathrm{prior}}[\operatorname{Var}_{\vec{\P}^{u_i}}(\ell_i \,|\, X_0)] \le M$, which determines the step size $\gamma_i \approx \sqrt{2\varepsilon / \widebar{V}_i}$ as well as the variance of the log-weights.

\textbf{Speed.} With a fixed step size $\gamma$, the gap after $I$ iterations is at most $\KL(\P^* | \vec{\P}^{u_0}) / (\gamma I)$ (\Cref{lem: mirror descent step}). The trust region acts as a step size of at least $\sqrt{2\varepsilon/M}$, which gives \eqref{eq: convergence rate}, i.e., $I = \mathcal{O}(\varepsilon^{-1/2})$ iterations for a given gap; the step grows automatically when $\widebar{V}_i$ is small, e.g., as $\vec{\P}^{u_i}$ approaches the target.

\textbf{Stability.} By \Cref{prop: geometric interpretation of update}, $\gamma_i \ell_i$ is, up to a function of $X_0$, the logarithm of the importance weight of the update. Its conditional variance, averaged over $X_0$, is thus $\gamma_i^2 \widebar{V}_i \approx 2\varepsilon$ at leading order. With a fixed step size, it is $\gamma^2 \widebar{V}_i$, which becomes large when $\vec{\P}^{u_i}$ puts mass where the target is small, e.g., in early iterations, so that the importance weights degenerate. The radius $\varepsilon$ thus trades stability for speed. Conversely, if $\widebar{V}_i$ is very large, the step size $\gamma_i \approx \sqrt{2\varepsilon / \widebar{V}_i}$ becomes so small that the learned updates stall, as we observe in some of our experiments. Bounding $\gamma_i$ from below (\Cref{app:aldp_detail}) restores progress but gives up the variance control, since then $\gamma_i^2 \widebar{V}_i > 2\varepsilon$; the trust region alone does not resolve this dilemma between stagnation and degenerate weights.

\section{Experiments}
\label{sec: experiments}

We evaluate our proposed method on high-dimensional sampling tasks designed to rigorously test its mode-seeking capabilities and scalability. Specifically, we consider a high-dimensional Many-Well (MW) potential characterized by exponentially many modes, a Gaussian Mixture Model (GMM) that affords precise control over mode placement, and alanine dipeptide, a molecular benchmark with several metastable conformations, one of which is rare. On MW, we report the marginal Earth Mover's Distance (EMD) and the error of the log-normalizing constant ($|\Delta \log Z|$), and on GMM, the Sinkhorn divergence (SD) \cite{genevay2018learning} and the total variation distance (TVD) of the mixture weights; definitions and further metrics are given in \Cref{app:exp_details,app: additional experiments}.

To isolate the contributions of the two ingredients of TR-SBS, the trust region and the importance weight, we compare \textbf{TR-SBS} with the original \textbf{ASBS} \citep{liu2025adjoint} and two ablations that each add one of them to ASBS (\Cref{fig: overview}): \textbf{ASBS-IW}, which adds only the importance weight, i.e., minimizes the forward KL divergence with unit step size, and \textbf{TR-ASBS}, which selects the step size by the trust region without importance weighting.
On GMM, we additionally compare with two diffusion-based samplers outside the ASBS family, \textbf{PIS} \cite{zhang2022path} and \textbf{DIS} \cite{berner2022optimal} (\Cref{app: pis dis}).

Qualitative visualizations and further results are provided in~\cref{app: additional experiments}, including a $\phi^4$ scalar field theory model (\Cref{sec: Ginzburg-Landau}) and the Lennard-Jones cluster LJ-13 (\Cref{app: lj13}).

\subsection{Many-Well potential}

To evaluate the sampler's ability to escape local minima and cross energy barriers, we employ the MW potential~\cite{wu2020stochastic,richter2023improved}. The energy function $\mathcal{E}$ is given by
\begin{equation}
    \mathcal{E}(x) = \sum_{i=1}^m (x_i^2 - \delta)^2 + \frac{1}{2} \sum_{j=m+1}^d x_j^2.
\end{equation}
Here, the first $m$ dimensions represent double-well potentials, creating $2^m$ modes. %

\begin{figure*}[t]
    \centering
    \setlength{\capshift}{0.175in}
    \includegraphics{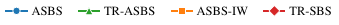}\\[-0.3em]
    \begin{subfigure}{0.24\textwidth}
        \includegraphics[width=\linewidth]{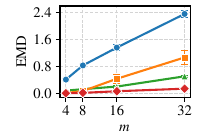}
        \shiftedcaption{EMD (vs. $m$)}
    \end{subfigure}
    \hfill
    \begin{subfigure}{0.24\textwidth}
        \includegraphics[width=\linewidth]{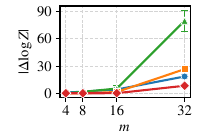}
        \shiftedcaption{$|\Delta \log Z|$ (vs. $m$)}
    \end{subfigure}
    \hfill
    \begin{subfigure}{0.24\textwidth}
        \includegraphics[width=\linewidth]{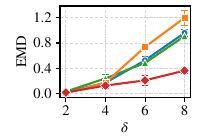}
        \shiftedcaption{EMD (vs. $\delta$)}
    \end{subfigure}
    \hfill
    \begin{subfigure}{0.24\textwidth}
        \includegraphics[width=\linewidth]{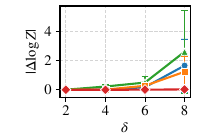}
        \shiftedcaption{$|\Delta \log Z|$ (vs. $\delta$)}
    \end{subfigure}
    
    \caption{\textbf{Results on Many-Well potential.} We evaluate robustness to increasing modes and robustness to mode distance. %
    \textbf{(a, b)} Performance vs. number of modes. 
    \textbf{(c, d)} Performance vs. mode distance. 
    Our proposed method (TR-SBS, red) achieves the lowest estimation error and marginal distance for large $m$ and $\delta$, handling both combinatorial complexity and high energy barriers.}
    \label{fig:mw_results}
\end{figure*}

\textbf{Results.} For $d=64$ and $m \in \{4, \dots, 32\}$ double-well dimensions (\Cref{fig:mw_results} (a, b)), \textbf{TR-SBS} (red) performs best in both metrics, which shows that combining the trust region with importance weighting is essential for accurate global sampling in multimodal landscapes. For $d=8$, $m=5$, and barrier distances $\delta \in \{2, \dots, 8\}$ (\Cref{fig:mw_results} (c, d)), TR-SBS attains the lowest marginal EMD and log-partition error for $\delta \geq 4$, i.e., it is robust to large separations and balances the probability mass between distant modes.

\begin{figure*}[t]
    \centering
    \setlength{\capshift}{0.175in}
    \includegraphics{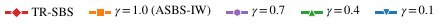}\\[-0.3em]
    \begin{subfigure}{0.24\textwidth}
        \includegraphics[width=\linewidth]{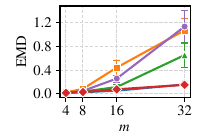}
        \shiftedcaption{EMD (vs. $m$)\label{fig:ablation_mw_summary}}
    \end{subfigure}
    \hfill
    \begin{subfigure}{0.24\textwidth}
        \includegraphics[width=\linewidth]{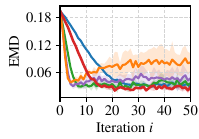}
        \shiftedcaption{EMD\label{fig:ablation_epoch_emd}}
    \end{subfigure}
    \hfill
    \begin{subfigure}{0.24\textwidth}
        \includegraphics[width=\linewidth]{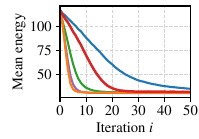}
        \shiftedcaption{Mean energy\label{fig:ablation_epoch_energy}} %
    \end{subfigure}
    \hfill
    \begin{subfigure}{0.24\textwidth}
        \includegraphics[width=\linewidth]{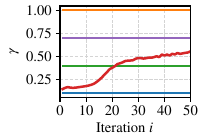}
        \shiftedcaption{Gamma\label{fig:ablation_gamma_iter}} %
    \end{subfigure}
    
    \caption{\textbf{Ablation study on the step size $\gamma$.} Legend is shared across all subplots. 
    \textbf{(a)} Adaptive TR-SBS matches the robustness of conservative $\gamma=0.1$.
    \textbf{(b, c)} Dynamics reveal the trade-off: Small $\gamma=0.1$ is stable (b) but slow (c). TR-SBS converges efficiently (c) while maintaining stability (b).
    \textbf{(d)} The adaptive $\gamma$ starts conservative to prevent collapse and increases to accelerate convergence.}
    \label{fig:ablation_gamma_mw}
\end{figure*}

\paragraph{Effect of the mirror descent step size $\gamma = (1+\lambda)^{-1}$}
We compare TR-SBS with $\gamma$-ASBS-IW (\Cref{fig: overview}) using fixed step sizes $\gamma \in \{1.0, 0.7, 0.4, 0.1\}$, where $\gamma = 1$ is ASBS-IW, on MW with $d=64$, analyzing both robustness to mode complexity ($m$) and training dynamics ($m=8$); see \Cref{fig:ablation_gamma_mw}.

\emph{Stability:} \Cref{fig:ablation_mw_summary} shows that aggressive fixed step sizes ($\gamma \ge 0.4$) fail to prevent mode collapse as combinatorial complexity increases ($m \to 32$). A conservative step size ($\gamma \approx 0.1$) is required to maintain low error, a level of robustness matched by our adaptive TR-SBS.
    
\emph{Efficiency vs. Stability:} Since each outer iteration uses $N_{\mathrm{FC}} N_{\mathrm{new}}$ target-gradient evaluations, the iteration axis in \Cref{fig:ablation_gamma_mw} is also proportional to their cumulative number, which scales as $\mathcal{O}(I N_{\mathrm{FC}} N_{\mathrm{new}})$. \Cref{fig:ablation_epoch_energy} shows that a small step $\gamma=0.1$ results in slow convergence of the mean energy $\mathbb{E}[\mathcal{E}(X_T)]$, whereas TR-SBS (red) converges more than twice as fast. However, \Cref{fig:ablation_epoch_emd} shows that aggressive steps converge to a higher error, while TR-SBS attains the low final error of conservative steps. As shown in \Cref{fig:ablation_gamma_iter}, the adaptive $\gamma$ starts small ($\approx 0.15$) for initial stability and increases as the training stabilizes, as expected from \Cref{sec: convergence}. Finally, we refer to an ablation study on the impact of the maximum noise scale in \Cref{app: sigma ablation}.

\subsection{Gaussian mixture model (GMM)}
\label{sec: gmm}

\begin{figure*}[t]
    \centering
    \setlength{\capshift}{0.2in}
    \includegraphics{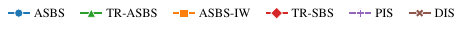}\\[-0.3em]
    \begin{subfigure}{0.24\textwidth}
        \includegraphics[width=\linewidth]{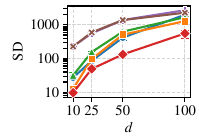}
        \shiftedcaption{SD (vs. $d$)}
    \end{subfigure}
    \hfill
    \begin{subfigure}{0.24\textwidth}
        \includegraphics[width=\linewidth]{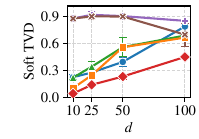}
        \shiftedcaption{TVD (vs. $d$)}
    \end{subfigure}
    \hfill
    \begin{subfigure}{0.24\textwidth}
        \includegraphics[width=\linewidth]{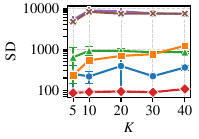}
        \shiftedcaption{SD (vs. $K$)}
    \end{subfigure}
    \hfill
    \begin{subfigure}{0.24\textwidth}
        \includegraphics[width=\linewidth]{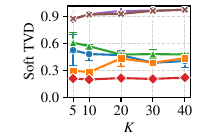}
        \shiftedcaption{TVD (vs. $K$)}
    \end{subfigure}
    
    \caption{\textbf{Results on GMM} (mean $\pm$ standard deviation over three seeds). 
    \textbf{(a, b)} Performance vs.\ dimension $d \in \{10, 25, 50, 100\}$ with $K=10$. 
    \textbf{(c, d)} Performance vs.\ number of components $K \in \{5, 10, 20, 30, 40\}$ with $d=10$. 
    TR-SBS (red) attains the lowest SD and TVD in every setting. PIS and DIS use 128 instead of 200 time steps and a different training budget (\Cref{app: pis dis}).}
    \label{fig:gmm_results}
\end{figure*}

We verify the scalability and robustness of the sampler using a GMM, $p_{\mathrm{target}}(x) = \sum_{k=1}^{K} \pi_k \, \mathcal{N}(x | \mu_k, s^2 I_d)$ with $s = \ln (1+e)$, means $\mu_k$ sampled uniformly from $[-L, L]^d$, and random mixing weights with $\max_k \pi_k / \min_k \pi_k \le 3$. %
We use the TVD between the true and empirical mixing weights as a primary metric for detecting mode collapse, alongside the SD to assess geometric fidelity. For each run, we evaluate the checkpoint with the lowest SD during training on $4{,}000$ independent samples; all settings are given in \Cref{app:gmm_detail}.

\textbf{Results.} For $K=10$, $L=7$, and $d \in \{10, 25, 50, 100\}$, \Cref{fig:gmm_results} (a, b) shows that TR-SBS attains the lowest SD and TVD in every dimension, and its advantage grows with $d$ (e.g., TVD $0.23$ vs.\ at least $0.40$ for the ASBS variants at $d=50$). Neither importance weighting with unit step size (ASBS-IW) nor the trust region without importance weighting (TR-ASBS) is consistently better than ASBS. Only their combination attains the lowest error for all $d$, which suggests that the two components are complementary. PIS and DIS concentrate on few components in this setting (TVD $\geq 0.70$ for all $d$). For $d=10$, widely separated components ($L=40$), and $K \in \{5, 10, 20, 30, 40\}$ (\Cref{fig:gmm_results} (c, d)), TR-SBS keeps the TVD at about $0.2$ for all $K$, while all other methods have a TVD of at least $0.28$ and an SD at least twice as large. The error of ASBS-IW grows with $K$, which shows that importance-weighted updates need the small steps enforced by the trust region. \Cref{fig:gmm_vis_k40} in \Cref{app: gmm visualization} visualizes the samples for $K=40$.

\begin{figure}[t]
\centering
\includegraphics[width=\linewidth]{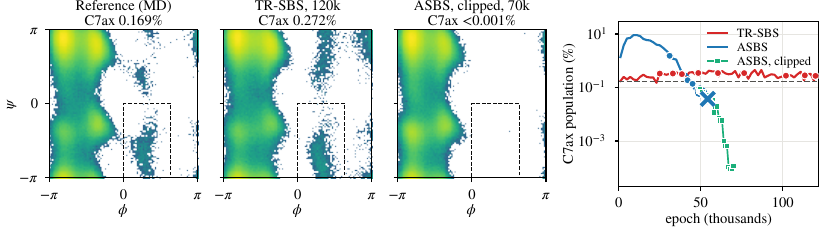}
\caption{Alanine dipeptide. Left: Ramachandran plots of the reference, of TR-SBS after $120{,}000$ epochs, and of ASBS with clipped corrector targets after $70{,}000$ epochs (\Cref{app: aldp results}); the dashed box marks C7ax. Right: C7ax population over training (lines: evaluations of $2{,}000$ samples, averaged over $2{,}000$ epochs; dots: $10^7$ samples, open dots from ASBS resumed at $48{,}500$ epochs, see \Cref{app: aldp results}; squares: ASBS with clipped corrector targets, $10^7$ samples; cross: divergence of ASBS without clipping; dashed: reference).}
\label{fig:aldp}
\vspace{0.6em}
\small

\captionof{table}{Alanine dipeptide ($10^7$ samples): ASBS before (\Cref{app: aldp results}) and after its divergence, ASBS with clipped corrector targets, and TR-SBS at the end of training. After the divergence, $19\%$ of the ASBS samples have the wrong chirality. Last row: from \cite{liu2025adjoint}, with a Gaussian prior and another budget (\Cref{app:aldp_detail}).}
\label{tab:aldp_results}
\begin{tabular}{llcccc}
\hline
Method & Epoch & KL$_\phi \downarrow$ & KL$_\psi \downarrow$ & $W_2 \downarrow$ & C7ax (\%) \\
\hline
ASBS (before divergence) & 54k   & 0.0457 & 0.0276 & 0.574 & 0.034 \\
ASBS (diverged)          & 58.5k & 0.2967 & 0.3786 & 1.784 & 0.193 \\
ASBS, clipped     & 70k   & 0.0252 & 0.0406 & 0.832 & $<$0.001 \\
TR-SBS                   & 120k  & 0.0144 & 0.0037 & 0.246 & 0.272 \\
\hline
ASBS \cite{liu2025adjoint} & -- & $0.02 \pm 0.00$ & $0.01 \pm 0.00$ & $0.25 \pm 0.01$ & -- \\
\hline
\end{tabular}
\end{figure}

\begingroup
\subsection{Alanine dipeptide}
\label{sec: aldp}
Finally, we sample the Boltzmann distribution of alanine dipeptide in internal coordinates ($d=60$), as in \cite{midgley2023flow,liu2025adjoint} (\Cref{app:aldp_detail}). Its conformations are characterized by the two backbone torsion angles $\phi$ and $\psi$, i.e., the rotations around the N-C$_\alpha$ and C$_\alpha$-C bonds, whose joint distribution is shown as a Ramachandran plot in \Cref{fig:aldp}. Besides its dominant conformations, the molecule has a rare metastable conformation, C7ax, which holds only about $0.17\%$ of the probability mass (dashed box). Transitions into the region $\phi > 0$, which contains C7ax, are a classical rare event problem in molecular simulation \cite{bolhuis2000reaction}, and even flows trained on molecular dynamics (MD) samples struggle to model this weakly populated mode \cite{midgley2023flow}. Missing it cannot be corrected by reweighting the final samples and biases estimates such as conformational free energies. \cite{liu2025adjoint} report that ASBS misses such low-density conformations and suggest importance weighting in training as a remedy, which TR-SBS uses. To make it possible for both methods to reach C7ax at all, we use the same prior for both, which spreads a small fraction of its samples along $\phi$ (\Cref{app:aldp_detail}). Following \cite{liu2025adjoint}, we report KL divergences of the torsion marginals and $W_2$ on $(\phi, \psi)$, and in addition the C7ax population. \Cref{fig:aldp} shows that ASBS first overpopulates C7ax by a factor of about $50$, then drains it to about a fifth of the reference population, and diverges between $54{,}000$ and $54{,}500$ epochs, also when training is resumed from an earlier checkpoint; afterwards, its samples are far from the target, e.g., $19\%$ of them have the wrong chirality (\Cref{tab:aldp_results}). With corrector targets clipped in addition to the clipping used for both methods, ASBS does not diverge within $70{,}000$ epochs but drains C7ax further, to $0.0001\%$ (\Cref{app: aldp results}). TR-SBS, whose updates are bounded by the trust region, keeps the C7ax population close to the reference throughout training and trains stably for $120{,}000$ epochs.
\endgroup

\section{Conclusion and limitations}
\label{sec: limitations}
We proposed a trust region approach for computing Schrödinger bridges from an unnormalized target density.
Building on the mirror descent interpretation of IPF, it is, to our knowledge, the first method to select the IPF step size adaptively, via a KL constraint on path space that locally controls the variance of the importance weights.
On Many-Well, GMM, and alanine dipeptide, it mitigates the mode collapse of ASBS.

The trust region radius $\varepsilon$, however, trades stability for speed (\Cref{sec: convergence}): a small trust region stabilizes the updates but requires more outer iterations, so that TR-SBS can need several times more target-gradient evaluations than ASBS, which converges faster but can collapse to a subset of the modes (\Cref{app: computational cost}). Moreover, note that our theoretical analysis concerns the idealized iteration: \Cref{thm:convergence_rate} and the variance argument are local and assume exact updates, whereas TR-SBS learns each step from samples, normalizes and clips the weights, and, on alanine dipeptide and LJ-13, bounds $\gamma_i$ from below and changes the matching anchor (\Cref{app: weight normalization,app:aldp_detail}). Like ASBS, it is also sensitive to the number of training epochs per update and to the diffusion scale (\Cref{app: sigma ablation}). Adapting the radius during training and analyzing the iteration with learned updates are natural next steps.

\newpage
\begin{ack}
This work was supported by JST CREST Grant Number JPMJCR24Q5 and JST ASPIRE Grant Number JPMJAP2329, Japan, and by the Horizon Europe Marie Sk\l{}odowska-Curie Actions Staff Exchanges project 101131557 (REMODEL).
\end{ack}

\bibliographystyle{abbrv}
\bibliography{references}

\begin{thebibliography}{10}

\bibitem{abdolmaleki2015model}
A.~Abdolmaleki, R.~Lioutikov, N.~Lau, L.~P. Reis, J.~Peters, and G.~Neumann.
\newblock Model-based relative entropy stochastic search.
\newblock {\em Advances in Neural Information Processing Systems}, 28, 2015.

\bibitem{amari2016information}
S.-i. Amari.
\newblock {\em Information geometry and its applications}.
\newblock Springer, 2016.

\bibitem{aubin2022mirror}
P.-C. Aubin-Frankowski, A.~Korba, and F.~L{\'e}ger.
\newblock Mirror descent with relative smoothness in measure spaces, with
  application to {S}inkhorn and {EM}.
\newblock {\em Advances in Neural Information Processing Systems},
  35:17263--17275, 2022.

\bibitem{berner2025discrete}
J.~Berner, L.~Richter, M.~Sendera, J.~Rector-Brooks, and N.~Malkin.
\newblock From discrete-time policies to continuous-time diffusion samplers:
  Asymptotic equivalences and faster training.
\newblock {\em Transactions on Machine Learning Research}, 2026.

\bibitem{berner2022optimal}
J.~Berner, L.~Richter, and K.~Ullrich.
\newblock An optimal control perspective on diffusion-based generative
  modeling.
\newblock {\em Transactions on Machine Learning Research}, 2024.

\bibitem{bernton2019schr}
E.~Bernton, J.~Heng, A.~Doucet, and P.~E. Jacob.
\newblock Schr\"odinger bridge samplers.
\newblock {\em arXiv preprint arXiv:1912.13170}, 2019.

\bibitem{blessing2025trust}
D.~Blessing, J.~Berner, L.~Richter, C.~Domingo-Enrich, Y.~Du, A.~Vahdat, and
  G.~Neumann.
\newblock Trust region constrained measure transport in path space for
  stochastic optimal control and inference.
\newblock In {\em Advances in Neural Information Processing Systems},
  volume~38, pages 165462--165510, 2025.

\bibitem{blessing2025underdamped}
D.~Blessing, J.~Berner, L.~Richter, and G.~Neumann.
\newblock Underdamped diffusion bridges with applications to sampling.
\newblock In {\em International Conference on Learning Representations}, 2025.

\bibitem{blessing2026bridge}
D.~Blessing, L.~Richter, J.~Berner, E.~Malitskiy, and G.~Neumann.
\newblock Bridge matching sampler: Scalable sampling via generalized
  fixed-point diffusion matching.
\newblock In {\em International Conference on Machine Learning}, pages
  8496--8541. PMLR, 2026.

\bibitem{bolhuis2000reaction}
P.~G. Bolhuis, C.~Dellago, and D.~Chandler.
\newblock Reaction coordinates of biomolecular isomerization.
\newblock {\em Proceedings of the National Academy of Sciences},
  97(11):5877--5882, 2000.

\bibitem{chen2024sequential}
J.~Chen, L.~Richter, J.~Berner, D.~Blessing, G.~Neumann, and A.~Anandkumar.
\newblock Sequential controlled {L}angevin diffusions.
\newblock In {\em International Conference on Learning Representations}, 2025.

\bibitem{chen2021likelihood}
T.~Chen, G.-H. Liu, and E.~A. Theodorou.
\newblock Likelihood training of {S}chr{\"o}dinger bridge using
  forward-backward {SDE}s theory.
\newblock In {\em International Conference on Learning Representations}, 2022.

\bibitem{choi2025non}
J.~Choi, Y.~Chen, M.~Tao, and G.-H. Liu.
\newblock Non-equilibrium annealed adjoint sampler.
\newblock In {\em Advances in Neural Information Processing Systems},
  volume~38, pages 90956--90988, 2025.

\bibitem{conn2000trust}
A.~R. Conn, N.~I.~M. Gould, and P.~L. Toint.
\newblock {\em Trust region methods}.
\newblock SIAM, 2000.

\bibitem{dai1991stochastic}
P.~Dai~Pra.
\newblock A stochastic control approach to reciprocal diffusion processes.
\newblock {\em Applied Mathematics and Optimization}, 23(3):313--329, 1991.

\bibitem{de2021diffusion}
V.~De~Bortoli, J.~Thornton, J.~Heng, and A.~Doucet.
\newblock Diffusion {S}chr{\"o}dinger bridge with applications to score-based
  generative modeling.
\newblock {\em Advances in neural information processing systems},
  34:17695--17709, 2021.

\bibitem{domingoenrich2025adjoint}
C.~Domingo-Enrich, M.~Drozdzal, B.~Karrer, and R.~T.~Q. Chen.
\newblock Adjoint matching: Fine-tuning flow and diffusion generative models
  with memoryless stochastic optimal control.
\newblock In {\em International Conference on Learning Representations}, 2025.

\bibitem{domingo2024stochastic}
C.~Domingo-Enrich, J.~Han, B.~Amos, J.~Bruna, and R.~T.~Q. Chen.
\newblock Stochastic optimal control matching.
\newblock {\em Advances in Neural Information Processing Systems},
  37:112459--112504, 2024.

\bibitem{fortet1940resolution}
R.~Fortet.
\newblock R{\'e}solution d'un syst{\`e}me d'{\'e}quations de {M}.
  {S}chr{\"o}dinger.
\newblock {\em Journal de math{\'e}matiques pures et appliqu{\'e}es},
  19(1-4):83--105, 1940.

\bibitem{genevay2018learning}
A.~Genevay, G.~Peyr{\'e}, and M.~Cuturi.
\newblock Learning generative models with {S}inkhorn divergences.
\newblock In {\em International Conference on Artificial Intelligence and
  Statistics}, pages 1608--1617. PMLR, 2018.

\bibitem{ginzburg2009theory}
V.~L. Ginzburg and L.~D. Landau.
\newblock On the theory of superconductivity.
\newblock In {\em On superconductivity and superfluidity: a scientific
  autobiography}, pages 113--137. Springer, 2009.

\bibitem{gottwald2024localized}
G.~A. Gottwald and S.~Reich.
\newblock Localized {S}chr{\"o}dinger bridge sampler.
\newblock {\em Journal of Computational Physics}, 548:114583, 2026.

\bibitem{gruhlke2026tensor}
R.~Gruhlke, J.~Berner, D.~Sommer, and L.~Richter.
\newblock Tensor train diffusion: Leveraging low-rank structures for
  high-dimensional score-based sampling.
\newblock In {\em International Conference on Machine Learning}, pages
  36905--36941. PMLR, 2026.

\bibitem{hartmann2024nonasymptotic}
C.~Hartmann and L.~Richter.
\newblock Nonasymptotic bounds for suboptimal importance sampling.
\newblock {\em SIAM/ASA Journal on Uncertainty Quantification}, 12(2):309--346,
  2024.

\bibitem{hartmann2017variational}
C.~Hartmann, L.~Richter, C.~Sch{\"u}tte, and W.~Zhang.
\newblock Variational characterization of free energy: Theory and algorithms.
\newblock {\em Entropy}, 19(11):626, 2017.

\bibitem{havens2025adjoint}
A.~J. Havens, B.~K. Miller, B.~Yan, C.~Domingo-Enrich, A.~Sriram, D.~S. Levine,
  B.~M. Wood, B.~Hu, B.~Amos, B.~Karrer, X.~Fu, G.-H. Liu, and R.~T.~Q. Chen.
\newblock Adjoint sampling: Highly scalable diffusion samplers via adjoint
  matching.
\newblock In {\em International Conference on Machine Learning}, pages
  22204--22237. PMLR, 2025.

\bibitem{huang2021schrodinger}
J.~Huang, Y.~Jiao, L.~Kang, X.~Liao, J.~Liu, and Y.~Liu.
\newblock {S}chr{\"o}dinger-{F}{\"o}llmer sampler.
\newblock {\em IEEE Transactions on Information Theory}, 71(2):1283--1299,
  2025.

\bibitem{karimi2023sinkhorn}
M.~R. Karimi, Y.-P. Hsieh, and A.~Krause.
\newblock {S}inkhorn flow as mirror flow: A continuous-time framework for
  generalizing the {S}inkhorn algorithm.
\newblock In {\em International Conference on Artificial Intelligence and
  Statistics}, pages 4186--4194. PMLR, 2024.

\bibitem{leonard2013survey}
C.~L{\'e}onard.
\newblock A survey of the {S}chr{\"o}dinger problem and some of its connections
  with optimal transport.
\newblock {\em Discrete and Continuous Dynamical Systems}, 34(4):1533--1574,
  2014.

\bibitem{liu2025adjoint}
G.-H. Liu, J.~Choi, Y.~Chen, B.~K. Miller, and R.~T.~Q. Chen.
\newblock Adjoint {S}chr{\"o}dinger bridge sampler.
\newblock In {\em Advances in Neural Information Processing Systems},
  volume~38, pages 15673--15708, 2025.

\bibitem{liu2023generalized}
G.-H. Liu, Y.~Lipman, M.~Nickel, B.~Karrer, E.~A. Theodorou, and R.~T.~Q. Chen.
\newblock Generalized {S}chr{\"o}dinger bridge matching.
\newblock In {\em International Conference on Learning Representations}, 2024.

\bibitem{midgley2023flow}
L.~I. Midgley, V.~Stimper, G.~N.~C. Simm, B.~Sch{\"o}lkopf, and J.~M.
  Hern{\'a}ndez-Lobato.
\newblock Flow annealed importance sampling bootstrap.
\newblock In {\em International Conference on Learning Representations}, 2023.

\bibitem{nam2025enhancing}
J.~Nam, B.~M{\'a}t{\'e}, A.~P. Toshev, M.~Kaniselvan, R.~G{\'o}mez-Bombarelli,
  R.~T.~Q. Chen, B.~Wood, G.-H. Liu, and B.~K. Miller.
\newblock Enhancing diffusion-based sampling with molecular collective
  variables.
\newblock In {\em International Conference on Learning Representations}, 2026.

\bibitem{nusken2021solving}
N.~N{\"u}sken and L.~Richter.
\newblock Solving high-dimensional {H}amilton-{J}acobi-{B}ellman {PDE}s using
  neural networks: perspectives from the theory of controlled diffusions and
  measures on path space.
\newblock {\em Partial Differential Equations and Applications}, 2(4):48, 2021.

\bibitem{otto2021differentiable}
F.~Otto, P.~Becker, N.~A. Vien, H.~C. Ziesche, and G.~Neumann.
\newblock Differentiable trust region layers for deep reinforcement learning.
\newblock In {\em International Conference on Learning Representations}, 2021.

\bibitem{ouyang2026diffusive}
R.~Ouyang, L.~Grenioux, and J.~M. Hern{\'a}ndez-Lobato.
\newblock A diffusive classification loss for learning energy-based generative
  models.
\newblock In {\em International Conference on Machine Learning}, pages
  95197--95252. PMLR, 2026.

\bibitem{peluchetti2023diffusion}
S.~Peluchetti.
\newblock Diffusion bridge mixture transports, {S}chr{\"o}dinger bridge
  problems and generative modeling.
\newblock {\em Journal of Machine Learning Research}, 24(374):1--51, 2023.

\bibitem{peluchetti2023non}
S.~Peluchetti.
\newblock Non-denoising forward-time diffusions.
\newblock {\em arXiv preprint arXiv:2312.14589}, 2023.

\bibitem{peters2010relative}
J.~Peters, K.~M{\"u}lling, and Y.~Altun.
\newblock Relative entropy policy search.
\newblock In {\em Proceedings of the AAAI Conference on Artificial
  Intelligence}, volume~24, pages 1607--1612, 2010.

\bibitem{richter2023improved}
L.~Richter and J.~Berner.
\newblock Improved sampling via learned diffusions.
\newblock In {\em International Conference on Learning Representations}, 2024.

\bibitem{rosenstein2021ginzburg}
B.~Rosenstein and D.~Li.
\newblock {\em Ginzburg--Landau Theory of Condensates}.
\newblock Cambridge University Press, 2021.

\bibitem{ruschendorf1995convergence}
L.~R{\"u}schendorf.
\newblock Convergence of the iterative proportional fitting procedure.
\newblock {\em The Annals of Statistics}, 23(4):1160--1174, 1995.

\bibitem{satorras2021en}
V.~G. Satorras, E.~Hoogeboom, and M.~Welling.
\newblock E(n) equivariant graph neural networks.
\newblock In {\em International Conference on Machine Learning}, 2021.

\bibitem{schrodinger1931umkehrung}
E.~Schr{\"o}dinger.
\newblock {\"U}ber die {U}mkehrung der {N}aturgesetze.
\newblock {\em Sitzungsberichte der Preussischen Akademie der Wissenschaften,
  Physikalisch-mathematische Klasse}, pages 144--153, 1931.

\bibitem{schulman2015trust}
J.~Schulman, S.~Levine, P.~Abbeel, M.~Jordan, and P.~Moritz.
\newblock Trust region policy optimization.
\newblock In {\em International conference on machine learning}, pages
  1889--1897. PMLR, 2015.

\bibitem{shi2023diffusion}
Y.~Shi, V.~De~Bortoli, A.~Campbell, and A.~Doucet.
\newblock Diffusion {S}chr{\"o}dinger bridge matching.
\newblock {\em Advances in Neural Information Processing Systems},
  36:62183--62223, 2023.

\bibitem{song2020score}
Y.~Song, J.~Sohl-Dickstein, D.~P. Kingma, A.~Kumar, S.~Ermon, and B.~Poole.
\newblock Score-based generative modeling through stochastic differential
  equations.
\newblock In {\em International Conference on Learning Representations}, 2021.

\bibitem{sun2024dynamical}
J.~Sun, J.~Berner, L.~Richter, M.~Zeinhofer, J.~M{\"u}ller, K.~Azizzadenesheli,
  and A.~Anandkumar.
\newblock Dynamical measure transport and neural {PDE} solvers for sampling.
\newblock {\em arXiv preprint arXiv:2407.07873}, 2024.

\bibitem{thalmeier2020adaptive}
D.~Thalmeier, H.~J. Kappen, S.~Totaro, and V.~G{\'o}mez.
\newblock Adaptive smoothing for path integral control.
\newblock {\em Journal of Machine Learning Research}, 21(191):1--37, 2020.

\bibitem{vargas2021machine}
F.~Vargas.
\newblock Machine-learning approaches for the empirical {S}chr{\"o}dinger
  bridge problem.
\newblock Technical report, University of Cambridge, Computer Laboratory, 2021.

\bibitem{vargas2023denoising}
F.~Vargas, W.~Grathwohl, and A.~Doucet.
\newblock Denoising diffusion samplers.
\newblock In {\em International Conference on Learning Representations}, 2023.

\bibitem{vargas2024transport}
F.~Vargas, S.~Padhy, D.~Blessing, and N.~N{\"u}sken.
\newblock Transport meets variational inference: Controlled {M}onte {C}arlo
  diffusions.
\newblock In {\em The Twelfth International Conference on Learning
  Representations}, 2024.

\bibitem{vargas2021solving}
F.~Vargas, P.~Thodoroff, A.~Lamacraft, and N.~Lawrence.
\newblock Solving {S}chr{\"o}dinger bridges via maximum likelihood.
\newblock {\em Entropy}, 23(9):1134, 2021.

\bibitem{wang2021deep}
G.~Wang, Y.~Jiao, Q.~Xu, Y.~Wang, and C.~Yang.
\newblock Deep generative learning via {S}chr{\"o}dinger bridge.
\newblock In {\em International conference on machine learning}, pages
  10794--10804. PMLR, 2021.

\bibitem{wu2020stochastic}
H.~Wu, J.~K{\"o}hler, and F.~No{\'e}.
\newblock Stochastic normalizing flows.
\newblock {\em Advances in neural information processing systems},
  33:5933--5944, 2020.

\bibitem{zhang2022path}
Q.~Zhang and Y.~Chen.
\newblock Path integral sampler: A stochastic control approach for sampling.
\newblock In {\em International Conference on Learning Representations}, 2022.

\end{thebibliography}

\newpage
\appendix
\crefalias{section}{appendix}\crefalias{subsection}{appendix}\crefalias{subsubsection}{appendix}

\addtocontents{toc}{\protect\setcounter{tocdepth}{2}}
\renewcommand{\contentsname}{Contents of the appendix}
\tableofcontents
\newpage

\section{Proofs and additional propositions}
\label{app: proofs}
In this section we provide the proofs for statements in the main part of the article and provide additional propositions.

\subsection{Notation}
\label{app: notation}
We denote with $\vec{\P}^u \in \mathcal{P}$ a path space measure related to a forward stochastic differential equation with control function $u \in \mathcal{U}$, where $\mathcal{P}$ is the set of probability measures on the space of continuous functions $C([0, T ], \R^d)$. Likewise, $\cev{\P}^v \in \mathcal{P}$ corresponds to a reverse-time SDE with control $v\in \mathcal{U}$. We denote by $\P_t$ the time marginal at time $t \in [0,T]$, by $\P_{0,T}$ the joint law of the endpoints $(X_0, X_T)$, and by $\P_{[t, T]}$ the time restriction to the interval $[t, T]$. The set $\mathcal{U} \subset C(\R^d \times [0, T], \R^d)$ denotes the space of admissible controls. With $\nabla$ we denote the gradient w.r.t. the spatial variable $x$. In the proofs we allow for a base drift $f$, i.e., $\vec{\P}^u$ is the law of $\mathrm d X_t = (f + \sigma u)(X_t, t)\,\mathrm d t + \sigma(t) \vec{\mathrm d} W_t$ and $\cev{\P}^v$ the law of $\mathrm d Y_t = (f - \sigma v)(Y_t, t)\,\mathrm d t + \sigma(t) \cev{\mathrm d} W_t$; the main text corresponds to $f = 0$. The diffusion coefficient $\sigma(t)$ is invertible and does not depend on the state. We omit function arguments, such as the time argument of $\sigma$, whenever they are clear from the context. $\P^{\mathrm{ref}} = \vec{\P}^{0}$ denotes the uncontrolled reference process and $\P^{\mathrm{ref}}_{t|0,T}$ its bridge, i.e., the law of $X_t$ given $(X_0, X_T)$. In conditional expectations given $X_t = x$, such as the cost-to-go $\mathcal{L}^{\mathrm{TRC}}_i(u, \lambda; x, t)$ and the value functions, the forward path measures $\vec{\P}^{u}_{[t, T]}$ and $\vec{\P}^{u_i}_{[t, T]}$ denote the laws of the forward SDE on $[t, T]$ with the common initial law $\vec{\P}^{u_i}_t$, so that $\frac{\mathrm d \vec{\P}^{u}_{[t, T]}}{\mathrm d \vec{\P}^{u_i}_{[t, T]}}$ has no boundary term at time $t$; we write $p_t \coloneqq \vec{\P}^{u_i}_t$.

\textbf{Expectations.} $\E_{\P}$ denotes the expectation of a path functional $X \mapsto G(X)$ under $X \sim \P$, and $\E_{\P}[\,\cdot \,|\, X_t = x]$ the conditional expectation given the state at time $t$. For endpoint-based objectives such as \eqref{eq: CE matching} and \eqref{eq: loss for v}, $\E_{\P_{0,T}\P^{\mathrm{ref}}_{t|0,T}}$ denotes sampling $(X_0, X_T) \sim \P_{0,T}$, $t \sim \mathcal{U}(0, T)$, and $X_t \sim \P^{\mathrm{ref}}_{t|0,T}(\cdot \,|\, X_0, X_T)$, with the time integral normalized accordingly. Unless stated otherwise, a Radon-Nikodym derivative $\frac{\mathrm d \P}{\mathrm d \mathbb{Q}}(X)$ is evaluated along the trajectory $X$ that is integrated in the surrounding expectation.

\subsection{Auxiliary results}
\label{app: auxiliary}
We collect the auxiliary identities used in the proofs below, stated in our notation with proofs or proof sketches. They correspond to results from \cite{karimi2023sinkhorn,richter2023improved,domingo2024stochastic,liu2025adjoint}.

\begin{lemma}[Terminal density ratio]
\label{lem: terminal density ratio}
Let $\P, \mathbb{Q} \in \mathcal{P}$ with $\mathbb{Q}_T \ll \P_T$ and $\mathbb{Q}(\cdot \,|\, X_T = y) = \P(\cdot \,|\, X_T = y)$ for $\P_T$-a.e.\ $y$. Then $\frac{\mathrm d \mathbb{Q}}{\mathrm d \P}(X) = \frac{\mathrm d \mathbb{Q}_T}{\mathrm d \P_T}(X_T)$. In particular, if $v_i$ is an exact backward IPF half-step for $u_i$, i.e., $\cev{\P}^{v_i} \in \argmin \{\KL(\mathbb{Q} | \vec{\P}^{u_i}) : \mathbb{Q}_T = p_\mathrm{target}\}$, then
\begin{equation}
\label{eq: terminal density ratio}
\frac{\mathrm d \cev{\P}^{v_i}}{\mathrm d \vec{\P}^{u_i}}(X) = \frac{p_\mathrm{target}}{\vec{\P}^{u_i}_T}(X_T).
\end{equation}
\end{lemma}
\begin{proof}
For bounded measurable $G$, disintegration with respect to $X_T$ gives $\E_{\mathbb{Q}}[G] = \int_{\R^d} \E_{\P}[G \,|\, X_T = y]\, \frac{\mathrm d \mathbb{Q}_T}{\mathrm d \P_T}(y)\, \P_T(\mathrm d y) = \E_{\P}\big[G\, \frac{\mathrm d \mathbb{Q}_T}{\mathrm d \P_T}(X_T)\big]$, which is the first claim. For the second claim, the chain rule of the KL divergence yields
$\KL(\mathbb{Q} | \vec{\P}^{u_i}) = \KL(\mathbb{Q}_T | \vec{\P}^{u_i}_T) + \E_{\mathbb{Q}_T}\big[\KL(\mathbb{Q}(\cdot | X_T) \,|\, \vec{\P}^{u_i}(\cdot | X_T))\big]$.
Under the constraint $\mathbb{Q}_T = p_\mathrm{target}$, the first term is fixed and the second term vanishes if and only if $\mathbb{Q}(\cdot | X_T) = \vec{\P}^{u_i}(\cdot | X_T)$. This minimizer is the law of the reverse-time SDE with the backward drift of the time reversal of $X^{u_i}$, started at $p_\mathrm{target}$, and is therefore of the form $\cev{\P}^{v}$. The first claim then gives \eqref{eq: terminal density ratio}. By \cite[Corollary E.2]{vargas2024transport}, the forward KL version of the half-step used in \Cref{sec: trust region SB} has the same minimizer.
\end{proof}

\begin{lemma}[Forward-backward path likelihood ratio, {\cite[Proposition~2.3 and Remark~A.1]{richter2023improved}}]
\label{lem: path likelihood ratio}
Let $X \sim \vec{\P}^u$ and $t \in [0, T]$. Then, $\vec{\P}^u$-almost surely,
\begin{align}
\label{eq: path RND general}
\begin{split}
\log \frac{\mathrm d \cev{\P}^{v}_{[t, T]}}{\mathrm d \vec{\P}^{u}_{[t, T]}}(X) = \log \frac{p_\mathrm{target}(X_T)}{\vec{\P}^u_t(X_t)} &- \int_t^T \sigma^{-1}(f + \sigma u) \cdot \sigma^{-1} \vec{\mathrm d} X_s + \frac{1}{2} \int_t^T \|\sigma^{-1}(f + \sigma u)\|^2 \mathrm d s \\
&+ \int_t^T \sigma^{-1}(f - \sigma v) \cdot \sigma^{-1} \cev{\mathrm d} X_s - \frac{1}{2} \int_t^T \|\sigma^{-1}(f - \sigma v)\|^2 \mathrm d s,
\end{split}
\end{align}
where the integrands are evaluated at $(X_s, s)$. For $f = 0$ this reads
\begin{align}
\label{eq: importance weights u_i v_i}
\begin{split}
\log \frac{\mathrm d \cev{\P}^{v}_{[t, T]}}{\mathrm d \vec{\P}^{u}_{[t, T]}}(X) = \log \frac{p_\mathrm{target}(X_T)}{\vec{\P}^u_t(X_t)} & + \frac{1}{2} \int_t^T \big(\|u\|^2 - \|v\|^2\big) \mathrm d s \\
& - \int_t^T u \cdot \sigma^{-1} \vec{\mathrm d} X_s - \int_t^T v \cdot \sigma^{-1} \cev{\mathrm d} X_s.
\end{split}
\end{align}
Moreover, writing $X = X^u$ with driving Brownian motion $W$, 
\begin{equation}
\label{eq: path RND ito}
\log \frac{\mathrm d \vec{\P}^{u}_{[t, T]}}{\mathrm d \cev{\P}^{v}_{[t, T]}}(X^u) = \log \frac{\vec{\P}^u_t(X^u_t)}{p_\mathrm{target}(X^u_T)} + \int_t^T \Big(\frac{1}{2}\|u + v\|^2 + \nabla \cdot (\sigma v - f)\Big) \mathrm d s + \int_t^T (u + v) \cdot \mathrm d W_s.
\end{equation}
\end{lemma}
\begin{proof}[Proof sketch]
Discretizing both processes with the Euler-Maruyama scheme, the ratio of the backward and forward Gaussian transition densities yields Riemann-It\^o sums with the forward drift at the left and the backward drift at the right endpoints, which converge to \eqref{eq: path RND general}; see \cite[Lemma~B.7]{berner2025discrete}. For \eqref{eq: path RND ito}, convert the backward integral into a forward one, $\int_t^T b \cdot \sigma^{-1} \cev{\mathrm d} X_s = \int_t^T b \cdot \sigma^{-1} \vec{\mathrm d} X_s + \int_t^T \nabla \cdot (\sigma b)\, \mathrm d s$ with $b = \sigma^{-1}(f - \sigma v)$, and insert $\sigma^{-1}\vec{\mathrm d} X^u_s = \sigma^{-1}(f + \sigma u)\, \mathrm d s + \mathrm d W_s$.
\end{proof}

\begin{lemma}[Conditional expectation matching]
\label{lem: matching}
Let $Y$ be a random vector, $Z$ a random variable, and $\zeta(Z) > 0$ a weight with $\E[\zeta(Z)\|Y\|^2] < \infty$. For every measurable $u$ with $\E[\zeta(Z)\|u(Z)\|^2] < \infty$,
\begin{equation}
\E\big[\zeta(Z)\|u(Z) - Y\|^2\big] = \E\big[\zeta(Z)\|u(Z) - \E[Y | Z]\|^2\big] + \E\big[\zeta(Z)\|Y - \E[Y | Z]\|^2\big].
\end{equation}
Hence both objectives differ by a constant independent of $u$, and both are minimized by $u(Z) = \E[Y | Z]$.
\end{lemma}
\begin{proof}
Expanding $u(Z) - Y = (u(Z) - \E[Y|Z]) + (\E[Y|Z] - Y)$, the cross term $2\E\big[\zeta(Z)(u(Z) - \E[Y|Z]) \cdot \E[\E[Y|Z] - Y \,|\, Z]\big]$ vanishes by the tower property.
\end{proof}

\begin{lemma}[Backward matching loss and the IPF half-step, cf.\ {\cite{liu2025adjoint,karimi2023sinkhorn}}]
\label{lem: backward matching}
Let $\P \coloneqq \vec{\P}^{u_{i+1}}$ be reciprocal (cf.\ \Cref{prop: preservation of reciprocal property}), i.e., $\P$ has the bridges $\P^{\mathrm{ref}}_{\cdot|0,T}$ and the endpoint density $\P_{0,T}(y, z) = \widehat{\varphi}_0(y)\, p^{\mathrm{ref}}_{T|0}(z|y)\, \varphi_T(z)$, and set $\widehat{\varphi}_t(x) \coloneqq \int_{\R^d} p^{\mathrm{ref}}_{t|0}(x|y)\, \widehat{\varphi}_0(y)\, \mathrm d y$. Then the minimizer of the backward loss \eqref{eq: loss for v} over all measurable $v$ is $v^*(x, t) = \sigma^\top \nabla \log \widehat{\varphi}_t(x)$. The reverse-time SDE \eqref{eq: backward SDE} with $v^*$ has the same backward transition kernels as the time reversal of $\P$; hence $\cev{\P}^{v^*}(\cdot | X_T) = \P(\cdot | X_T)$, i.e., $v^*$ is the exact backward IPF half-step of \Cref{lem: terminal density ratio}.
\end{lemma}
\begin{proof}[Proof sketch]
Since $\P$ is reciprocal, $\P_t = \widehat{\varphi}_t \varphi_t$ with $\varphi_t(x) \coloneqq \int_{\R^d} p^{\mathrm{ref}}_{T|t}(z|x)\, \varphi_T(z)\, \mathrm d z$, and the conditional law of $X_0$ given $X_t = x$ has density proportional to $\widehat{\varphi}_0(y)\, p^{\mathrm{ref}}_{t|0}(x|y)$. By \Cref{lem: matching}, the minimizer of \eqref{eq: loss for v} is $\sigma^\top \E_{\P}[\nabla_x \log p^{\mathrm{ref}}_{t|0}(x | X_0) \,|\, X_t = x] = \sigma^\top \nabla \log \widehat{\varphi}_t(x)$. By Nelson's relation, the time reversal of $\P$ has the backward drift $f - \sigma\sigma^\top \nabla \log \widehat{\varphi}_t = f - \sigma v^*$, so it shares the backward transition kernels and the conditional law given $X_T$ with $\cev{\P}^{v^*}$; cf.\ \cite{liu2025adjoint,karimi2023sinkhorn}.
\end{proof}

\begin{lemma}[One-step inequality of KL mirror descent]
\label{lem: mirror descent step}
Let $F(\P) \coloneqq \KL(\P_T | p_\mathrm{target})$ and $\mathcal{A} \coloneqq \{\P \in \mathcal{P} : \P_0 = p_\mathrm{prior}\}$. For $\P_i \in \mathcal{A}$ and $\gamma \in (0, 1]$, let $\P_{i+1} \in \mathcal{A}$ be defined by
\begin{equation}
\label{eq: md update}
\frac{\mathrm d \P_{i+1}}{\mathrm d \P_i}(X) = \frac{e^{-\gamma g(X)}}{\E_{\P_i}[e^{-\gamma g} \,|\, X_0]}, \qquad g(X) \coloneqq \log \frac{\P_{i,T}}{p_\mathrm{target}}(X_T).
\end{equation}
Then, for every $\mathbb{Q} \in \mathcal{A}$ for which the quantities below are finite,
\begin{equation}
\label{eq: md one step}
\gamma \big(F(\P_{i+1}) - F(\mathbb{Q})\big) \le \KL(\mathbb{Q} | \P_i) - \KL(\mathbb{Q} | \P_{i+1}).
\end{equation}
In particular, choosing $\mathbb{Q} = \P_i$ shows $F(\P_{i+1}) \le F(\P_i)$.
\end{lemma}
\begin{proof}
The function $g$ is the first variation $\delta F(\P_i)$ up to an additive constant; we write $\langle g, \P - \mathbb{Q} \rangle \coloneqq \E_{\P}[g] - \E_{\mathbb{Q}}[g]$. 
\emph{Three-point identity.} Let $Z(x) \coloneqq \E_{\P_i}[e^{-\gamma g} \,|\, X_0 = x]$. Since all measures in $\mathcal{A}$ have the initial marginal $p_\mathrm{prior}$, for every $\mathbb{Q} \in \mathcal{A}$ we have $\KL(\mathbb{Q} | \P_i) - \KL(\mathbb{Q} | \P_{i+1}) = \E_{\mathbb{Q}}\big[\log \frac{\mathrm d \P_{i+1}}{\mathrm d \P_i}\big] = -\gamma \E_{\mathbb{Q}}[g] - \E_{p_\mathrm{prior}}[\log Z]$. For $\mathbb{Q} = \P_{i+1}$, this gives $\KL(\P_{i+1} | \P_i) = -\gamma \E_{\P_{i+1}}[g] - \E_{p_\mathrm{prior}}[\log Z]$. Subtracting both identities yields
\begin{equation}
\gamma \langle g, \P_{i+1} - \mathbb{Q} \rangle = \KL(\mathbb{Q} | \P_i) - \KL(\mathbb{Q} | \P_{i+1}) - \KL(\P_{i+1} | \P_i).
\end{equation}
\emph{Relative smoothness and convexity.} For every $\P$ with $F(\P) < \infty$, a direct computation gives $F(\P) - F(\P_i) - \langle g, \P - \P_i \rangle = \KL(\P_T | \P_{i,T})$, which lies in $[0, \KL(\P | \P_i)]$ by the data processing inequality. Applying the upper bound to $\P = \P_{i+1}$ and the lower bound to $\P = \mathbb{Q}$ and subtracting, we obtain $F(\P_{i+1}) - F(\mathbb{Q}) \le \langle g, \P_{i+1} - \mathbb{Q} \rangle + \KL(\P_{i+1} | \P_i)$. Multiplying by $\gamma$ and inserting the three-point identity,
\begin{equation}
\gamma \big(F(\P_{i+1}) - F(\mathbb{Q})\big) \le \KL(\mathbb{Q} | \P_i) - \KL(\mathbb{Q} | \P_{i+1}) - (1 - \gamma)\KL(\P_{i+1} | \P_i),
\end{equation}
and the last term is non-positive since $\gamma \le 1$. For $\mathbb{Q} = \P_i$, the right-hand side of \eqref{eq: md one step} equals $-\KL(\P_i | \P_{i+1}) \le 0$.
\end{proof}

\subsection{Proofs of \Cref{prop: geometric interpretation of update,prop: optimality for TR-SB}}
\label{app: proofs of props}

\begin{proof}[Proof of \Cref{prop: geometric interpretation of update}]

First note that, for a fixed Lagrange multiplier $\lambda \ge 0$, it holds
\begin{equation}
    u^\lambda_{i+1} = \argmin_{u \in \U} \mathcal{L}^{\mathrm{TR}}_i(u, \lambda) = \argmin_{u \in \U} \left\{ \frac{1}{1+\lambda}\mathcal{L}^{\mathrm{TR}}_i(u, \lambda)\right\}.
\end{equation}
Then, we can compute
    \begin{subequations}
    \begin{align}
        u_{i+1}^{\lambda} &= \argmin_{u \in \U} \left\{ \frac{1}{1+\lambda} \KL(\vec{\P}^u |\cev{\P}^{v_{i}}) + \frac{\lambda}{1+\lambda} \KL(\vec{\P}^u|\vec{\P}^{u_i}) \right\} \\
        &= \argmin_{u \in \U}\mathbb{E}\left[\frac{1}{1+\lambda} \log \frac{\mathrm{d}\vec{\P}^u}{\mathrm{d}\cev{\P}^{v_{i}}}(X^u) + \frac{\lambda}{1+\lambda} \log \frac{\mathrm{d}\vec{\P}^u}{\mathrm{d}\vec{\P}^{u_{i}}}(X^u)  \right]  \\
        &= \argmin_{u \in \U} \mathbb{E}\left[ \log \left(\left(\frac{\mathrm{d}\vec{\P}^u}{\mathrm{d}\cev{\P}^{v_{i}}}(X^u)\right)^{\frac{1}{1+\lambda}} \left(\frac{\mathrm{d}\vec{\P}^u}{\mathrm{d}\vec{\P}^{u_{i}}}(X^u)\right)^{\frac{\lambda}{1+\lambda}}\right)  \right]  \\
        &= \argmin_{u \in \U} \mathbb{E}\left[ \log \frac{\mathrm{d}\vec{\P}^u}{\mathrm{d}\Big((\cev{\P}^{v_i})^{\frac{1}{1+\lambda}}(\vec{\P}^{u_{i}})^{\frac{\lambda}{1+\lambda}}\Big)}(X^u) \right] \\
        &= \argmin_{u \in \U} \KL\left(\vec{\P}^u \Big| (\cev{\P}^{v_i})^{\frac{1}{1+\lambda}}(\vec{\P}^{u_i})^{\frac{\lambda}{1+\lambda}}\right).
    \end{align}
    \end{subequations}
    Let $\mathbb{G}$ denote the normalized geometric average $\mathbb{G} \propto (\cev{\P}^{v_i})^{\frac{1}{1+\lambda}}(\vec{\P}^{u_{i}})^{\frac{\lambda}{1+\lambda}}$, i.e., $\frac{\mathrm d \mathbb{G}}{\mathrm d \vec{\P}^{u_i}} \propto \big(\frac{\mathrm d \cev{\P}^{v_i}}{\mathrm d \vec{\P}^{u_i}}\big)^{\frac{1}{1+\lambda}}$. The minimization is over path measures with $\vec{\P}^u_0 = p_\mathrm{prior}$. By the chain rule of the KL divergence with respect to $X_0$,
    \begin{equation}
    \KL(\vec{\P}^u | \mathbb{G}) = \KL(p_\mathrm{prior} | \mathbb{G}_0) + \E_{X_0 \sim p_\mathrm{prior}}\big[\KL(\vec{\P}^u(\cdot | X_0) \,|\, \mathbb{G}(\cdot | X_0))\big],
    \end{equation}
    where the first term does not depend on $u$. Hence the minimizer satisfies $\vec{\P}^{u_{i+1}^{\lambda}}(\cdot | X_0) = \mathbb{G}(\cdot | X_0)$, i.e., it is the geometric average with its initial marginal replaced by $p_\mathrm{prior}$. Since $\vec{\P}^{u_i}_0 = p_\mathrm{prior}$, this gives
    \begin{align}
    \label{eq: normalized annealing}
    \frac{\mathrm{d}\vec{\P}^{u^{\lambda}_{i+1}}}{\mathrm{d}\vec{\P}^{u_{i}}}(X) = \frac{\mathrm d \mathbb{G}(\cdot | X_0)}{\mathrm d \vec{\P}^{u_i}(\cdot | X_0)}(X) = \frac{\left(\frac{\mathrm d \cev{\P}^{v_i}}{\mathrm d \vec{\P}^{u_i}}(X)\right)^{\frac{1}{1+\lambda}}}{\E_{\vec{\P}^{u_i}}\left[\left(\frac{\mathrm d \cev{\P}^{v_i}}{\mathrm d \vec{\P}^{u_i}}\right)^{\frac{1}{1+\lambda}} \,\middle|\, X_0\right]},
    \end{align}
    which is the first identity of the statement together with its normalization. That this measure is induced by a control $u^\lambda_{i+1} \in \mathcal{U}$ follows from \Cref{prop: optimality for TR-SB}. The second equality follows from \Cref{lem: terminal density ratio}, which relates the Radon-Nikodym derivative of path measures to the likelihood ratio of their terminal marginals. This completes the proof.
\end{proof}

\begin{proof}[Proof of \Cref{prop: optimality for TR-SB}]
We remark that while the main text considers SDEs without base drift, here we establish the result for a more general case with drift $(f + \sigma u)$. The setting in the main text is recovered by setting $f=0$.

The proof consists of two parts. First, we establish a lower bound for the objective functional $\mathcal{L}^{\mathrm{TRC}}_i$ using Jensen's inequality. Second, we explicitly construct a control $u^* (= u^{\lambda}_{i+1})$ that attains this lower bound, thereby proving that the lower bound coincides with the value function defined as $V_{i+1}^\lambda(x, t) = \inf_{u} \mathcal{L}^{\mathrm{TRC}}_i(u, \lambda; x, t)$.
\paragraph{Part 1: Lower bound via Jensen's inequality.}
For any control $u \in \mathcal{U}$, we expand the Lagrangian defined in \cref{eq: J_TRC_pointwise}. Using the chain rule for Radon-Nikodym derivatives, we have:
\begin{subequations}
\begin{align}
    \mathcal{L}^{\mathrm{TRC}}_i(u, \lambda; x, t) 
    &= \mathbb{E}_{\vec{\P}^u}\left[ \log \left( \frac{\mathrm d \vec{\P}^{u_i}_{[t, T]}}{\mathrm d \cev{\P}^{v_i}_{[t, T]}} \frac{\mathrm d \vec{\P}^{u}_{[t, T]}}{\mathrm d \vec{\P}^{u_i}_{[t, T]}} \right) + \lambda \log \frac{\mathrm d \vec{\P}^{u}_{[t, T]}}{\mathrm d \vec{\P}^{u_i}_{[t, T]}} \,\Bigg|\, X_t^u = x \right] \\
    &= \mathbb{E}_{\vec{\P}^u}\left[ -(1+\lambda)\log \left(\frac{\mathrm d \vec{\P}^{u_i}_{[t, T]}}{\mathrm d \cev{\P}^{v_i}_{[t, T]}}\right)^{-\frac{1}{1+\lambda}} - (1+ \lambda) \log \frac{\mathrm d \vec{\P}^{u_i}_{[t, T]}}{\mathrm d \vec{\P}^{u}_{[t, T]}} \,\Bigg|\, X_t^u = x \right].
\end{align}
\end{subequations}
Applying Jensen's inequality to the strictly convex function $\phi(y) = -\log y$, we obtain the following lower bound:
\begin{subequations}
\begin{align}
\label{eq: jensen_lower_bound}
    \mathcal{L}^{\mathrm{TRC}}_i(u, \lambda; x, t) 
    &\ge -(1+\lambda)\log \mathbb{E}_{\vec{\P}^u}\left[ \left(\frac{\mathrm d \vec{\P}^{u_i}_{[t, T]}}{\mathrm d \cev{\P}^{v_i}_{[t, T]}}\right)^{-\frac{1}{1+\lambda}} \frac{\mathrm d \vec{\P}^{u_i}_{[t, T]}}{\mathrm d \vec{\P}^{u}_{[t, T]}} \,\Bigg|\, X_t^u = x \right] \\
    &= -(1+\lambda)\log \mathbb{E}_{\vec{\P}^{u_i}}\left[ \left(\frac{\mathrm d \vec{\P}^{u_i}_{[t, T]}}{\mathrm d \cev{\P}^{v_i}_{[t, T]}}\right)^{-\frac{1}{1+\lambda}} \,\Bigg|\, X_t^{u_i} = x \right].
\end{align}
\end{subequations}
Note that the right-hand side is independent of $u$. Thus, $V_{i+1}^\lambda(x, t)$ is bounded from below by this quantity.

\paragraph{Part 2: Attainability via martingale comparison.}
We now construct the optimal control $u^*$ that minimizes the objective. The lower bound in \cref{eq: jensen_lower_bound} is attained if and only if Jensen's inequality becomes an equality. This requires the random variable inside the logarithm to be constant almost surely with respect to the optimal path measure $\vec{\P}^{u^*}$. 

Let us denote this state-dependent constant by $C(x, t)$. By definition, at the optimum, the value of the objective function becomes:
\begin{equation}
    \inf_{u \in \mathcal{U}}\mathcal{L}^{\mathrm{TRC}}_i(u, \lambda; x, t)  = -(1+\lambda) \log \E_{\vec{\P}^{u^*}}[C(x, t)] = -(1+\lambda) \log C(x, t).
\end{equation}
Since the value function $V_{i+1}^\lambda(x, t)$ is defined as this minimal cost, we establish the fundamental relationship:
\begin{equation}
    V_{i+1}^\lambda(x, t) = -(1+\lambda) \log C(x, t) \quad \iff \quad \log C(x, t) = -\frac{V_{i+1}^\lambda(x, t)}{1+\lambda}.
\end{equation}

Now, writing the condition ``inside of log equals $C$'' explicitly:
\begin{equation}
\label{eq: optimality_log_relation}
     \left(\frac{\mathrm d \vec{\P}^{u_i}_{[t, T]}}{\mathrm d \cev{\P}^{v_i}_{[t, T]}}\right)^{-\frac{1}{1+\lambda}} \frac{\mathrm d \vec{\P}^{u_i}_{[t, T]}}{\mathrm d \vec{\P}^{u^*}_{[t, T]}} = C(x, t).
\end{equation}
Taking the logarithm and rearranging terms:
\begin{equation}
    \log \frac{\mathrm d \vec{\P}^{u^*}_{[t, T]}}{\mathrm d \vec{\P}^{u_i}_{[t, T]}} = -\frac{1}{1+\lambda} \log \frac{\mathrm d \vec{\P}^{u_i}_{[t, T]}}{\mathrm d \cev{\P}^{v_i}_{[t, T]}} - \log C(x, t).
\end{equation}

We identify $u^*$ by matching the martingale parts (coefficients of the Brownian motion $\mathrm d W_s$) on both sides of \eqref{eq: optimality_log_relation}, which holds for every lower limit $s \in [t, T]$ in place of $t$ (with $x$ replaced by $X_s$), so that both sides are processes in $s$.
For the first term on the RHS, we recall the explicit representation of the log-likelihood ratio from \Cref{lem: path likelihood ratio}, see \eqref{eq: path RND ito}:
\begin{equation}
\label{eq: richter_formula}
   \log \frac{\mathrm d \vec{\P}^{u_i}_{[t, T]}}{\mathrm d \cev{\P}^{v_i}_{[t, T]}} =  \log \frac{p_t(X_t^{u_i})}{p_\mathrm{target}(X^{u_i}_T)} + \int_t^T \left( \frac{1}{2} \|u_i+v_i\|^2 + \nabla \cdot (\sigma v_i - f) \right)\mathrm ds + \int_t^T (u_i + v_i) \cdot \mathrm dW_s,
\end{equation}
where $p_s \coloneqq \vec{\P}^{u_i}_s$. As a process in the lower limit $s$, the stochastic integral contributes the martingale part $-(u_i + v_i) \cdot \mathrm dW_s$, and the boundary term $\log p_s(X_s)$ contributes $\sigma^\top \nabla \log p_s \cdot \mathrm dW_s$.
Similarly, by the Girsanov theorem, the LHS is given by:
\begin{equation}
    \log \frac{\mathrm d \vec{\P}^{u^*}_{[t, T]}}{\mathrm d \vec{\P}^{u_i}_{[t, T]}} = \frac{1}{2} \int_t^T \|u^* - u_i \|^2 \mathrm ds +  \int_t^T (u^* - u_i)\cdot \mathrm d W_s.
\end{equation}
For the normalizing constant term $\log C(x, t)$, applying It\^o's formula to $V_{i+1}^\lambda(X_s, s)$ under the dynamics $\mathrm d X_s = (f + \sigma u^*)\mathrm d s + \sigma \mathrm d W_s$:
\begin{align}
    \mathrm d (\log C) &= -\frac{1}{1+\lambda} \Bigg[ \underbrace{\left( \partial_t V_{i+1}^\lambda + (f + \sigma u^*) \cdot \nabla V_{i+1}^\lambda + \frac{1}{2}\operatorname{Tr}(\sigma \sigma^\top \nabla^2 V_{i+1}^\lambda) \right)}_{\text{drift part}} \mathrm d s \nonumber \\
    &\qquad + \underbrace{\nabla V_{i+1}^\lambda \cdot \sigma \mathrm d W_s}_{\text{martingale part}} \Bigg].
\end{align}

Equating the coefficients of the martingale part ($\mathrm{d}W_s$) across the equation yields:
\begin{equation}
    (u^* - u_i) = -\frac{1}{1+\lambda} (u_i + v_i) - \frac{1}{1+\lambda} \sigma^\top \nabla \big(V_{i+1}^\lambda - \log p_t\big).
\end{equation}
Solving for $u^*$, we obtain the optimal control law:
\begin{equation}
    u^* = \frac{\lambda}{1+\lambda} u_i - \frac{1}{1+\lambda} v_i - \frac{1}{1+\lambda} \sigma^\top \nabla \big(V_{i+1}^\lambda - \log p_t\big).
\end{equation}
By construction, this control $u^*$ eliminates the martingale part of the log-likelihood ratio in \eqref{eq: optimality_log_relation}, rendering the term inside the expectation in \eqref{eq: jensen_lower_bound} deterministic (equal to $C(x,t)$).
Consequently, we have
\begin{equation}
    V^{\lambda}_{i+1}(x, t) = -(1+\lambda)\log \mathbb{E}_{\vec{\P}^{u_i}}\left[ \left(\frac{\mathrm d \vec{\P}^{u_i}_{[t, T]}}{\mathrm d \cev{\P}^{v_i}_{[t, T]}}\right)^{-\frac{1}{1+\lambda}} \,\Bigg|\, X_t^{u_i} = x \right]
\end{equation}
\end{proof}

\begin{remark}[Relation to \cite{karimi2023sinkhorn} and the HJB equation]
The explicit formulas derived in \Cref{prop: optimality for TR-SB} are equivalent to Theorem~4.2 of \cite{karimi2023sinkhorn} when $v_i$ is the exact backward half-step.
The main difference lies in the derivation: whereas our proof is probabilistic, relying on Jensen's inequality and martingale matching, \cite{karimi2023sinkhorn} obtain the same result analytically via the associated Hamilton-Jacobi-Bellman (HJB) equation.
For completeness, we present below an alternative proof in our notation following this approach.

More precisely, their method applies a logarithmic (Fleming or Cole-Hopf) transformation \cite[Proposition~C.1]{karimi2023sinkhorn}, which in our notation reads
$V_{i+1}^\lambda = -(1+\lambda)\log \Psi$,
to linearize the nonlinear HJB equation associated with $\mathcal{L}^{\mathrm{TRC}}_i$.
The resulting function $\Psi$ admits a Feynman-Kac representation, which, after a Girsanov change of measure, coincides with our expectation formula in \Cref{eq: value function lambda}.
This connection is also implicitly reflected in our probabilistic proof: matching the \emph{drift terms} ($\mathrm ds$ terms) in \eqref{eq: optimality_log_relation}, rather than the martingale terms used to identify $u^*$, recovers precisely the same HJB equation.
\end{remark}

\begin{proof}[Alternate Proof of \Cref{prop: optimality for TR-SB} via the HJB equation.]
\label{proof: alternate proof of optimality thm}
We provide an alternative proof of \Cref{prop: optimality for TR-SB} based on the HJB equation, following the approach of \cite{karimi2023sinkhorn} but expressed in our notation.

The proof proceeds in three main steps. First, we decompose the objective functional to isolate the dynamic component of the control problem. Second, we derive the associated nonlinear HJB equation and linearize it using a logarithmic transformation. Finally, we solve the resulting linear PDE via the Feynman-Kac formula and apply a change of measure to recover the variational representation~\cite{hartmann2017variational}.

\paragraph{Part 1: Derive the SOC formulation by using Girsanov theorem}
We first derive the SOC formulation of the objective functional:
\begin{equation}
    \mathcal{L}^{\mathrm{TRC}}_i(u, \lambda; x, t) 
        = \mathbb{E}_{\vec{\P}^u}\left[ \log \frac{\mathrm d \vec{\P}^{u}_{[t, T]}}{\mathrm d \cev{\P}^{v_i}_{[t, T]}}  + \lambda \log \frac{\mathrm d \vec{\P}^{u}_{[t, T]}}{\mathrm d \vec{\P}^{u_i}_{[t, T]}} \,\Bigg|\, X_t^u = x \right].
\end{equation}
We expand the terms using \Cref{lem: path likelihood ratio} and the standard Girsanov theorem, substituting the following expressions:
\begin{gather}
    \label{eq: richter_formula_again}
       \log \frac{\mathrm d \vec{\P}^{u_i}_{[t, T]}}{\mathrm d \cev{\P}^{v_i}_{[t, T]}}=  \log \frac{p_t(X_t^{u_i})}{p_\mathrm{target}(X_T^{u_i})}+\int_t^T \left( \frac{1}{2} \|u_i+v_i\|^2 + \nabla \cdot (\sigma v_i - f) \right)\mathrm ds + \int_t^T (u_i + v_i) \cdot \mathrm dW_s,\\
       \log \frac{\mathrm d \vec{\P}^{u}_{[t, T]}}{\mathrm d \vec{\P}^{u_i}_{[t, T]}} = \frac{1}{2} \int_t^T \|u - u_i \|^2 \mathrm ds +  \int_t^T (u - u_i) \cdot \mathrm d W_s.
\end{gather}
Substituting these into $\mathcal{L}^{\mathrm{TRC}}_i$ and collecting the terms, we obtain:
\begin{align}
    \begin{split}
    &\mathcal{L}^{\mathrm{TRC}}_i(u, \lambda; x, t) \\
    &= \mathbb{E}_{\vec{\P}^u}\Bigg[ \tfrac{1+\lambda}{2}\int^{T}_t\left(\left\|u + \left(\tfrac{1}{1+\lambda} v_{i} - \tfrac{\lambda}{1+\lambda}u_{i}\right)\right\|^2 + \tfrac{\lambda}{(1+\lambda)^2}\| v_{i} + u_i\|^2 + \tfrac{2}{1+\lambda} \nabla \cdot (\sigma v_i-f)\right)\mathrm{d}s \\
    & \qquad\qquad + \log \frac{p_t(X_t^{u})}{p_\mathrm{target}(X_T^u)}\,\Bigg|\, X_t^u = x \Bigg].
    \end{split}
\end{align}
Since $\log p_t(X_t^u)$ is constant conditioned on $X_t^u=x$, we can separate it from the optimization objective. We define the dynamic cost functional $\widetilde{J}(u; x, t)$ and the potential term $h(x, s)$ as follows:
\begin{align}
\mathcal{L}^{\mathrm{TRC}}_i(u, \lambda; x, t)
&= \log p_t(x) + \widetilde{J}(u; x, t), \\
\widetilde{J}(u; x, t)
&\coloneqq
\mathbb{E}_{\vec{\mathbb{P}}^u}\Bigg[
\int_t^T
\tfrac{1+\lambda}{2}
\left(
\left\|
u + \left(
\tfrac{1}{1+\lambda} v_i
- \tfrac{\lambda}{1+\lambda} u_i
\right)
\right\|^2
+ h(X_s^u, s)
\right)
\,\mathrm{d}s  \notag \\
&\qquad\qquad
- \log p_{\mathrm{target}}(X_T^u)
\,\Bigg|\, X_t^u = x
\Bigg], \\
h(x, s)
&\coloneqq
\frac{\lambda}{(1+\lambda)^2}
\| v_i + u_i \|^2
+
\frac{2}{1+\lambda}
\nabla \cdot (\sigma v_i - f).
\end{align}

\paragraph{Part 2: The nonlinear PDE for the value function}
We denote the value function for the dynamic part of the cost as $\widetilde{V}(x, t) \coloneqq \inf_u \widetilde{J}(u; x, t)$. Based on the running cost derived in Part 1, $\widetilde{V}$ must satisfy the Hamilton-Jacobi-Bellman (HJB) equation:
\begin{equation}
    \pderiv{\widetilde{V}}{t} + \min_u \left\{ (f + \sigma u) \cdot \nabla \widetilde{V} + \frac{1}{2} \operatorname{Tr}(\sigma \sigma^\top \nabla^2 \widetilde{V}) + \tfrac{1+\lambda}{2}\left(\left\|u - \left(\tfrac{\lambda}{1+\lambda}u_{i} - \tfrac{1}{1+\lambda}v_{i}\right)\right\|^2 + h \right)  \right\} = 0,
\end{equation}
with the terminal condition $\widetilde{V}(x, T) = - \log p_\mathrm{target}(x)$.

The minimizer $u^*$ of the expression inside the curly braces can be found by taking the gradient with respect to $u$ and setting it to zero. The relevant terms for minimization are $\sigma u \cdot \nabla \widetilde{V} + \frac{1+\lambda}{2}\|u - (\tfrac{\lambda}{1+\lambda}u_{i} - \tfrac{1}{1+\lambda}v_{i})\|^2$. The first-order condition yields:
\begin{equation}
    \sigma^\top \nabla \widetilde{V} + (1+\lambda)\left(u^* - \left(\tfrac{\lambda}{1+\lambda}u_{i} - \tfrac{1}{1+\lambda}v_{i}\right)\right) = 0.
\end{equation}
Solving for $u^*$, we obtain the optimal control:
\begin{equation}
    u^* = \left(\tfrac{\lambda}{1+\lambda}u_{i} - \tfrac{1}{1+\lambda}v_{i}\right) - \frac{1}{1+\lambda}\sigma^\top \nabla \widetilde{V}.
\end{equation}
Substituting this optimal control $u^*$ back into the HJB equation, the drift term and the quadratic running cost term combine as follows:
\begin{align}
    &\text{Drift term: } \quad (f + \sigma u^*) \cdot \nabla \widetilde{V} = \left(f + \sigma\left(\tfrac{\lambda}{1+\lambda}u_{i} - \tfrac{1}{1+\lambda}v_{i}\right)\right) \cdot \nabla \widetilde{V} - \frac{1}{1+\lambda} \|\sigma^\top \nabla \widetilde{V}\|^2, \\
    &\text{Quad term: } \quad \frac{1+\lambda}{2}\left\| -\frac{1}{1+\lambda}\sigma^\top \nabla \widetilde{V} \right\|^2 = \frac{1}{2(1+\lambda)} \|\sigma^\top \nabla \widetilde{V}\|^2.
\end{align}
Summing these simplifies the nonlinear gradient term to $-\frac{1}{2(1+\lambda)} \|\sigma^\top \nabla \widetilde{V}\|^2$. Thus, we arrive at the following nonlinear PDE for $\widetilde{V}$:
\begin{equation} \label{eq:hjb_nonlinear}
    \pderiv{\widetilde{V}}{t} + \left(f + \sigma\left(\tfrac{\lambda}{1+\lambda}u_{i} - \tfrac{1}{1+\lambda}v_{i}\right)\right) \cdot \nabla \widetilde{V} + \frac{1}{2} \operatorname{Tr}(\sigma \sigma^\top \nabla^2 \widetilde{V}) - \frac{1}{2(1+\lambda)}\|\sigma^\top \nabla \widetilde{V}\|^2 + \frac{1+\lambda}{2} h = 0.
\end{equation}

\paragraph{Part 3: Linearization and solution via the Feynman-Kac formula}
To solve the nonlinear PDE \eqref{eq:hjb_nonlinear}, we apply the Cole-Hopf transformation with a scaling factor matching the nonlinearity. Define a new function $E(x, t)$ such that:
\begin{equation}
    \widetilde{V}(x, t) = -(1+\lambda) \log E(x, t).
\end{equation}
The derivatives transform as follows:
\begin{equation}
    \pderiv{\widetilde{V}}{t} = -(1+\lambda)\frac{1}{E}\pderiv{E}{t}, \quad \nabla \widetilde{V} = -(1+\lambda)\frac{\nabla E}{E}, \quad \nabla^2 \widetilde{V} = -(1+\lambda)\frac{\nabla^2 E}{E} + (1+\lambda)\frac{\nabla E (\nabla E)^\top}{E^2}.
\end{equation}
Substituting these into \cref{eq:hjb_nonlinear}, the nonlinear gradient terms cancel out perfectly:
\begin{align}
    &\text{Hessian part: } \frac{1}{2} \operatorname{Tr}\left(\sigma \sigma^\top (1+\lambda)\frac{\nabla E (\nabla E)^\top}{E^2}\right) = \frac{1+\lambda}{2} \frac{\|\sigma^\top \nabla E\|^2}{E^2}, \\
    &\text{Nonlinear part: } -\frac{1}{2(1+\lambda)} \left\| \sigma^\top \left( -(1+\lambda)\frac{\nabla E}{E} \right) \right\|^2 = -\frac{1+\lambda}{2} \frac{\|\sigma^\top \nabla E\|^2}{E^2}.
\end{align}
Thus, multiplying the remaining terms by $-\frac{E}{1+\lambda}$, we obtain the linear PDE for $E(x, t)$:
\begin{equation} \label{eq:pde_linear}
    \pderiv{E}{t} + \left( f + \sigma\left(\tfrac{\lambda}{1+\lambda}u_{i} - \tfrac{1}{1+\lambda}v_{i}\right) \right) \cdot \nabla E + \frac{1}{2}\operatorname{Tr}(\sigma \sigma^\top \nabla^2 E) - \frac{1}{2} h E = 0.
\end{equation}
The terminal condition transforms to $E(x, T) = \exp\left(-\frac{\widetilde{V}(x,T)}{1+\lambda}\right) = (p_\mathrm{target}(x))^{\frac{1}{1+\lambda}}$.

This is a backward Kolmogorov equation with a potential term. Let $\widetilde{u}_i \coloneqq \tfrac{\lambda}{1+\lambda}u_{i} - \tfrac{1}{1+\lambda}v_{i}$. By the Feynman-Kac formula, the solution is given by:
\begin{equation}
    E(x, t) = \mathbb{E}_{\vec{\P}^{\widetilde{u}_i}} \left[ \exp\left( - \frac{1}{2}\int_t^T h(X_s, s) \mathrm{d}s \right) (p_\mathrm{target}(X_T))^{\frac{1}{1+\lambda}} \,\Bigg|\, X_t = x \right].
\end{equation}

\paragraph{Part 4: Explicit calculation of the expectation}
We now change the measure in the expectation from $\vec{\P}^{\widetilde{u}_i}$ to $\vec{\P}^{u_i}$. The drift difference is $\widetilde{u}_i - u_i = -\frac{1}{1+\lambda}(u_i + v_i)$. The Girsanov density is:
\begin{equation}
    \frac{\mathrm{d}\vec{\P}^{\widetilde{u}_i}}{\mathrm{d}\vec{\P}^{u_i}} = \exp\left( -\frac{1}{1+\lambda}\int_t^T (u_i + v_i) \cdot \mathrm{d}W_s^{u_i} - \frac{1}{2(1+\lambda)^2}\int_t^T \|u_i + v_i\|^2 \mathrm{d}s \right).
\end{equation}
Substituting this into the Feynman-Kac formula for $E(x, t)$, we combine the exponents:
\begin{align}
E(x,t)
&=
\mathbb{E}_{\vec{\P}^{u_i}}
\bigg[
\exp\bigg(
-\tfrac{1}{1+\lambda}
\Big\{
\int_t^T
\Big(
\tfrac{1}{2}\|u_i+v_i\|^2
+
\nabla\cdot(\sigma v_i-f)
\Big)\mathrm{d}s
\nonumber\\
&\qquad\qquad
+
\int_t^T (u_i+v_i)\cdot \mathrm{d}W_s^{u_i}
-
\log p_\mathrm{target}(X_T)
\Big\}
\bigg)
\,\bigg|\, X_t=x
\bigg].
\end{align}
The term inside the brackets is exactly the log-likelihood ratio \eqref{eq: richter_formula_again}, excluding the initial term $\log p_t(x)$. Thus, substituting $E(x, t)$ back into
\begin{subequations}
\begin{align}
    V_{i+1}^\lambda(x, t) &\coloneqq \inf_{u \in \mathcal{U}} \mathcal{L}^{\mathrm{TRC}}_i(u, \lambda; x, t) =  \log p_t(x) + \inf_{u \in \mathcal{U}} \widetilde{J}(u; x, t) =\log p_t(x)  + \widetilde{V}(x, t) \\
    &= \log p_t(x) - (1+\lambda)\log E(x, t),
\end{align}
\end{subequations}
the $\log p_t(x)$ terms cancel out, yielding the desired result:
\begin{align}
    V_{i+1}^\lambda(x, t)=-(1+\lambda)\log \mathbb{E}_{\vec{\P}^{u_i}}\left[ \left(\frac{\mathrm d \vec{\P}^{u_i}_{[t, T]}}{\mathrm d \cev{\P}^{v_i}_{[t, T]}}\right)^{-\frac{1}{1+\lambda}} \,\Bigg|\, X_t^{u_i} = x \right].
\end{align}
Moreover, since $\widetilde{V} = V_{i+1}^\lambda - \log p_t$, the optimal control obtained in Part 2 reads $u^* = \frac{\lambda}{1+\lambda}u_i - \frac{1}{1+\lambda}v_i - \frac{1}{1+\lambda}\sigma^\top \nabla \big(V_{i+1}^\lambda - \log p_t\big)$, as stated in \Cref{prop: optimality for TR-SB}.
\end{proof}

\subsection{Stochastic optimal control formulation for SB based on the reciprocal property}

\begin{proposition}[Preservation of the reciprocal property]
\label{prop: preservation of reciprocal property}
Assume that the path measure at iteration $i$, denoted by $\vec{\P}^{u_i}$, is a reciprocal process with respect to the reference measure $\P^{\mathrm{ref}}$. Specifically, assume it admits the factorization:
\begin{align}
    \frac{\mathrm{d}\vec{\P}^{u_i}}{\mathrm{d}\P^{\mathrm{ref}}}(X) = \exp\left( \log \widehat{\varphi}_{i-1}(X_0) + \log \varphi_i(X_T) \right) = \widehat{\varphi}_{i-1}(X_0) \varphi_i(X_T).
\end{align}
If $v_i$ is the exact backward half-step for $u_i$ (\Cref{lem: terminal density ratio}), the updated measure $\vec{\P}^{u_{i+1}}$ obtained via the update rule \eqref{eq: gamma IPF} remains a reciprocal process.
\end{proposition}

\begin{proof}
By Proposition \ref{prop: geometric interpretation of update}, the Radon-Nikodym derivative of the updated measure is given by the multiplicative update:
\begin{align}
    \frac{\mathrm{d}\vec{\P}^{u_{i+1}}}{\mathrm{d}\vec{\P}^{u_{i}}}(X) 
    = \frac{\Psi_i(X_T)}{c_i(X_0)}, \qquad \Psi_i(X_T) \coloneqq \left( \frac{p_{\mathrm{target}}(X_T)}{\vec{\P}^{u_i}_T(X_T)} \right)^{\frac{1}{1+\lambda_i}}.
\end{align}
Note that $\Psi_i$ depends solely on the terminal state $X_T$, and the normalization $c_i(X_0) \coloneqq \E_{\vec{\P}^{u_i}}[\Psi_i(X_T) \,|\, X_0]$, which ensures $\vec{\P}^{u_{i+1}}_0 = p_{\mathrm{prior}}$, solely on the initial state $X_0$.
Multiplying this density by the hypothesis for step $i$, we obtain:
\begin{subequations}
\begin{align}
    \frac{\mathrm{d}\vec{\P}^{u_{i+1}}}{\mathrm{d}\P^{\mathrm{ref}}}(X) 
    &= \frac{\mathrm{d}\vec{\P}^{u_{i+1}}}{\mathrm{d}\vec{\P}^{u_{i}}}(X)  \frac{\mathrm{d}\vec{\P}^{u_{i}}}{\mathrm{d}\P^{\mathrm{ref}}}(X) \\
    &= \frac{\Psi_i(X_T)}{c_i(X_0)}  \left[ \widehat{\varphi}_{i-1}(X_0) \varphi_i(X_T) \right] \\
    &= \frac{\widehat{\varphi}_{i-1}(X_0)}{c_i(X_0)}  \left[ \varphi_i(X_T) \Psi_i(X_T) \right].
\end{align}
\end{subequations}
By defining the updated potentials as $\widehat{\varphi}_{i} \coloneqq \widehat{\varphi}_{i-1} / c_i$ and $\varphi_{i+1} := \varphi_i \cdot \Psi_i$, the measure $\vec{\P}^{u_{i+1}}$ retains the form of a reciprocal process in the same class as $\P^{\mathrm{ref}}$.
\end{proof}

\begin{remark}
In our algorithm, the controls are initialized as $u_0 = v_0 = 0$, which implies $\vec{\P}^{u_0} = \P^{\mathrm{ref}}$. Since the reference measure is trivially reciprocal (with $\widehat{\varphi}_{-1} = \varphi_0 \equiv 1$), it follows by induction that $\vec{\P}^{u_i}$ is a reciprocal process for all $i \ge 0$. For $i = 0$, $v_0 = 0$ is in general not the backward half-step, but for $f = 0$, \Cref{lem: path likelihood ratio} gives $\frac{\mathrm d \cev{\P}^{v_0}}{\mathrm d \vec{\P}^{u_0}}(X) = \frac{p_{\mathrm{target}}(X_T)}{p_{\mathrm{prior}}(X_0)}$, so that the proof applies with $\Psi_0 = p_{\mathrm{target}}^{\gamma_0}$ up to a function of $X_0$.
Crucially, Proposition \ref{prop: preservation of reciprocal property} holds for any value of the dual variable $\lambda_i$. Therefore, the preservation of the reciprocal structure is guaranteed even in our Trust Region IPF scheme where the step size $\gamma_i = \frac{1}{1+\lambda_i}$ is adaptively determined at each iteration $i$.
Finally, we note that this result constitutes the continuous path-measure counterpart of Lemma 2 in \citep{karimi2023sinkhorn}.
\end{remark}

\begin{proposition}[Equivalence of Reciprocal and Trust Region SOC]
\label{prop:reciprocal_soc_equivalence}
Consider the optimization at iteration $i$.
First, assume that the previous forward control $u_i$ is a \textit{reciprocal control} associated with a potential $\varphi_i$.
Second, let the backward control $v_i$ be fixed, corresponding to the backward potential $\widehat{\varphi}_i$ via $v_i = \sigma^{\top} \nabla \log \widehat{\varphi}_i$.

We restrict the search for the updated forward control $u$ to the class of controls of reciprocal form, i.e., $h$-transforms of the reference process with a terminal function,
\begin{align}
    \mathcal{U}_{\mathrm{REC}} \coloneqq \left\{ u = \sigma^{\top} \nabla \log \varphi_t \,\middle|\, \varphi \colon \R^d \to (0, \infty), \ \varphi_t(x) \coloneqq \E_{\vec{\P}^{\mathrm{ref}}}[\varphi(X_T) \,|\, X_t = x] \right\},
\end{align}
so that $\frac{\mathrm{d}\vec{\P}^{u}}{\mathrm{d}\vec{\P}^{\mathrm{ref}}} = a(X_0)\, \varphi(X_T)$ for a function $a$ of $X_0$ alone. The potential $\varphi$ is determined by $u$ up to a positive factor.
For any $u \in \mathcal{U}_{\mathrm{REC}}$, let $\varphi$ denote its associated forward potential. The Reciprocal SOC objective function is defined as:
\begin{align}
    \label{eq:recip_form}
    \mathcal{L}^{\mathrm{REC}}_i(u, \lambda; x, t) \coloneqq \mathbb{E}_{\vec{\P}^u}\left[ \frac{1}{2}\int^{T}_t \| u(X^u_s) \|^2 \mathrm{d}s + \Psi_i(X^u_T, u) \,\bigg|\, X^u_t=x \right],
\end{align}
where the terminal cost $\Psi_i$ utilizes the potentials of $u$ and $u_i$:
\begin{align}
    \Psi_i(y, u) \coloneqq \log \widehat{\varphi}_i(y) + \mathcal{E}(y) + \lambda \log \frac{\varphi(y)}{\varphi_i(y)}.
\end{align}
Then, for any $u \in \mathcal{U}_{\mathrm{REC}}$, the following identity holds:
\begin{align}
    \mathcal{L}^{\mathrm{TRC}}_i(u, \lambda; x, t) = \mathcal{L}^{\mathrm{REC}}_i(u, \lambda; x, t) - \lambda \log \varphi_t(x) + C(x, t),
\end{align}
where $C(x, t)$ is independent of $u$. In particular, if $\varphi$ is normalized such that $\varphi_t(x) = 1$, which leaves $u$ unchanged, then $\mathcal{L}^{\mathrm{TRC}}_i$ and $\mathcal{L}^{\mathrm{REC}}_i$ differ by a constant.
\end{proposition}

\begin{proof}
First, we justify the ansatz. According to Proposition \ref{prop: preservation of reciprocal property}, if the current iterate $\vec{\P}^{u_i}$ is a reciprocal process, then the optimal solution for the trust region control (TRC) problem is also a reciprocal process. Therefore, it is theoretically justified to restrict the search space to the reciprocal class. 
For $u \in \mathcal{U}_{\mathrm{REC}}$, conditionally on $X_t = x$, the density of $\vec{\P}^u_{[t,T]}$ with respect to the reference process is $\varphi(X_T)/\varphi_t(x)$ (\Cref{app: notation}), and $\E_{\vec{\P}^u}[\frac{1}{2}\int_t^T \|u\|^2 \mathrm ds \,|\, X_t = x] = \E_{\vec{\P}^u}[\log \varphi(X_T) \,|\, X_t = x] - \log \varphi_t(x)$ by the Girsanov theorem.

Recall the definition of the TRC loss function:
\begin{equation}
    \label{eq: TRC-loss}
    \mathcal{L}^{\mathrm{TRC}}_i(u, \lambda; x, t) 
        = \mathbb{E}_{\vec{\P}^u}\left[ \log \frac{\mathrm d \vec{\P}^{u}_{[t, T]}}{\mathrm d \cev{\P}^{v_i}_{[t, T]}}  + \lambda \log \frac{\mathrm d \vec{\P}^{u}_{[t, T]}}{\mathrm d \vec{\P}^{u_i}_{[t, T]}} \,\Bigg|\, X_t^u = x \right].
\end{equation}

For the first term, we utilize the identity relating the KL divergence to the potentials and control cost \citep[Eq.~(59)]{liu2025adjoint}. Note that the terminal density of the backward process $\cev{\P}^{v_i}$ matches the target, i.e., $q_T = p_{\mathrm{target}} = \mathcal{Z}^{-1} e^{-\mathcal{E}}$. Thus, we obtain:
\begin{align}
\label{eq:reciprocal_term1}
\begin{split}
&\mathbb{E}_{\vec{\P}^{u}} \left[ \log \frac{\mathrm d \vec{\P}^{u}_{[t, T]}}{\mathrm d \cev{\P}^{v_i}_{[t, T]}} \,\bigg|\, X_t = x\right] \\
&\quad = \mathbb{E}_{\vec{\P}^{u}}\left[\frac{1}{2}\int_t^T \|u(s, X_s^{u})\|^2 \mathrm{d}s + \log \widehat{\varphi}_i(X_T^u) + \mathcal{E}(X_T^u) + C_1(x) \,\bigg|\, X_t =x\right],
\end{split}
\end{align}
where $C_1(x) = \log p_t(x) - \log \widehat{\varphi}_i(x) + \log \mathcal{Z}$, with $p_t = \vec{\P}^{u_i}_t$ (\Cref{app: notation}), represents terms that depend only on the initial state $x$ (and the fixed backward process), which are constant with respect to the optimization variable $u$.

For the second term (the regularization term), we decompose the Radon-Nikodym derivative via the reference measure. Since both $\vec{\P}^u_{[t,T]}$ and $\vec{\P}^{u_i}_{[t,T]}$ start from $p_t$ at time $t$ and have the densities $\varphi(X_T)/\varphi_t(x)$ and $\varphi_i(X_T)/\varphi_i(x, t)$ with respect to the reference process given $X_t = x$, we have:
\begin{align}
    \label{eq:reciprocal_term2}
    \mathbb{E}_{\vec{\P}^u} \left[ \log \frac{\mathrm d \vec{\P}^{u}_{[t, T]}}{\mathrm d \vec{\P}^{u_i}_{[t, T]}} \,\bigg|\, X_t=x \right]
    &= \mathbb{E}_{\vec{\P}^u} \left[ \log \left(\frac{\mathrm{d}\vec{\P}_{[t,T]}^u}{\mathrm{d}\vec{\P}_{[t,T]}^{\mathrm{ref}}} \bigg/ \frac{\mathrm{d}\vec{\P}_{[t,T]}^{u_i}}{\mathrm{d}\vec{\P}_{[t,T]}^{\mathrm{ref}}} \right) \,\bigg|\, X_t=x\right] \notag \\
    &= \mathbb{E}_{\vec{\P}^u} \left[ \log \frac{\varphi(X_T)}{\varphi_i(X_T)} + C_2(x)\,\bigg|\, X_t=x \right],
\end{align}
where $C_2(x) = \log \varphi_i(x, t) - \log \varphi_t(x)$.

Substituting \eqref{eq:reciprocal_term1} and \eqref{eq:reciprocal_term2} into the objective \eqref{eq: TRC-loss}, and grouping all terms independent of $u$ into $C(x, t) \coloneqq C_1(x) + \lambda \log \varphi_i(x, t)$, we arrive at the Reciprocal SOC formulation:
\begin{align}
    \mathcal{L}^{\mathrm{TRC}}_i(u, \lambda; x, t) = \mathcal{L}^{\mathrm{REC}}_i(u, \lambda; x, t) - \lambda \log \varphi_t(x) + C(x, t).
\end{align}
This concludes the proof.
\end{proof}

\begin{remark}
    This proposition provides the specific realization of the general relationship indicated in Eq.~\eqref{eq: L_TRC_relation} for our Trust Region Control objective $\mathcal{L}^{\mathrm{TRC}}_i$. 
    The formulation of $\mathcal{L}^{\mathrm{REC}}_i$ implicitly leverages the optimality condition that the solution must belong to the reciprocal class.
    Consequently, minimizing $\mathcal{L}^{\mathrm{REC}}$ over potentials normalized by $\varphi_t(x) = 1$ is equivalent to minimizing $\mathcal{L}^{\mathrm{TRC}}_i$. 
    A crucial advantage of this reformulation is that the trust region regularization term $D_{\mathrm{KL}}(\vec{\P}^u | \vec{\P}^{u_i})$, which originally involves path-dependent integration, collapses into a purely terminal cost term $\lambda \log (\varphi(X_T)/\varphi_i(X_T))$ under the reciprocal assumption, up to the term $-\lambda \log \varphi_t(x)$, which only fixes the normalization of $\varphi$.
\end{remark}

\begin{proposition}[Optimality for the reciprocal SOC problem]
\label{prop: optimality for REC-SOC}
Let $\vec{\P}^{u_i}$ be reciprocal with forward potential $\varphi_i$, i.e., $u_i = \sigma^\top \nabla \log \varphi_i$, and let $v_i = \sigma^\top \nabla \log \widehat{\varphi}_i$ be the exact backward half-step for $u_i$ (\Cref{lem: backward matching}). Define
\begin{align}
    \mathscr{V}^{\lambda}_{i+1}(x,t) &\coloneqq - \log \mathbb{E}_{\P^{\mathrm{ref}}} \big[ e^{-G(X_T)} \,\big|\, X_t = x \big], \label{eq: value function reciprocal} \\
    G &\coloneqq \tfrac{1}{1+\lambda}\big(\log \widehat{\varphi}_i(\cdot, T)+\mathcal{E}\big) - \tfrac{\lambda}{1+\lambda} \log \varphi_i. \label{eq: G reciprocal}
\end{align}
Then the updated control $u_{i+1}^{\lambda} = \argmin_{u \in \mathcal{U}} \mathcal{L}^{\mathrm{TR}}_i(u, \lambda)$ of \Cref{prop: geometric interpretation of update} lies in $\mathcal{U}_{\mathrm{REC}}$ and admits the representation
\begin{equation}
    u_{i+1}^{\lambda}(x, t) = - \sigma^\top \nabla \mathscr{V}_{i+1}^{\lambda}(x, t).
\end{equation}
\end{proposition}

\begin{proof}
Write $\mathscr{V}^{\lambda}_{i+1}(x, t) = -\log Z(x, t)$ with $Z(x, t) \coloneqq \E_{\P^{\mathrm{ref}}}[e^{-G(X_T)} \,|\, X_t = x]$. Since $\vec{\P}^{u_i}$ is reciprocal and $v_i$ is its exact backward half-step, $\vec{\P}^{u_i}_T = \varphi_i \widehat{\varphi}_i(\cdot, T)$ (\Cref{lem: backward matching}), so that the weight in \Cref{prop: geometric interpretation of update} reads $w_i = p_{\mathrm{target}} / \vec{\P}^{u_i}_T = e^{-\mathcal{E}} / (\mathcal{Z} \varphi_i \widehat{\varphi}_i(\cdot, T))$. By \Cref{prop: geometric interpretation of update} and the reciprocal factorization $\frac{\mathrm d \vec{\P}^{u_i}}{\mathrm d \P^{\mathrm{ref}}} = a_i(X_0) \varphi_i(X_T)$,
\begin{align}
    \frac{\mathrm d \vec{\P}^{u_{i+1}^\lambda}}{\mathrm d \P^{\mathrm{ref}}}(X) = \frac{a_i(X_0)}{c_i(X_0)} \varphi_i(X_T)\, w_i(X_T)^{\frac{1}{1+\lambda}} = \widetilde{a}(X_0)\, e^{-G(X_T)},
\end{align}
where $c_i(X_0) = \E_{\vec{\P}^{u_i}}[w_i^{\frac{1}{1+\lambda}} \,|\, X_0]$ and $\widetilde{a}$ collects all factors depending on $X_0$ alone, using $1 - \frac{1}{1+\lambda} = \frac{\lambda}{1+\lambda}$. Hence $\vec{\P}^{u_{i+1}^\lambda}$ is the Doob $h$-transform of the reference process with terminal function $e^{-G}$, i.e., $u_{i+1}^\lambda \in \mathcal{U}_{\mathrm{REC}}$ with potential $\varphi = e^{-G}$ and $\varphi_t = Z(\cdot, t)$, and its control is $u_{i+1}^\lambda = \sigma^\top \nabla \log Z = -\sigma^\top \nabla \mathscr{V}^\lambda_{i+1}$.
Equivalently, $\mathscr{V}^\lambda_{i+1}$ solves the HJB equation $\partial_t \mathscr{V} + f \cdot \nabla \mathscr{V} + \frac{1}{2}\operatorname{Tr}(\sigma \sigma^\top \nabla^2 \mathscr{V}) - \frac{1}{2}\|\sigma^\top \nabla \mathscr{V}\|^2 = 0$ with terminal condition $\mathscr{V}(\cdot, T) = G$, i.e., it is the value function of the SOC problem with running cost $\frac{1}{2}\|u\|^2$ and terminal cost $G$, which is the terminal cost $\Psi_i$ of \Cref{prop:reciprocal_soc_equivalence} evaluated at its own optimal potential $\varphi = e^{-G}$.
\end{proof}

\subsection{Proof of the main results in~\cref{sec: learning constrained optimal control}}
\label{app: proof CE matching}
\begin{proof}[Proof of Proposition \ref{eq: CE as matching objective}]
The derivation proceeds in three steps: identifying the optimal drift via the value function, applying the matching property of conditional expectations, and performing a change of measure to the sampling distribution.

\paragraph{Part 1: Characterization of the optimal drift.}
Let $u_{i+1}$ denote the optimal control minimizing the forward KL divergence $\mathcal{L}^\mathrm{CE}_i(u) = D_{\mathrm{KL}}(\vec{\P}^{u_{i+1}} | \vec{\P}^u)$. 
Equivalently, minimizing the KL divergence corresponds to minimizing the expected $L^2$ distance between the drifts under the optimal path measure $\vec{\P}^{u_{i+1}}$:
\begin{equation}
    \mathcal{L}^\mathrm{CE}_i(u) = \mathbb{E}_{\vec{\P}^{u_{i+1}}} \left[ \frac{1}{2} \int_0^T \| u(X_t, t) - u_{i+1}(X_t, t) \|^2 \mathrm{d}t \right].
\end{equation}
From Proposition \ref{prop: optimality for REC-SOC}, the optimal control is given by $u_{i+1}(x, t) = -\sigma^\top \nabla \mathscr{V}_{i+1}^{\lambda_i}(x, t)$.
The value function admits the Feynman-Kac representation:
\begin{equation}
    \mathscr{V}_{i+1}^{\lambda_i}(x, t) = -\log \mathbb{E}_{\vec{\P}^{\mathrm{ref}}} \left[ \exp\left( - G(X_T) \right) \mid X_t = x \right],
\end{equation}
where the effective terminal potential $G(X_T)$ is defined based on Eq.~\eqref{eq: value function reciprocal} as:
\begin{equation}
    G(X_T) \coloneqq \frac{1}{1+\lambda_i} \left( \log \widehat{\varphi}_i(X_T) + \mathcal{E}(X_T) \right) - \frac{\lambda_i}{1+\lambda_i} \log \varphi_i(X_T).
\end{equation}

We explicitly derive the expression for the optimal control. Recall that the value function is given by $\mathscr{V}_{i+1}^{\lambda}(x, t) = -\log Z(x, t)$, where $Z(x, t)$ serves as the local normalizing constant (partition function) defined by the conditional expectation under the reference measure:
\begin{align}
    Z(x, t) \coloneqq \mathbb{E}_{\vec{\P}^{\mathrm{ref}}} \left[ \exp\left( - G(X_T) \right) \mid X_t = x \right].
\end{align}
The optimal control is related to the gradient of the value function by $u_{i+1}(x, t) = -\sigma^{\top} \nabla \mathscr{V}_{i+1}^{\lambda_i}(x, t)$.
Applying the chain rule to the logarithmic term $\nabla (-\log Z) = -\frac{\nabla Z}{Z}$, and using that, for $f = 0$, $X_T = X_t + \int_t^T \sigma(s)\, \mathrm dW_s$ under the reference process, so that $\nabla_x$ commutes with the conditional expectation, we have:
\begin{subequations}
\begin{align}
    u_{i+1}(x, t) &=  -\sigma^{\top} \nabla \left( -\log \mathbb{E}_{\vec{\P}^{\mathrm{ref}}} [ e^{-G(X_T)} \mid X_t = x ] \right)\\
    &=-\sigma^{\top} \left( - \frac{\nabla \mathbb{E}_{\vec{\P}^{\mathrm{ref}}} [ e^{-G(X_T)} \mid X_t = x ]}{\mathbb{E}_{\vec{\P}^{\mathrm{ref}}} [ e^{-G(X_T)} \mid X_t = x ]} \right) \nonumber \\
    &= \sigma^{\top} \frac{1}{Z(x, t)} \nabla \mathbb{E}_{\vec{\P}^{\mathrm{ref}}} \left[ e^{-G(X_T)} \mid X_t = x \right] \\
    &= \sigma^{\top} \frac{1}{Z(x, t)}  \mathbb{E}_{\vec{\P}^{\mathrm{ref}}} \left[ \nabla \left( e^{-G(X_T)} \right) \mid X_t = x \right] \\
    &= \sigma^{\top} \frac{1}{Z(x, t)} \mathbb{E}_{\vec{\P}^{\mathrm{ref}}} \left[ e^{-G(X_T)} \left( - \nabla G(X_T) \right) \mid X_t = x \right] \\
    &= -\sigma^{\top} \mathbb{E}_{\vec{\P}^{\mathrm{ref}}}\left[\frac{\mathrm{d}\vec{\P}_{[t, T]}^{u_{i+1}}}{\mathrm{d}\vec{\P}^{\mathrm{ref}}_{[t, T]}} \nabla G(X_T) \mid X_t = x \right] \\
    &= -\sigma^{\top} \mathbb{E}_{\vec{\P}^{u_{i+1}}}\left[\nabla G(X_T) \mid X_t = x \right]
\end{align}
\end{subequations}
Here, the expectation is taken with respect to the optimal bridge process (which shares the same bridge dynamics as the reference), conditioned on the current state.
Computing the gradient $-\sigma^{\top} \nabla G(X_T)$ explicitly using the relations $v_i = \sigma^{\top} \nabla \log \widehat{\varphi}_i$ and $u_i = \sigma^{\top} \nabla \log \varphi_i$, we define the target vector field $Y_T$:
\begin{align}
    Y_T \coloneqq -\sigma^{\top} \nabla G(X_T) 
    &= - \frac{1}{1+\lambda_i}(v_i(X_T) + \sigma^{\top} \nabla \mathcal{E}(X_T)) + \frac{\lambda_i}{1+\lambda_i} u_i(X_T).
\end{align}
Here, $\sigma = \sigma(t)$ is evaluated at the time $t$ of the regressed control, and $v_i(X_T) \coloneqq \sigma(t)^\top \nabla \log \widehat{\varphi}_i(X_T)$ and $u_i(X_T) \coloneqq \sigma(t)^\top \nabla \log \varphi_i(X_T)$. These are the controls $v_i(X_T, T)$ and $u_i(X_T, T)$ at time $T$ multiplied by $\sigma(t)^\top \sigma(T)^{-\top}$, which is the identity for constant $\sigma$; in \eqref{eq: CE matching}, $v_i(X_T, T)$ and $u_i(X_T, T)$ are to be read in this sense. Our implementation uses a scalar $\sigma(t)$, parametrizes the controls as $u = \sigma u_\theta$ and $v = \sigma v_\phi$ (\Cref{sec:noise_schedule}), and regresses $u_\theta(X_t, t)$ onto $\sigma(t)^{-1} Y_T = -\nabla G(X_T)$, which contains no time-dependent factor.
Thus, the optimal control is the projection of this target: $u_{i+1}(X_t, t) = \mathbb{E}_{\vec{\P}^{u_{i+1}}}[Y_T \mid X_t]$.

\paragraph{Part 2: Conditional flow matching.}
Substituting $u_{i+1}$ back into the objective $\mathcal{L}^\mathrm{CE}_i(u)$, we obtain:
\begin{equation}
    \mathcal{L}^\mathrm{CE}_i(u) = \mathbb{E}_{\vec{\P}^{u_{i+1}}} \left[ \frac{1}{2} \int_0^T \| u(X_t) - \mathbb{E}_{\vec{\P}^{u_{i+1}}}[Y_T \mid X_t] \|^2 \mathrm{d}t \right].
\end{equation}
Invoking the \textit{matching property} of conditional expectations (\Cref{lem: matching}, cf.\ \citep{domingo2024stochastic}), minimizing $\| u - \mathbb{E}[Y|X] \|^2$ is equivalent to minimizing $\| u - Y \|^2$ up to a constant independent of $u$ (the variance of $Y$ given $X$). Therefore:
\begin{align}
\label{eq: CE before iw}
    \mathcal{L}^\mathrm{CE}_i(u) &= \mathbb{E}_{\vec{\P}^{u_{i+1}}} \left[ \frac{1}{2} \int_0^T \| u(X_t) - Y_T \|^2 \mathrm{d}t \right] + C \nonumber \\
    &= \mathbb{E}_{\vec{\P}^{u_{i+1}}} \left[ \frac{1}{2} \int_0^T \left\| u(X_t) + \frac{v_i(X_T) + \sigma^\top \nabla \mathcal{E}(X_T)}{1+\lambda_i} - \frac{\lambda_i}{1+\lambda_i} u_i(X_T) \right\|^2 \mathrm{d}t \right] + C.
\end{align}

\paragraph{Part 3: Change of measure.}
Finally, to evaluate this expectation using samples from the current policy $\vec{\P}^{u_i}$, we apply a change of measure. From Proposition \ref{prop: geometric interpretation of update}, see \eqref{eq: normalized annealing}, the Radon-Nikodym derivative (the importance weight) is computed by:
\begin{equation}
    \frac{\mathrm{d}\vec{\P}^{u_{i+1}}}{\mathrm{d}\vec{\P}^{u_i}}(X) = \left( \frac{\mathrm{d}\cev{\P}^{v_i}}{\mathrm{d}\vec{\P}^{u_i}}(X) \right)^{\frac{1}{1+\lambda_i}} \Bigg/ \E_{\vec{\P}^{u_i}}\left[\left( \frac{\mathrm{d}\cev{\P}^{v_i}}{\mathrm{d}\vec{\P}^{u_i}} \right)^{\frac{1}{1+\lambda_i}} \,\Bigg|\, X_0\right].
\end{equation}
Thus, the path expectation $\mathbb{E}_{\vec{\P}^{u_{i+1}}}$ can be implemented by sampling endpoints $(X_0, X_T)$ from $\vec{\P}^{u_i}$, re-weighting them, and then sampling the bridge path from $\P^{\mathrm{ref}}_{t|0,T}$. Note that we can rely on the reference bridge because $\vec{\P}^{u_i}$ belongs to the reciprocal class. 
Substituting the importance weight into~\cref{eq: CE before iw} yields the objective \eqref{eq: CE matching}.
\end{proof}

\subsection{TR-IPF and its convergence}
\label{app: convergence}
\Cref{alg: TR-IPF} states TR-IPF as an idealized iteration on path measures. With $\lambda_i = 0$ for all $i$, it reduces to IPF \eqref{eq: IPF iterations}, and with a fixed $\lambda_i = \lambda$ to the $\gamma$-IPF update \eqref{eq: gamma IPF} with $\gamma = (1+\lambda)^{-1}$. TR-SBS (\Cref{alg: TR-SBS high level}) implements each step with samples: the forward update by the matching objective \eqref{eq: CE matching}, the backward half-step by \eqref{eq: loss for v}, and the multiplier by the dual objective \eqref{eq: dual function}.

\begin{algorithm}[H]
\caption{Trust region IPF (TR-IPF), idealized iteration on path measures.}
\label{alg: TR-IPF}
\small
\begin{algorithmic}[1]
\Require Prior $p_\mathrm{prior}$, unnormalized target $p_\mathrm{target}$, initial control $u_0$, trust region radius $\varepsilon > 0$, number of iterations $I$
\For{$i = 0, \ldots, I-1$}
\State \parbox[t]{\dimexpr\linewidth-\algorithmicindent}{\textbf{Backward half-step:} $\cev{\P}^{v_i} = \argmin \{\KL(\mathbb{Q} | \vec{\P}^{u_i}) : \mathbb{Q}_T = p_\mathrm{target}\}$, so that $\frac{\mathrm d \cev{\P}^{v_i}}{\mathrm d \vec{\P}^{u_i}}(X) = w_i(X) \coloneqq \frac{p_\mathrm{target}}{\vec{\P}^{u_i}_T}(X_T)$ (\Cref{lem: terminal density ratio}).}
\State \parbox[t]{\dimexpr\linewidth-\algorithmicindent}{\textbf{Step size:} solve the saddle-point problem \eqref{eq: saddle-point problem} for $\lambda_i$ and set $\gamma_i = (1+\lambda_i)^{-1}$.}
\State \parbox[t]{\dimexpr\linewidth-\algorithmicindent}{\textbf{Forward update:} $\frac{\mathrm d \vec{\P}^{u_{i+1}}}{\mathrm d \vec{\P}^{u_i}}(X) = w_i(X)^{\gamma_i} \big/ \E_{\vec{\P}^{u_i}}[w_i^{\gamma_i} \,|\, X_0]$ (\Cref{prop: geometric interpretation of update}), which solves \eqref{eq: TR objective}.}
\EndFor
\State \Return $\vec{\P}^{u_I}$
\end{algorithmic}
\end{algorithm}

We restate \Cref{thm:convergence_rate} with its assumptions in full.
\begin{proposition}[Convergence rate of TR-IPF; detailed statement of \Cref{thm:convergence_rate}]
\label{thm:convergence_rate_app}
Let $(\vec{\P}^{u_i})_{i\ge 0}$ be the forward path measures generated by TR-IPF (\Cref{alg: TR-IPF}). TR-SBS generates them under the idealized assumption that each of its steps (the forward and backward matching problems and the Lagrange multiplier) is computed exactly and that $u_0 = v_0 = 0$, so that all iterates are reciprocal (\Cref{prop: preservation of reciprocal property}); since $v_0$ is in general not the backward half-step for $u_0$, this holds from the first backward update on, i.e., with $u_1$ in place of $u_0$. Let $p_{\text{prior}}$ and $p_{\mathrm{target}}$ be the initial and target distributions, respectively, and define the objective function as $F(\P) := D_{\mathrm{KL}}(\P_T | p_{\mathrm{target}})$. Let $\mathcal A := \{\P \mid \P_0 = p_{\mathrm{prior}}\}$ denote the admissible class of path measures satisfying the initial boundary condition, and let $\P^* \in \arg\min_{\P\in\mathcal A} F(\P)$ be the optimal global path measure.

Suppose that the trust region constraint is active for every iteration $i=0,\ldots,I-1$, i.e., it tightly holds that $D_{\mathrm{KL}}(\vec{\P}^{u_{i+1}}|\vec{\P}^{u_i}) = \varepsilon$ for a fixed trust region radius $\varepsilon>0$. For each $i$, let $\cev{\P}^{v_i}$ be the backward path measure at iteration $i$, and define the log-likelihood ratio $\ell_i := \log(\mathrm{d}\cev{\P}^{v_i}/\mathrm{d}\vec{\P}^{u_i})$. Assume that there exists a constant $M>0$ such that the variance of the update direction is bounded: $\operatorname{Var}_{\vec{\P}^{u_i}}(\ell_i) \le M$ for all $i\ge 0$.

Then $F(\vec{\P}^{u_{i+1}}) \le F(\vec{\P}^{u_i})$ for all $i$, and the gap after $I \ge 1$ steps is bounded at leading order by
\begin{equation}
    F(\vec{\P}^{u_I})-F(\P^*) = \min_{1\le i\le I} \left\{ F(\vec{\P}^{u_i})-F(\P^*) \right\} \lesssim \frac{D_{\mathrm{KL}}(\P^*|\vec{\P}^{u_0})\sqrt M}{I\sqrt{2\varepsilon}}.
\end{equation}
Consequently, to achieve an optimization gap of at most $\Delta>0$, the required number of outer iterations $I$ is $\mathcal O\left( \frac{D_{\mathrm{KL}}(\P^*|\vec{\P}^{u_0})\sqrt M}{\Delta\sqrt{\varepsilon}} \right)$.
\end{proposition}

\begin{proof}
The proof proceeds in three main steps.
\paragraph{Part 1: TR-IPF as mirror descent}
By \Cref{eq: CE as matching objective}, the actual forward control loss function is equivalent to $D_{\mathrm{KL}}(\vec{\P}^{u_{i+1}} | \vec{\P}^u)$ up to a constant independent of $u$. Furthermore, \Cref{prop: geometric interpretation of update} shows that the target measure $\vec{\P}^{u_{i+1}}$ satisfies the relation $\frac{\mathrm{d}\vec{\P}^{u_{i+1}}}{\mathrm{d}\vec{\P}^{u_i}} \propto \left( \frac{\mathrm{d}\cev{\P}^{v_i}}{\mathrm{d}\vec{\P}^{u_i}} \right)^{\gamma_i}$, where $\gamma_i = \frac{1}{1+\lambda_i}$ and the normalization is conditional on $X_0$, see \eqref{eq: normalized annealing}. Given this relation, optimizing the loss function 
\begin{equation}
    \min_{u \in \mathcal{U}} D_{\mathrm{KL}}(\vec{\P}^{u_{i+1}} | \vec{\P}^u)
\end{equation}
is equivalent to the optimization of 
\begin{equation}
    \min_{u \in \mathcal{U}} \left\{ D_{\mathrm{KL}}(\vec{\P}^u | \cev{\P}^{v_i}) + \frac{1-\gamma_i}{\gamma_i} D_{\mathrm{KL}}(\vec{\P}^u | \vec{\P}^{u_i}) \right\},
\end{equation}
which exactly corresponds to the forward control update of the $\gamma$-IPF. Furthermore, $\vec{\P}^{u_{i+1}}$ matches the minimizer of the exact population-level problem:
\begin{equation}
    \min_{\P \in \mathcal{A}} \left\{ D_{\mathrm{KL}}(\P | \cev{\P}^{v_i}) + \frac{1-\gamma_i}{\gamma_i} D_{\mathrm{KL}}(\P | \vec{\P}^{u_i}) \right\}.
\end{equation}
This exact matching is constructively shown in \Cref{prop: optimality for TR-SB}, guaranteeing that the optimal solution parameterized by the control $u$ coincides perfectly with the optimal solution in the general path measure space $\P$.

On the other hand, by \Cref{lem: backward matching}, the optimization of 
\begin{equation}
    \mathcal{L}^B_i(v) = \mathbb{E}_{\vec{\P}^{u_{i+1}}_{0,T} \P^{\mathrm{ref}}_{t|0,T}} \left[ \zeta_t \left\| v(X_t, t) - \sigma^\top \nabla \log p^{\mathrm{ref}}_{t|0}(X_t|X_0) \right\|^2 \right]
\end{equation}
yields the exact backward half-step $v_{i+1}$ for $u_{i+1}$, since $\vec{\P}^{u_{i+1}}$ is reciprocal (\Cref{prop: preservation of reciprocal property}). Hence \Cref{lem: terminal density ratio} gives $\ell_i(X) = \log \frac{p_\mathrm{target}}{\vec{\P}^{u_i}_T}(X_T)$ for all $i \ge 1$. Therefore, the overall alternating optimization procedure constitutes the $\gamma$-IPF, and the forward update coincides with the mirror descent update \eqref{eq: md update} with $\P_i = \vec{\P}^{u_i}$, $g = -\ell_i$, and the adaptively chosen step size $\gamma = \gamma_i = \frac{1}{1+\lambda_i} \in (0, 1]$, cf.~\Cref{sec:ipf_as_mirror_descent} and \cite{karimi2023sinkhorn}.

\paragraph{Part 2: Lower bound on the adaptive step size.}
The geometric form of the generalized IPF update implies that the next path measure is obtained by exponentially tilting the current path measure toward $\cev{\P}^{v_i}$. Then the update can be written as $\frac{\mathrm{d}\vec{\P}^{u_{i+1}}}{\mathrm{d}\vec{\P}^{u_i}} = \exp\left\{ \gamma_i \ell_i - \psi_i(\gamma_i; X_0) \right\}$, where $\psi_i(\gamma; x) := \log \mathbb E_{\vec{\P}^{u_i}} \left[ \exp(\gamma \ell_i) \,|\, X_0 = x \right]$ is the conditional log-partition function. Thus, for fixed $i$, the curve
\begin{equation}
    \gamma \mapsto \P_{i,\gamma}, \qquad \frac{\mathrm{d}\P_{i,\gamma}}{\mathrm{d}\vec{\P}^{u_i}} = \exp\left\{ \gamma \ell_i - \psi_i(\gamma; X_0) \right\},
\end{equation}
defines, conditionally on $X_0$, a one-dimensional exponential family on path space, with base measure $\vec{\P}^{u_i}(\cdot | X_0)$, natural parameter $\gamma$, and sufficient statistic $\ell_i$. In particular, we have $\P_{i,0} = \vec{\P}^{u_i}$ and $\P_{i,\gamma_i} = \vec{\P}^{u_{i+1}}$, and all $\P_{i,\gamma}$ have the initial marginal $p_\mathrm{prior}$.

By the standard local quadratic approximation of the Kullback-Leibler divergence in information geometry~\citep{amari2016information}, the KL divergence between two nearby points in a smooth parametric statistical model is, to second order, one half of the squared parameter displacement measured in the Fisher information metric: $D_{\mathrm{KL}}(P_{\theta+d\theta}|P_\theta) = \frac12 d\theta^\top \mathcal{I}(\theta)d\theta + o(\|d\theta\|^2)$.
Applying this fact to the family $\{\P_{i,\gamma}\}_{\gamma}$ at $\gamma=0$ conditionally on $X_0$ and averaging over $X_0 \sim p_\mathrm{prior}$ gives
\begin{equation}
    D_{\mathrm{KL}}(\P_{i,\gamma_i}|\P_{i,0}) = \frac{\gamma_i^2}{2} \mathcal{I}_i(0) + o(\gamma_i^2),
\end{equation}
where the Fisher information along this exponential family is given by the averaged conditional variance of the sufficient statistic, $\mathcal{I}_i(0) = \E_{X_0 \sim p_\mathrm{prior}}\big[\operatorname{Var}_{\vec{\P}^{u_i}}(\ell_i \,|\, X_0)\big] \le \operatorname{Var}_{\vec{\P}^{u_i}}(\ell_i)$.
Consequently, the KL divergence can be expanded as
\begin{equation}
    D_{\mathrm{KL}}(\vec{\P}^{u_{i+1}}|\vec{\P}^{u_i}) = \frac{\gamma_i^2}{2} \E_{X_0}\big[\operatorname{Var}_{\vec{\P}^{u_i}}(\ell_i \,|\, X_0)\big] + o(\gamma_i^2).
\end{equation}

Under the active trust region constraint, we have $D_{\mathrm{KL}}(\vec{\P}^{u_{i+1}} | \vec{\P}^{u_i}) = \varepsilon$. Using the uniform variance bound $\operatorname{Var}_{\vec{\P}^{u_i}}(\ell_i) \le M$ and neglecting the higher-order term in the small-step (small-$\varepsilon$) regime, we obtain at leading order:
\begin{equation}
    \varepsilon \lesssim \frac{\gamma_i^2}{2}M.
\end{equation}
Therefore, we establish the asymptotic lower bound for the adaptive step size as:
\begin{equation}
    \gamma_i \gtrsim \sqrt{\frac{2\varepsilon}{M}} =: \underline{\gamma}_{\varepsilon}.
\end{equation}

\paragraph{Part 3: Applying and telescoping the descent lemma.}
By Part 1, we can apply \Cref{lem: mirror descent step} with $\P_i = \vec{\P}^{u_i}$, $\gamma = \gamma_i \le 1$, and $\mathbb{Q} = \P^*$. This yields the following bound on the objective progress with respect to the global optimal path measure $\P^*$:
\begin{equation}
    \gamma_i \bigl( F(\vec{\P}^{u_{i+1}}) - F(\P^*) \bigr) \le D_{\mathrm{KL}}(\P^* | \vec{\P}^{u_i}) - D_{\mathrm{KL}}(\P^* | \vec{\P}^{u_{i+1}}).
\end{equation}
Summing this inequality over the outer iterations from $i=0$ to $I-1$, the right-hand side telescopes. Using the nonnegativity of the Kullback-Leibler divergence ($D_{\mathrm{KL}}(\P^* | \vec{\P}^{u_I}) \ge 0$), we obtain the accumulated bound:
\begin{equation}
    \sum_{i=0}^{I-1} \gamma_i \bigl( F(\vec{\P}^{u_{i+1}}) - F(\P^*) \bigr) \le D_{\mathrm{KL}}(\P^* | \vec{\P}^{u_0}) - D_{\mathrm{KL}}(\P^* | \vec{\P}^{u_I}) \le D_{\mathrm{KL}}(\P^* | \vec{\P}^{u_0}).
\end{equation}

\paragraph{Part 4: Best-iterate rate.}
Let $\Delta_I := \min_{1\le i\le I} \left\{ F(\vec{\P}^{u_i}) - F(\P^*) \right\}$ be the minimal objective gap over the first $I$ updates. By definition, we have $F(\vec{\P}^{u_{i+1}}) - F(\P^*) \ge \Delta_I$ for all $0\le i<I$. Substituting this into the accumulated sum from the telescoping bound yields:
\begin{equation}
    \Delta_I \sum_{i=0}^{I-1} \gamma_i \le \sum_{i=0}^{I-1} \gamma_i \bigl( F(\vec{\P}^{u_{i+1}}) - F(\P^*) \bigr) \le D_{\mathrm{KL}}(\P^* | \vec{\P}^{u_0}).
\end{equation}
From the local quadratic approximation in Part 2, we have the uniform lower bound for the step size at leading order in the small-$\varepsilon$ regime: $\gamma_i \gtrsim \underline{\gamma}_{\varepsilon} = \sqrt{2\varepsilon/M}$. Thus, the accumulated step size is bounded from below by $\sum_{i=0}^{I-1}\gamma_i \gtrsim I\sqrt{2\varepsilon/M}$. 

Combining this with the telescoping bound gives
\begin{equation}
    \Delta_I \left( I\sqrt{\frac{2\varepsilon}{M}} \right) \lesssim D_{\mathrm{KL}}(\P^* | \vec{\P}^{u_0}).
\end{equation}
Rearranging this inequality, we obtain the final best-iterate convergence rate:
\begin{equation}
    \min_{1\le i\le I} \left\{ F(\vec{\P}^{u_i}) - F(\P^*) \right\} = \Delta_I \lesssim \frac{D_{\mathrm{KL}}(\P^* | \vec{\P}^{u_0}) \sqrt M}{I\sqrt{2\varepsilon}}.
\end{equation}
Since $F(\vec{\P}^{u_{i+1}}) \le F(\vec{\P}^{u_i})$ by \Cref{lem: mirror descent step}, the minimum is attained at $i = I$, so that the bound also holds for the last iterate.
To ensure that the optimization gap on the right-hand side is at most $\Delta>0$, it suffices to require
\begin{equation}
    I \ge \frac{D_{\mathrm{KL}}(\P^* | \vec{\P}^{u_0}) \sqrt M}{\Delta\sqrt{2\varepsilon}}.
\end{equation}
Hence, the required number of outer iterations satisfies
\begin{equation}
    I = \mathcal O\left( \frac{D_{\mathrm{KL}}(\P^* | \vec{\P}^{u_0}) \sqrt M}{\Delta\sqrt{\varepsilon}} \right).
\end{equation}
This completes the proof. 
\end{proof}

\subsection{Remarks on the variance of importance weights}
\label{app: variance of importance weights}

Recall from \Cref{sec: learning constrained optimal control} that, at iteration $i$ of the Trust Region Schrödinger Bridge Sampler (TR-SBS), the next control is obtained by minimizing the cross-entropy objective
\begin{equation}
    \mathcal{L}_i^\mathrm{CE}(u) := \KL(\vec{\P}^{u_{i+1}} | \vec{\P}^{u}).
\end{equation}
To make this objective tractable, we employ importance sampling. Since we can generate trajectories $X^{u_i}$ from the previous iterate, we rewrite
\begin{equation}
    \mathcal{L}_i^\mathrm{CE}(u)
    =
    \E_{\vec{\P}^{u_i}}\!\left[
         \left( \log \frac{\mathrm d \vec{\P}^{u_{i+1}}}{\mathrm d \vec{\P}^{u}}(X^{u_i}) \right)
        \frac{\mathrm d \vec{\P}^{u_{i+1}}}{\mathrm d \vec{\P}^{u_i}}(X^{u_i})
    \right].
\end{equation}

In general, importance reweighting introduces exponential terms that may lead to large variance and numerical instabilities, especially in high-dimensional settings; see, e.g., \cite{hartmann2024nonasymptotic}. In our case, however, the trust region constraint
\begin{equation}
    \KL(\vec{\P}^{u} | \vec{\P}^{u_i}) \le \varepsilon
\end{equation}
(cf.~\eqref{eq: TR objective}) implicitly controls the variance of the importance weights. This connection is made precise by the inequality
\begin{equation}
    \operatorname{Var}_{\vec{\P}^{u_i}}\!\left(
        \frac{\mathrm d \vec{\P}^{u_{i+1}}}{\mathrm d \vec{\P}^{u_i}}
    \right)
    \ge
    \exp\!\Big(
        \KL(\vec{\P}^{u_{i+1}}| \vec{\P}^{u_i})
    \Big) - 1.
\end{equation}
Hence, if the KL divergence grows, the variance of the importance weights is forced to increase exponentially. Controlling the KL divergence is therefore a necessary condition to avoid variance explosion.

Empirically, we observe that enforcing this constraint is sufficient to keep the variance small in practice. We stress, however, that the following argument is local: it relies on a second-order expansion around $\frac{\mathrm d \vec{\P}^{u_{i+1}}}{\mathrm d \vec{\P}^{u_i}} \approx 1$ and does not provide a global bound on the variance. For small values of $\varepsilon$, as typically chosen in applications, we consistently find
\begin{equation}
    \operatorname{Var}_{\vec{\P}^{u_i}}\!\left(
        \frac{\mathrm d \vec{\P}^{u_{i+1}}}{\mathrm d \vec{\P}^{u_i}}
    \right)
    \approx 2\varepsilon,
\end{equation}
see Appendix~I.3 in \cite{blessing2025trust}.
This behavior follows at leading order in $\varepsilon$ from the exponential family argument in Part~2 of the proof of \Cref{thm:convergence_rate_app}. By \Cref{prop: geometric interpretation of update}, the weight $W \coloneqq \frac{\mathrm d \vec{\P}^{u_{i+1}}}{\mathrm d \vec{\P}^{u_i}} = \exp\{\gamma_i \ell_i - \psi_i(\gamma_i; X_0)\}$ satisfies $\E_{\vec{\P}^{u_i}}[W \,|\, X_0] = 1$, and expanding in $\gamma_i$ gives $W - 1 = \gamma_i \big(\ell_i - \E_{\vec{\P}^{u_i}}[\ell_i \,|\, X_0]\big) + \mathcal{O}(\gamma_i^2)$. Hence
\begin{equation}
    \operatorname{Var}_{\vec{\P}^{u_i}}(W) = \gamma_i^2 \widebar{V}_i + o(\gamma_i^2) = 2 \KL(\vec{\P}^{u_{i+1}} | \vec{\P}^{u_i}) + o(\gamma_i^2) \approx 2\varepsilon,
\end{equation}
with $\widebar{V}_i = \E_{X_0 \sim p_\mathrm{prior}}[\operatorname{Var}_{\vec{\P}^{u_i}}(\ell_i \,|\, X_0)]$ as in \Cref{sec: convergence}, where we used $\KL(\vec{\P}^{u_{i+1}} | \vec{\P}^{u_i}) = \frac{\gamma_i^2}{2} \widebar{V}_i + o(\gamma_i^2)$ from Part~2. The same expansion shows that the log-weight $\log W$ has variance $2\varepsilon$ at leading order. Both statements are local: the neglected terms involve higher cumulants of $\ell_i$, which can be large when $\vec{\P}^{u_i}$ is far from $\cev{\P}^{v_i}$.

\subsection{Relation to trust region methods on path space \texorpdfstring{\cite{blessing2025trust}}{}}
\label{app: relation to blessing}

Our trust region formulation is inspired by \cite{blessing2025trust}, who constrain consecutive updates of a stochastic optimal control problem in KL divergence on path space. The present work differs in the following three respects.

\textbf{Moving-target trust region formulation (\Cref{prop: geometric interpretation of update}).} In \cite{blessing2025trust}, the updates are optimized toward a fixed target path measure. In contrast, the target $\cev{\P}^{v_i}$ of our forward update changes after each backward IPF step. We formulate the trust region update for this moving-target sequence and interpret the resulting dual variable as an adaptive $\gamma$-IPF step size $\gamma_i = (1+\lambda_i)^{-1}$.

\textbf{SB-specific stochastic-control reformulation (\Cref{prop: optimality for TR-SB,prop:reciprocal_soc_equivalence}).} By exploiting the reciprocal structure of the IPF iterates, we reformulate the path space trust region update as an equivalent terminal-cost stochastic optimal control problem. This reformulation leads to the importance-weighted generalized adjoint-matching objective of \Cref{eq: CE as matching objective}, which can be evaluated tractably using endpoints sampled from the previous iterate.

\textbf{Trust region IPF interpretation and convergence analysis (\Cref{thm:convergence_rate}).} We interpret the trust region iteration as a $\gamma$-IPF, or mirror descent, procedure with adaptive step size and establish a convergence guarantee for the resulting moving-target SB sequence.

\subsection{Relation to ASBS and limiting cases}
\label{app: relation to ASBS}\label{sec: relation to ASBS}
\Cref{fig: overview} places TR-SBS among related IPF-based samplers. Moving from the first to the second row implements the path space update with samples of $\vec{\P}^{u_i}$, which requires the Radon-Nikodym weight to change the measure to the desired next iterate; without this weight (third row), one obtains fixed-point iterations such as ASBS (\Cref{sec: learning constrained optimal control}). Moving from left to right changes how the mirror-descent step size is selected. ASBS-IW and TR-ASBS isolate the two ingredients of TR-SBS and serve as ablations in \Cref{sec: experiments}. ASBS itself arises as a limiting case: when the KL constraint in \eqref{eq: TR objective} is inactive, complementary slackness gives $\lambda_i = 0$ and hence $\gamma_i = 1$, so TR-IPF reduces to IPF, TR-ASBS to ASBS, and TR-SBS to ASBS-IW. As $\varepsilon \to 0$, the update vanishes, with $\lambda_i \to \infty$ and $\gamma_i \to 0$.

\begingroup
\subsection{Anchoring the matching target at the regressed time}
\label{app: anchor}
On LJ-13 and alanine dipeptide, we replace the anchor $u_i(X_T, T)$ in \eqref{eq: CE matching} by $u_i(X_t, t)$, the previous control at the time and state of the regressed control (\Cref{app:aldp_detail}). The following lemma compares a single update with the original and with the modified anchor, starting from the same iterate $\vec{\P}^{u_i}$, which we assume to be obtained from $u_0 = 0$ by exact updates as in \Cref{eq: CE as matching objective}. We write $\gamma_i = 1/(1+\lambda_i)$; for a time-dependent $\sigma$, $u_i(X_T, T)$ and $v_i(X_T, T)$ are multiplied by $\sigma(t)^\top \sigma(T)^{-\top}$, as in \eqref{eq: CE matching} (\Cref{app: proof CE matching}).

\begin{lemma}[Anchor at the regressed time]
\label{lem: anchor}
Let
\begin{equation*}
    \widetilde{\mathcal{L}}_i^{\mathrm{F}}(u) \coloneqq \mathbb{E}_{\vec{\P}^{u_i}_{0, T}\vec{\P}^{\mathrm{ref}}_{t|0,T}} \! \bigg[ \Big\| u(X_t, t) \!+\! \gamma_i \big( v_i(X_T, T) \!+\! \sigma^{\top} \nabla \mathcal{E}(X_T) \big) \!-\! (1-\gamma_i)\, u_i(X_t, t) \Big\|^2 \frac{\mathrm{d}\vec{\P}^{u_{i+1}}}{\mathrm d \vec{\P}^{u_i}} (X) \bigg]
\end{equation*}
denote the loss \eqref{eq: CE matching} with the anchor $u_i(X_T, T)$ replaced by $u_i(X_t, t)$, and define the importance-weighted adjoint matching target {\relpenalty=10000\binoppenalty=10000 $\widetilde{u}_{i+1}(x, t) \coloneqq -\E_{\vec{\P}^{u_{i+1}}}[v_i(X_T, T) + \sigma^\top \nabla \mathcal{E}(X_T) \mid X_t = x]$}. Then:
\begin{enumerate}
    \item The previous control satisfies $u_i(x, t) = \E_{\vec{\P}^{u_i}}[u_i(X_T, T) \mid X_t = x]$, whereas the minimizer of \eqref{eq: CE matching} is $u_{i+1}(x, t) = (1-\gamma_i)\, \E_{\vec{\P}^{u_{i+1}}}[u_i(X_T, T) \mid X_t = x] + \gamma_i\, \widetilde{u}_{i+1}(x, t)$.
    \item The minimizer of $\widetilde{\mathcal{L}}_i^{\mathrm{F}}$ is $(1-\gamma_i)\, u_i + \gamma_i\, \widetilde{u}_{i+1}$.
    \item For $\gamma_i = 1$, both minimizers equal $\widetilde{u}_{i+1}$.
    \item For every $\gamma_i \in (0, 1]$, if $\vec{\P}^{u_i}$ is the Schr\"odinger bridge, both minimizers equal $u_i$, i.e., the Schr\"odinger bridge is a fixed point of both iterations.
\end{enumerate}
\end{lemma}

\begin{proof}
1. For $i = 0$, $u_0 = 0$ and the first identity is trivial. For $i \ge 1$, $\vec{\P}^{u_i}$ results from an exact update, so that, as shown in \Cref{app: proof CE matching} for $\vec{\P}^{u_{i+1}}$, it is the reference process tilted by a terminal potential,
\begin{equation}
    \frac{\mathrm d \vec{\P}^{u_i}_{[t, T]}}{\mathrm d \vec{\P}^{\mathrm{ref}}_{[t, T]}} = \frac{\varphi_i(X_T)}{\varphi_i(X_t, t)},
\end{equation}
with $\varphi_i(x, t) = \E_{\vec{\P}^{\mathrm{ref}}}[\varphi_i(X_T) \mid X_t = x]$ and $u_i = \sigma^\top \nabla \log \varphi_i$. The computation in \Cref{app: proof CE matching}, with $e^{-G(X_T)}$ and $Z(x, t)$ replaced by $\varphi_i(X_T)$ and $\varphi_i(x, t)$, gives $u_i(x, t) = \sigma^\top \E_{\vec{\P}^{u_i}}[\nabla \log \varphi_i(X_T) \mid X_t = x] = \E_{\vec{\P}^{u_i}}[u_i(X_T, T) \mid X_t = x]$. The second identity is $u_{i+1}(x, t) = \E_{\vec{\P}^{u_{i+1}}}[Y_T \mid X_t = x]$ (\Cref{app: proof CE matching}) with $Y_T$ split into its two terms.

2. By Part 3 of the proof of \Cref{eq: CE as matching objective}, the weighted expectation in \eqref{eq: CE matching}, and hence in $\widetilde{\mathcal{L}}_i^{\mathrm{F}}$, is an expectation under $\vec{\P}^{u_{i+1}}$. By the matching property (\Cref{lem: matching}), the minimizer of $\widetilde{\mathcal{L}}_i^{\mathrm{F}}$ is therefore the conditional expectation of its regression target given $X_t$ under $\vec{\P}^{u_{i+1}}$. Since $u_i(X_t, t)$ is a function of $X_t$, this conditional expectation is $(1-\gamma_i)\, u_i(X_t, t) + \gamma_i\, \widetilde{u}_{i+1}(X_t, t)$.

3. For $\gamma_i = 1$, the anchor has weight zero in both losses.

4. If $\vec{\P}^{u_i}$ is the Schr\"odinger bridge, its terminal marginal is $p_{\mathrm{target}}$, so that the backward half-step does not change it, $\cev{\P}^{v_i} = \vec{\P}^{u_i}$. By \Cref{prop: geometric interpretation of update}, the importance weight is then $1$ and $\vec{\P}^{u_{i+1}} = \vec{\P}^{u_i}$, so that the two conditional expectations of the anchor in statements 1 and 2 coincide and equal $u_i$. Moreover, since $\vec{\P}^{u_{i+1}} = \cev{\P}^{v_i}$, \Cref{eq: CE as matching objective} with $\lambda_i = 0$ shows that $\widetilde{u}_{i+1}$ is the control of the update with $\gamma_i = 1$, whose path measure is $\cev{\P}^{v_i} = \vec{\P}^{u_i}$; hence $\widetilde{u}_{i+1} = u_i$, and both minimizers equal $(1-\gamma_i)\, u_i + \gamma_i\, u_i = u_i$.
\end{proof}

The modification thus replaces the conditional expectation of the anchor under the updated path measure by that under the previous one. Away from the Schr\"odinger bridge, the two differ in general, so that $\widetilde{\mathcal{L}}_i^{\mathrm{F}}$ is not the cross-entropy loss \eqref{eq: cross-entropy loss}. Moreover, the minimizer in statement 2 need not be of the form $\sigma^\top \nabla \log \varphi$ with $\varphi$ as in the proof, so that the next iterate need not be the reference process tilted by a terminal potential. The lemma then does not apply to the next update, and sampling $X_t$ from the reference bridge is no longer exact. The iterates with the modified anchor are therefore not those of TR-IPF, and our analysis in \Cref{sec: convergence} does not cover them. Statement 4 is not affected: an iteration started at the Schr\"odinger bridge stays there.

We nevertheless consider the modification reasonable, for three reasons. First, the Schr\"odinger bridge remains a fixed point (statement 4). Second, by statement 1, the minimizers of the two losses differ by $(1-\gamma_i)\big(\E_{\vec{\P}^{u_{i+1}}} - \E_{\vec{\P}^{u_i}}\big)[u_i(X_T, T) \mid X_t]$. This difference is caused only by the tilt $\big(p_{\mathrm{target}}/\vec{\P}^{u_i}_T\big)^{\gamma_i}(X_T)$, up to normalization, that distinguishes $\vec{\P}^{u_{i+1}}$ from $\vec{\P}^{u_i}$, so that it vanishes at the Schr\"odinger bridge and is small when this tilt is close to constant. Since it equals $\gamma_i \operatorname{Cov}_{\vec{\P}^{u_i}}\big(u_i(X_T, T), \log w_i(X_T) \,\big|\, X_t\big) + \mathcal{O}(\gamma_i^2)$, it is, however, of the same order in $\gamma_i$ as the update itself, so that the modified iteration differs from TR-IPF already at first order. Third, the modification removes one source of estimation error of the regression target from finite samples (\Cref{app:aldp_detail}): in \eqref{eq: CE matching}, the previous control enters the target only through its terminal values $u_i(X_T, T)$, whose conditional expectation $\E_{\vec{\P}^{u_{i+1}}}[u_i(X_T, T) \mid X_t]$ has to be estimated by the regression, whereas in $\widetilde{\mathcal{L}}_i^{\mathrm{F}}$, it enters directly at the time and state of the regressed control. Empirically, TR-SBS fails on LJ-13 without the two modifications of \Cref{app:aldp_detail} (\Cref{app: lj13}).
\endgroup

\section{Experiment details}
\label{app:exp_details}

In this section, we provide a comprehensive description of the experimental setup, evaluation metrics, and implementation details for the four benchmarks used to evaluate the mode-seeking capabilities and scalability of our method: the Gaussian Mixture Model (GMM), the Many-Well (MW) potential, the Ginzburg-Landau (GL) model, and alanine dipeptide.

\subsection{Gaussian mixture model (GMM)}
\label{app:gmm_detail}

To assess the ability to capture multiple isolated modes in high-dimensional spaces, we conduct evaluations under two distinct settings: one focusing on scalability with respect to dimensionality, and the other on robustness to an increasing number of modes.

\subsubsection{Problem setup}
The target density $p(x)$ is defined as a mixture of $K$ multivariate Gaussian components in $\mathbb{R}^d$:
\begin{equation}
    p(x) = \sum_{k=1}^{K} \pi_k \, \mathcal{N}(x | \mu_k, \Sigma_k).
\end{equation}
The component means are sampled uniformly from a hypercube $\mu_k \sim \operatorname{Uniform}([-L, L]^d)$. For the covariances, we use isotropic matrices $\Sigma_k = \sigma^2 I_d$ with $\sigma = \operatorname{Softplus}(1.0) \approx 1.31$ across all experiments. In both settings, the mixing weights $\pi_k$ are \textit{non-uniform} with a maximum probability ratio $\max_k \pi_k / \min_k \pi_k$ of $3.0$. The target (means and weights) is generated once with a fixed seed and shared by all methods and training seeds.

\paragraph{Experimental configurations}
The specific configurations for the target distribution and the prior distribution differ between the two experiments to create appropriate difficulty levels for each task.

\begin{enumerate}
    \item \textbf{Scalability with dimensionality:} 
    We fix $K=10$ and vary $d \in \{10, 25, 50, 100\}$. The means are sampled with $L=7$. The prior distribution is set to $\mathcal{N}(0, 5.0^2 I_d)$.
    
    \item \textbf{Robustness to increasing modes:} 
    We fix $d=10$ and vary $K \in \{5, 10, 20, 30, 40\}$. To accommodate a larger number of modes without excessive overlap, we use a wide range $L=40$. The prior distribution covers this region with $\mathcal{N}(0, 30^2 I_d)$.
\end{enumerate}

\paragraph{Training and model selection}
For each method and setting, we train with three seeds. The ASBS-family methods are trained for at most $150$ outer iterations, and PIS and DIS for $40{,}000$ gradient steps (\Cref{app: pis dis}). During training, we periodically compute the SD to the target and keep the checkpoint with the lowest value. This checkpoint is then evaluated on $4{,}000$ independent samples with a fixed evaluation seed, and we report the mean and sample standard deviation of these evaluations over the three seeds. Note that this model selection uses samples from the target, which are available for this synthetic benchmark.
    
\subsubsection{Evaluation metrics}
We employ the following metrics to evaluate the recovery of the multi-modal distribution:
\begin{itemize}
    \item \textbf{Total variation distance (TVD):} Measures the $L_1$ discrepancy between the true mixing weights $\pi$ and the empirical weights $\widehat{\pi}$ estimated from $N$ samples.
    \begin{equation}
        \text{TVD}(\pi, \widehat{\pi}) = \frac{1}{2} \sum_{k=1}^K | \widehat{\pi}_k - \pi_k |
    \end{equation}
    To compute the empirical weights $\widehat{\pi}_k$ specifically for the soft assignment metric, we utilize the posterior responsibility $\gamma_{ik}$ of component $k$ for each generated sample $x_i$. This is calculated using the ground-truth parameters as follows:
    \begin{equation}
        \gamma_{ik} = \frac{\pi_k \mathcal{N}(x_i | \mu_k, \Sigma_k)}{\sum_{j=1}^K \pi_j \mathcal{N}(x_i | \mu_j, \Sigma_j)}.
    \end{equation}
    The soft empirical weight is then derived by averaging these responsibilities over the batch:
    \begin{equation}
        \widehat{\pi}_k = \frac{1}{N} \sum_{i=1}^N \gamma_{ik}.
    \end{equation}
    
    \item \textbf{Sinkhorn divergence:} To assess geometric fidelity beyond categorical mass assignment, we compute the Sinkhorn Divergence $S_{\varepsilon}(\mu, \nu)$ between the true and empirical distributions ($\varepsilon=10^{-3}$).
\end{itemize}
For the final evaluation, we use $N=4{,}000$ generated samples and the same number of ground-truth samples. In \Cref{app: gmm extra metrics}, we additionally report the 2-Wasserstein distance ($W_2$), the sliced 2-Wasserstein distance, and the normalized effective sample size of the raw path weights $\frac{\mathrm d \cev{\P}^{v}}{\mathrm d \vec{\P}^{u}}$ (raw path ESS).

\subsection{Many-Well potential (MW)}
\label{app:mw_detail}

To evaluate the sampler's ability to navigate high-dimensional landscapes with a combinatorial number of modes and cross high energy barriers, we employ the Many-well potential. We conduct two sets of experiments: one focusing on the robustness to the number of modes (increasing $m$), and the other on the robustness to the geometric distance between modes (increasing $\delta$).

\subsubsection{Problem setup}
The energy function $\mathcal{E}(x)$ for a configuration $x \in \mathbb{R}^d$ is composed of non-convex double-well potentials in the first $m$ dimensions and a harmonic potential in the remaining $d-m$ dimensions:
\begin{equation}
    \mathcal{E}(x) = \sum_{i=1}^m (x_i^2 - \delta)^2 + \frac{1}{2} \sum_{j=m+1}^d x_j^2,
\end{equation}
where $m$ denotes the number of active double-well dimensions and $\delta$ controls the location of the minima ($x_i = \pm \sqrt{\delta}$). The total number of symmetric global minima is $2^m$.

\paragraph{Experimental configurations}
We configure the experiments as follows to test specific capabilities of the sampler:

\begin{enumerate}
    \item \textbf{Robustness to increasing modes (varying $m$):} 
    We fix the total dimensionality at $d=64$ and the barrier parameter at $\delta=4.0$. We vary the number of active double-wells $m \in \{4, 8, 16, 32\}$. At $m=32$, the landscape contains $2^{32}$ modes, posing an extreme challenge for global mass discovery.
    
    \item \textbf{Robustness to mode distance (varying $\delta$):} 
    We fix the dimensionality at $d=8$ with $m=5$ active wells ($32$ modes). We vary the well parameter $\delta \in \{2, 4, 6, 8\}$. Increasing $\delta$ expands the geometric distance between modes and heightens the energy barriers, testing the sampler's ability to escape local minima.
\end{enumerate}

\subsubsection{Evaluation metrics}
To quantitatively assess the sample quality, we leverage the separability of the Many-Well potential, which allows us to compute ground-truth statistics analytically or via high-precision numerical integration.
\begin{itemize}
    \item \textbf{Marginal Earth Mover's distance (EMD):} We compute the squared 2-Wasserstein distance, i.e., the optimal transport cost with squared Euclidean cost, between the empirical and ground-truth marginal distributions for each dimension $j$. We report the mean ($\text{EMD}_{\text{mean}}$) across all $d$ dimensions.

    \item \textbf{Log-Partition Function Estimation Error ($\Delta_{\log Z}$):} 
    To evaluate the global mass balance across the $2^m$ symmetric modes, we define the estimation error as the deviation of the estimated partition function $\widehat{Z}$ from the ground truth $Z_{\text{true}}$:
    \begin{equation}
        \Delta_{\log Z} = \left| \log \widehat{Z} - \log Z_{\text{true}} \right|.
    \end{equation}
    The estimator $\widehat{Z}$ is derived by exploiting the symmetry of the potential. We define a specific quadrant $B = \{x \in \mathbb{R}^d : x_1, \dots, x_m > 0\}$, which theoretically contains exactly $2^{-m}$ of the total probability mass. Using the empirical mass of this quadrant $\widehat{\pi}_B$, we estimate the global constant as $\widehat{Z} = Z_{\text{true}} \cdot (2^m \widehat{\pi}_B)^{-1}$.
    
    The ground-truth constant $Z_{\text{true}}$ is obtained analytically due to the separability of the energy function:
    \begin{equation}
        \log Z_{\text{true}} = m \log Z_{DW}(\delta) + (d-m) \log \sqrt{2\pi}.
    \end{equation}
    Here, $Z_{DW}(\delta)$ denotes the partition function of the 1D double-well component, which we calculate via high-precision numerical integration:
    \begin{equation}
        Z_{DW}(\delta) = \int_{-\infty}^{\infty} \exp\left( - (u^2 - \delta)^2 \right) du.
    \end{equation}
    In practice, this integral is approximated using numerical quadrature over a sufficiently large bounded interval to ensure negligible truncation error.
\end{itemize}

\subsection{Ginzburg-Landau ($\phi^4$ scalar field theory) model}
\label{app:gl_detail}

To evaluate our method on a physically motivated problem with strong spatial correlations and continuous phase transitions, we employ the discretized Ginzburg-Landau model.

\subsubsection{Problem setup}
The target distribution is defined by the energy functional:
\begin{equation}
    \mathcal{E}(x) = \frac{1}{2}\sum_{i=1}^{d} (x_i - x_{i+1})^2 + \frac{1}{2}\sum_{i=1}^m (x_i^2 - \delta)^2 + \frac{1}{2}\sum_{i=m+1}^d x_i^2,
\end{equation}
where $d$ is the total dimensionality, $m$ is the number of active double-well dimensions, and $\delta$ dictates the location of the minima. We apply periodic boundary conditions such that $x_{d+1} = x_1$, with a coupling strength of $1$.

Based on our configuration, we use $d=16$, $m=5$, and $\delta=6$.

\subsubsection{Evaluation metrics}
Unlike the Gaussian Mixture Model or the Many-Well potential, the Ginzburg-Landau energy lacks a tractable normalizing constant and ground-truth samples. Therefore, we evaluate performance through the following intrinsic metrics:

\begin{itemize}
    \item \textbf{Normalized effective sample size (ESS):} To measure the statistical efficiency and stability of the sampling, we calculate the normalized ESS derived from the importance weights $w = \frac{\mathrm{d}\cev{\P}^{v}}{\mathrm{d}\vec{\P}^{u}}(X^{u})$. The values are normalized to the range $[0, 1]$, where a higher value indicates that the trust region constraint successfully maintained low-variance updates.
    
    \item \textbf{Symmetry violation (mirror marginal EMD):} Since the target energy obeys $\mathcal{E}(x) = \mathcal{E}(-x)$, the predictive distribution should ideally satisfy $p(x) = p(-x)$. To quantify this, we compute the 2-Wasserstein distance between the marginals of the predictive distribution $\vec{\P}^{u}_{T}$ and of its own reflection $-\vec{\P}^{u}_{T}$, instead of comparing with ground-truth samples as in the MW experiments. We report the mean EMD across all dimensions ($\widebar{W}_2(X_T, -X_T)$). This allows us to assess global mode balance and symmetry preservation in the absence of ground-truth data.
\end{itemize}

\begingroup
\subsection{Alanine dipeptide}
\label{app:aldp_detail}

\subsubsection{Problem setup}
We sample the Boltzmann distribution of alanine dipeptide (22 atoms) at $300\,$K, modeled with the Amber ff96 force field and the OBC implicit solvent model in OpenMM, as in \cite{midgley2023flow}. Following \cite{midgley2023flow,liu2025adjoint}, the sampler operates in internal coordinates (bond lengths, bond angles, and dihedral angles; $d=60$ after removing the six external degrees of freedom). The target density includes the Jacobian determinant of the coordinate transform. We use $10^6$ reference samples from molecular dynamics. Alanine dipeptide is chiral: the four substituents of C$_\alpha$ can be arranged in two mirror-image forms, L and D, which cannot be converted into each other without breaking bonds. The energy is the same for both forms, so the target also has modes of the D-form, while the reference samples contain only the L-form. We determine the form of a sample from the sign of $(x_{\mathrm{N}} - x_{\mathrm{C}_\alpha}) \cdot \big((x_{\mathrm{C}} - x_{\mathrm{C}_\alpha}) \times (x_{\mathrm{C}_\beta} - x_{\mathrm{C}_\alpha})\big)$, where $x_{\mathrm{C}_\alpha} \in \R^3$ is the Cartesian position of C$_\alpha$ and $x_{\mathrm{N}}$, $x_{\mathrm{C}}$, and $x_{\mathrm{C}_\beta}$ are those of the nitrogen, the carbonyl carbon, and the methyl carbon bonded to it; this sign is the same for all reference samples.

\paragraph{Prior.} Both methods use the prior $(1-w)\,\mathcal{N}(0, I) + w\,\rho_\phi$ with $w = 0.02$. In $\rho_\phi$, the Gaussian component along the direction of a rigid rotation about the torsion $\phi$ is replaced by a uniform distribution over one full turn of $\phi$, cut at $\phi = 2.2$, where the target density is negligible. Thus, $2\%$ of the initial states cover all values of $\phi$, including the C7ax region $0 \le \phi \le 2$, while the bulk of the prior is unchanged. This prior differs from the Gaussian prior $\mathcal{N}(0, 0.25\,I)$ of \cite{liu2025adjoint}, with which ASBS essentially does not find C7ax (see their Ramachandran plot).

\paragraph{Hyperparameters.} The controls $u$ and $v$ are time-conditional MLPs with Fourier time features, $8$ layers, and $256$ hidden units, whose outputs are clipped componentwise to $[-100, 100]$. We use $\sigma_{\mathrm{prior}} = 1$, $\sigma_{\min} = 10^{-3}$, $\sigma_{\max} = 0.5$, $N_t = 200$, $I = 40$, $N_{\mathrm{FC}} = 2{,}000$, $N_{\mathrm{BC}} = 1{,}000$, and $L = 100$, i.e., $120{,}000$ epochs in total, and Adam with learning rate $10^{-4}$ and gradient-norm clipping at $100$. Training runs data-parallel on $4$ NVIDIA A40 GPUs with, per GPU, batch size $B = 500$, buffer size $N_{\mathrm{buf}} = 40{,}000$, and $N_{\mathrm{new}} = 1{,}000$. TR-SBS uses $\varepsilon = 0.2$, $c_{\mathrm{IW}} = 10$, $N_{\mathrm{state}} = 500$ per GPU, i.e., $2{,}000$ initial states for the dual problem, and $N_{\mathrm{path}} = 20$. ASBS uses the same settings with the losses of \cite{liu2025adjoint}, i.e., adjoint matching at a single time $t \sim \mathcal{U}(0, T)$ and corrector matching at $t = T$ only, without importance weights and with $\gamma_i = 1$. For both methods, samples with D-chirality are discarded before they enter the buffers, and $\nabla \mathcal{E}(X_T)$ in the regression target is clipped to norm $100$. We report one training run (seed $0$) per method. Compared with the alanine dipeptide setup of \cite{liu2025adjoint}, which uses the prior $\mathcal{N}(0, 0.25\,I)$, $I = 15$, $N_{\mathrm{FC}} = 4{,}000$, $N_{\mathrm{BC}} = 2{,}000$ ($90{,}000$ epochs), $1{,}000$ new samples per epoch, and a buffer of $10^4$ samples, we use a wider prior that also covers C7ax, more and shorter stages, and more new samples per epoch and larger buffers on four GPUs. The target, the coordinates, the network, the noise schedule, and the evaluation metrics are the same.

\paragraph{Modifications of TR-SBS.} On alanine dipeptide, we use two modifications. First, the step size is bounded from below, $\gamma_i \ge 0.05$. Second, in \eqref{eq: CE matching}, we replace the anchor $u_i(X_T, T)$ by $u_i(X_t, t)$, the previous control at the same time and state as the regressed control. In \eqref{eq: CE matching}, the previous control enters the regression target only through its terminal values $u_i(X_T, T)$, so that its values at intermediate times and states are determined by the regression as conditional expectations given $X_t$. For small $\gamma_i$, this term dominates the target, as its weight is $\lambda_i/(1+\lambda_i) = 1 - \gamma_i$, whereas the update itself is of order $\gamma_i$. Estimation errors of these conditional expectations, due to finite samples or to a learned control that is not exactly consistent with its own terminal values, then enter every update with a weight close to one and can accumulate over many small steps. With the modified anchor, the previous control enters the regression target at the time and state of the regressed control, so that only the term of order $\gamma_i$ has to be estimated. For $\lambda_i = 0$ (ASBS), the anchor drops out, so ASBS is unaffected; for large $\lambda_i$, the regression target approaches the previous control at every time, so that small steps leave the control unchanged. \Cref{lem: anchor} (\Cref{app: anchor}) characterizes the resulting update: its minimizer is a convex combination of the previous control and the importance-weighted adjoint matching target, it agrees with the minimizer of \eqref{eq: CE matching} for $\gamma_i = 1$, and the Schr\"odinger bridge remains a fixed point. Away from the Schr\"odinger bridge, however, the modified loss is not exactly the cross-entropy loss, and our analysis in \Cref{sec: convergence} applies to the unmodified loss.

\subsubsection{Evaluation metrics}
Following \cite{liu2025adjoint}, we report
\begin{itemize}
    \item the KL divergences $\KL(p_{\mathrm{ref}} | p_{\mathrm{model}})$ between the marginal histograms ($200$ bins on $[-\pi, \pi]$, with $10^{-5}$ added to the bin probabilities) of the backbone torsions $\phi$ and $\psi$ and of the remaining torsions $\gamma_1$, $\gamma_2$, $\gamma_3$,
    \item $W_2$ between $10^4$ model and $10^4$ reference samples of $(\phi, \psi)$, with the Euclidean distance between the angles.
\end{itemize}
In addition, we report the populations of the rare conformations C7ax and $\alpha_L$, i.e., the fraction of all samples that have L-chirality and satisfy $0 \le \phi \le 2$ and $\psi < 0$ (C7ax) or $\psi \ge 0$ ($\alpha_L$). The reference populations are $0.169\%$ and $0.049\%$. Except for $W_2$, all model metrics use $10^7$ samples. Except for ASBS after its divergence ($19\%$), at most $0.022\%$ of the samples of all reported checkpoints have D-chirality.
\endgroup

\subsection{Implementation details and hyperparameters}
\label{app:implementation}
The code is available at \url{https://github.com/take-koshizuka/tr_sbs}.

\begin{figure}[t]
\begin{algorithm}[H]
\caption{\small Trust Region Schrödinger Bridge Sampler (TR-SBS), overview (see \Cref{alg:tr-sbs-detail} for details).}
\label{alg: TR-SBS high level}
\small
\begin{algorithmic}[1]
\Require Initial controls $u_0, v_0 = \mathbf{0}$, number of iterations $I$
\For{$i = 0, \ldots, I-1$}

\State \parbox[t]{\dimexpr\linewidth-\algorithmicindent}{Sample trajectories $X^{u_i} \sim \vec{\P}^{u_i}$ by integrating the forward SDE~\eqref{eq: forward SDE}, and compute terminal controls $u_i(X_T^{u_i}, T)$ and the Radon-Nikodym derivative $w = \frac{\mathrm d \cev{\P}^{v_i}}{\mathrm d \vec{\P}^{u_i}}(X^{u_i})$ as in~\eqref{eq: importance weights u_i v_i}.}

\State Construct buffer $\mathcal{B} = \{ (X^{u_i}_0, X^{u_i}_T), u_i(X_T^{u_i},T), w \}$.

\State \parbox[t]{\dimexpr\linewidth-\algorithmicindent}{Compute the Lagrange multiplier $\lambda_i = \argmax_{\lambda \in \mathbb{R}_+} \mathcal{L}^{\mathrm{Dual}}_{i}(\lambda)$ via~\eqref{eq: dual function}, estimated with $N_\mathrm{path}$ trajectories $X^{u_i} \sim \vec{\P}^{u_i}(\cdot \,|\, X_0)$ for each of $N_\mathrm{state}$ initial states $X_0 \sim p_\mathrm{prior}$.}

\State \parbox[t]{\dimexpr\linewidth-\algorithmicindent}{Compute $u_{i+1} = \argmin_u \mathcal{L}_i^{\mathrm{F}}(u)$ using~\eqref{eq: CE matching}, buffer $\mathcal{B}$, and the weights $\frac{\mathrm d \vec{\P}^{u_{i+1}}}{\mathrm d \vec{\P}^{u_i}} \propto w^{\frac{1}{1+\lambda_i}}$ as in~\eqref{eq:annealing}.}

\State Compute $v_{i+1} = \argmin_v \mathcal{L}^\mathrm{B}_i(v)$ via~\eqref{eq: loss for v} with endpoints $(X_0, X_T) \sim \vec{\P}^{u_{i+1}}_{0,T}$.

\EndFor
\State \Return control $u_I$ solving the Schrödinger Bridge
\end{algorithmic}
\end{algorithm}
\end{figure}

\begin{algorithm}
\caption{Trust Region Schr\"odinger Bridge Sampler (TR-SBS)}
\label{alg:tr-sbs-detail}
\begin{algorithmic}[1]
\Require Sample-able source $X_0\!\sim\!p_{\mathrm{prior}}$, differentiable target log density $\log p_{\mathrm{target}}(x)$ \textit{(with $\mathcal{E}(x)=-\log p_{\mathrm{target}}(x)$)}, neural networks for forward control $u$ and backward control $v$.
\Require Replay buffers $\mathcal B_u,\mathcal B_v$ with max capacity $N_{\mathrm{buf}}$, number of new samples per iteration $N_{\mathrm{new}}$.
\Require Number of outer loop $I$, numbers of inner epochs $N_{\mathrm{FC}},N_{\mathrm{BC}}$, batch size $B$, number of gradient steps $L$.
\Require A uniform time grid of $[0, T]$ given by $\mathcal{T}=\{t_0,\ldots, t_{N_t}\}$. 
\State Initialize $u_0, v_0 \coloneqq 0$, \quad $\mathcal B_u\leftarrow\varnothing$, \quad $\mathcal B_v\leftarrow\varnothing$.
\State \textbf{for} $i=0,1,\ldots,I-1$ \textbf{do} \Comment{outer-loop index}
\end{algorithmic}

\makeatletter
\begin{tcolorbox}[highlightbox, left=0.5em]
\begin{algorithmic}[1]
\setcounter{ALG@line}{2}
\State \textbf{Optimize multiplier parameter $\lambda$:} \Comment{$\triangleright$ Solving dual problem}
\State Sample $N_{\mathrm{state}}$ initial states $X_0 \sim p_{\mathrm{prior}}$.
\State For each $X_0$, simulate $N_{\mathrm{path}}$ trajectories $\{X_t^{u_{i}}\}_{t \in \mathcal{T}}$ under the current control $u_i$.
\State Solve for $\lambda_i = \argmax_{\lambda > 0} \widehat{\mathcal{L}}^{\mathrm{Dual}}_i(\lambda)$  via Brent (scipy.optimize) using the Monte Carlo estimate of:
\Statex
\begin{equation*}
    \widehat{\mathcal{L}}^{\mathrm{Dual}}_i(\lambda) =
    (1+\lambda)\,\mathbb{E}_{X_0}\!\left[-\log \mathbb{E}_{X_t^{u_{i}}|X_0}\!\left[\left(\frac{p_{\mathrm{prior}}(X_0)\prod_{n=0}^{N_t-1}p(X^{u_i}_{t_{n+1}}\mid X^{u_i}_{t_{n}})}{p_{\mathrm{target}}(X^{u_i}_T)\prod_{n=0}^{N_t-1}q(X^{u_i}_{t_{n}}\mid X^{u_i}_{t_{n+1}})}\right)^{-\tfrac{1}{1+\lambda}}\right]\right]-\lambda\varepsilon
\end{equation*}
\Statex where $p$ and $q$ are forward and backward transition kernels, respectively.
\end{algorithmic}
\end{tcolorbox}

\begin{tcolorbox}[highlightbox, left=0.5em]
\begin{algorithmic}[1]
\setcounter{ALG@line}{5}
\State \textbf{Forward control (FC) update:}
\State Initialize $u \leftarrow u_{i}$.
\For{$j = 1, 2, \ldots, N_{\mathrm{FC}}$} \Comment{$\triangleright$ Forward Control (FC) matching}
\State Update replay buffer $\mathcal B_u \leftarrow \textsc{PopulateBuffer}(u_i, v_i, \mathcal B_u, N_{\mathrm{new}}, N_{\mathrm{buf}})$.
\State Sample a mini-batch $\{(X_0^{(k)}, X_T^{(k)}, w^{(k)})\}_{k=1}^B$ from $\mathcal{B}_u$. 
\State Sample time $t \sim \mathcal{U}(0, T)$ and bridge state $X_t^{(k)} \sim p^{\mathrm{ref}}_{t|0,T}(\cdot|X_0^{(k)}, X_T^{(k)})$ for each pair $(X_0^{(k)}, X_T^{(k)})$. 
\State Take $L$ gradient steps with respect to the FC objective $\mathcal{L}^{\mathrm{F}}_i(u)$ (\cref{eq: CE matching}), with weights $w^{\frac{1}{1+\lambda_i}}$ normalized and clipped as described in \Cref{app: weight normalization}:
\Statex
\begin{align}
    \mathcal{L}_i^{\mathrm{F}}(u) &= \mathbb{E}_{X_{0, T}^{(k)} \sim \mathcal{B}_u} \sum_{\tau \in \{ 0, t, T\}}\mathbb{E}_{X_{\tau}^{(k)} \mid X_{0, T}^{(k)} } \Bigg[ w^{\frac{1}{1+\lambda_i}}\bigg\| u(X_{\tau}^{(k)}, \tau) - \sigma(\tau)\sigma(T)^{-1}\, m_i^*(X^{(k)}) \bigg\|^2 \Bigg],\\
     m_i^*(X^{(k)}) &\coloneqq -\frac{1}{1 + \lambda_i} \left(v_i(X^{(k)}_T, T) + \sigma(T)^{\top} \nabla \mathcal{E}(X^{(k)}_T)\right) + \frac{\lambda_i}{1 + \lambda_i} u_i^{(k)}(X^{(k)}_T, T). 
\end{align}
\EndFor
\State Update $u_{i+1} \leftarrow u$.
\end{algorithmic}
\end{tcolorbox}
\begin{tcolorbox}[highlightbox, left=0.5em]
\begin{algorithmic}[1]
\setcounter{ALG@line}{15}
\State \textbf{Backward control (BC) update:}
\State Initialize $v \leftarrow v_{i}$.
\For{$j = 1, 2, \ldots, N_{\mathrm{BC}}$} \Comment{$\triangleright$ Backward Control (BC) matching}
    \State Update replay buffer $\mathcal B_v \leftarrow \textsc{PopulateBuffer}(u_{i+1}, v_i, \mathcal B_v, N_{\mathrm{new}}, N_{\mathrm{buf}})$. 
    \State Sample a mini-batch $\{(X_0^{(k)}, X_T^{(k)}, \cdot)\}_{k=1}^B$ from $\mathcal{B}_v$ and discard unused fields.
    \State Sample time $t \sim \mathcal{U}(0, T)$ and bridge state
$X_t^{(k)} \sim p^{\mathrm{ref}}_{t|0,T}(\cdot \mid X_0^{(k)}, X_T^{(k)})$ for each pair $(X_0^{(k)}, X_T^{(k)})$.
    \State Minimize the BC objective $\mathcal{L}^{\mathrm{B}}_i(v)$ (with $\zeta_\tau = (1+\sigma^2(1-\tau)/\tau)^{-1}$):
    \Statex \quad
    \begin{align*}
        \mathcal{L}^{\mathrm{B}}_i(v)
        &= \mathbb{E}_{(X_0^{(k)},X_T^{(k)}) \sim \mathcal{B}_v}
        \sum_{\tau \in \{t, T\}}
        \mathbb{E}_{X_{\tau}^{(k)} \mid X_{0,T}^{(k)}}
        \Bigg[
            \zeta_{\tau}\Big\|
                v(X_{\tau}^{(k)}, \tau)
                -  \sigma(\tau)^{\top} \nabla \log p^{\mathrm{ref}}_{\tau|0}(X_{\tau}^{(k)} \mid X_0^{(k)})
            \Big\|^2
        \Bigg]
    \end{align*}
\EndFor
\State Update $v_{i+1} \leftarrow v$.
\end{algorithmic}
\end{tcolorbox}

\makeatother
\begin{algorithmic}[1]
\setcounter{ALG@line}{22}
\State \textbf{end for}
\Ensure Kinetic-optimal drift $u^* \approx u_{I}$
\end{algorithmic}
\end{algorithm}

\begin{algorithm}
\caption{PopulateBuffer: Auxiliary routine for replay buffer construction}
\label{alg:populate-buffer}
\begin{algorithmic}[1]
\State \textbf{Routine} $\textsc{PopulateBuffer}(u_i, v_i, \mathcal B, N_{\mathrm{new}}, N_{\mathrm{buf}})$
\State Sample $N_{\mathrm{new}}$ trajectories $(X_0, \dots, X_T^{u_i})$ from the current model $u_i$.
\For{$\ell = 1, \ldots, N_{\mathrm{new}}$}
    \State Compute $w^{(\ell)} = w(X^{u_i,(\ell)})$ via path-integral:
    \Statex \quad
    \begin{equation}
        \log w^{(\ell)}
        = \log \frac{\mathrm{d}\cev{\P}^{v_i}}{\mathrm{d}\vec{\P}^{u_i}}(X^{u_i,(\ell)})
        \approx
        -\log\!\left(
        \frac{
            p_{\mathrm{prior}}(X_0^{(\ell)})
            \prod_{n=0}^{N_t-1} p(X^{u_i,(\ell)}_{t_{n+1}}\!\mid X^{u_i,(\ell)}_{t_n})
        }{
            p_{\mathrm{target}}(X^{u_i,(\ell)}_T)
            \prod_{n=0}^{N_t-1} q(X^{u_i,(\ell)}_{t_n}\!\mid X^{u_i,(\ell)}_{t_{n+1}})
        }\right).
    \end{equation}
    \State $\mathcal B \leftarrow \mathcal B \cup \{(X^{(\ell)}_0, X^{(\ell)}_T, w^{(\ell)})\}$.
\EndFor
\State \textbf{if} $|\mathcal B| > N_{\mathrm{buf}}$ \textbf{then}
\State \quad Remove the oldest $|\mathcal B| - N_{\mathrm{buf}}$ elements from $\mathcal B$ (FIFO).
\State \textbf{end if}
\State \textbf{return} $\mathcal B$.
\end{algorithmic}
\end{algorithm}

In this section, we detail the implementation of TR-SBS, covering the underlying SDEs, numerical integration schemes, network architectures, and data management strategies. The detailed pseudocode is provided in~\cref{alg:tr-sbs-detail}. All experiments were conducted on an internal computing cluster. The training and evaluation for each configuration were performed using a single NVIDIA A40 GPU with 48GB of VRAM, except for the training on alanine dipeptide, which used four (\Cref{app:aldp_detail}).

\subsubsection{Geometric noise schedule and SDE formulation}
\label{sec:noise_schedule}

We adopt the Variance Exploding (VE) SDE framework \citep{song2020score} as our reference dynamics, where the base drift is zero ($f(x,t) := 0$). The time-dependent diffusion coefficient $\sigma(t)$ is that of the VE SDE with the geometric noise levels $\sigma_{\min} (\sigma_{\max}/\sigma_{\min})^{s}$, $s \in [0, 1]$, of \cite{song2020score}, run in reverse time $s = 1 - t$, so that the noise is largest at the prior ($t = 0$) and smallest at the target ($t = 1$):
\begin{equation}
    \sigma(t) \coloneqq \sigma_{\min} \left( \frac{\sigma_{\max}}{\sigma_{\min}} \right)^{1-t} \sqrt{2 \ln \frac{\sigma_{\max}}{\sigma_{\min}}}.
\end{equation}

\paragraph{Forward and backward dynamics}
The Schrödinger bridge problem is formulated as finding a control policy $u$ that minimizes transport cost while satisfying boundary conditions. The forward process $X^u_t$ and the corresponding reverse-time backward process $Y^v_t$ are governed by:
\begin{align}
    \label{eq:forward_sde}
    \mathrm{d} X^u_t &= \sigma(t) u(X^u_t,t)\mathrm{d} t + \sigma(t) \vec{\mathrm{d}} W_t, \qquad & X^u_0 \sim p_{\mathrm{prior}}, \\
    \label{eq:backward_sde}
    \mathrm{d} Y^v_t &= -\sigma(t) v(Y^v_t,t)\mathrm{d} t + \sigma(t) \cev{\mathrm{d}} W_t, \qquad & Y^v_T \sim p_{\mathrm{target}},
\end{align}
where $u \in \mathcal{U}$ is the forward control and $v$ is the backward control. The objective is to minimize the KL divergence with respect to the reference path measure (where $u \equiv 0$). 
Except for alanine dipeptide (\Cref{app:aldp_detail}) and LJ-13 (\Cref{app: lj13}), we used a zero-mean Gaussian distribution as the prior distribution $p_{\mathrm{prior}} = \mathcal{N}(\mathbf{0}, \sigma_{\mathrm{prior}}^2 \mathbf{I})$.

\paragraph{Parameterization}
We parameterize the controls using neural networks $u_\theta$ and $v_\phi$. To stabilize training and align with the scale of the diffusion coefficient, we scale the network outputs by $\sigma(t)$:
\begin{align}
    u(x, t) &= \sigma(t) u_\theta(x, t), \\
    v(x, t) &= \sigma(t) v_\phi(x, t).
\end{align}
Substituting these into the SDEs, the effective drift terms become $\sigma(t)^2 u_\theta(X_t, t)$ and $-\sigma(t)^2 v_\phi(Y_t, t)$, ensuring that the learnable drift magnitudes remain comparable to the noise level throughout the process.

\subsubsection{Numerical integration and trajectory weighting}

To ensure numerical stability and accuracy, especially in regimes with limited discretization steps, we employ an exact integration scheme for the noise variance rather than a simple Euler-Maruyama approximation.

\paragraph{Exact integration of diffusion variance}
For a discretization grid $\mathcal{T}=\{t_0, \ldots, t_{N_t}\}$, we compute the exact accumulated variance $\widebar{\sigma}_k^2$ over each interval $[t_k, t_{k+1}]$ analytically:
\begin{equation}
    \widebar{\sigma}_k^2 \coloneqq \int_{t_k}^{t_{k+1}} \sigma_s^2 \, ds = \sigma_{\max}^2 \left[ \left( \frac{\sigma_{\min}}{\sigma_{\max}} \right)^{2t_k} - \left( \frac{\sigma_{\min}}{\sigma_{\max}} \right)^{2t_{k+1}} \right].
\end{equation}
This formulation captures the precise energy of the noise injected into the system, avoiding discretization errors associated with approximating $\sigma_t^2 \Delta t$ when $\sigma_t$ varies rapidly.

\paragraph{State update rule}
Using the calculated variance $\widebar{\sigma}_k^2$, the forward state update follows a modified Euler-Maruyama scheme:
\begin{equation}
    X_{t_{k+1}} = X_{t_k} + \sigma(t_k) u(X_{t_k}, t_k)\Delta t_k + \widebar{\sigma}_k \cdot \varepsilon_k, \quad \varepsilon_k \sim \mathcal{N}(0, I),
\end{equation}
where $\Delta t_k = t_{k+1} - t_k$.
This ensures that the stochastic term remains consistent with the underlying continuous-time process.

\paragraph{Computation of importance weights}
The computation of the Radon-Nikodym derivative $w(\mathbf{X})$ is crucial for the trust region update. Consistent with Algorithm \ref{alg:populate-buffer}, the log-weight for a trajectory $\mathbf{X} = (X_{t_0}, \dots, X_{t_{N_t}})$ is given by:
\begin{equation}
    \log w(\mathbf{X}) \approx \log p_{\mathrm{target}}(X_T) + \sum_{k=0}^{N_t-1} \log q(X_{t_k} \mid X_{t_{k+1}}) - \log p_{\mathrm{prior}}(X_0) - \sum_{k=0}^{N_t-1} \log p(X_{t_{k+1}} \mid X_{t_k}).
\end{equation}
Here, the transition probabilities $p$ (forward) and $q$ (backward) are defined by the Gaussian transitions derived from the exact variance integration:
\begin{align}
    \log p(X_{t_{k+1}} \mid X_{t_k}) &= -\frac{1}{2\widebar{\sigma}_k^2} \big\| X_{t_{k+1}} - X_{t_k} - \sigma(t_k) u(X_{t_k}, t_k)\Delta t_k \big\|^2 + Z_k, \\
    \log q(X_{t_k} \mid X_{t_{k+1}}) &= -\frac{1}{2\widebar{\sigma}_k^2} \big\| X_{t_k} - X_{t_{k+1}} - \sigma(t_{k+1}) v(X_{t_{k+1}}, t_{k+1})\Delta t_k \big\|^2 + Z_k.
\end{align}
Note that the normalization constant $Z_k$ cancels out in the likelihood ratio.

\paragraph{Normalization and clipping of the importance weights}
\label{app: weight normalization}
In the forward matching loss, we use the tempered weights $w^{\frac{1}{1+\lambda_i}}$. For numerical stability, the tempered log-weights are shifted such that the weights have unit mean over each training mini-batch, and the resulting weights are clipped at a maximum value $c_\mathrm{IW}$ (\Cref{tab:all_hyperparams}). The batch normalization replaces the normalization $\E_{\vec{\P}^{u_i}}[w^{\frac{1}{1+\lambda_i}} \,|\, X_0]$ in \Cref{prop: geometric interpretation of update}, which cannot be estimated from a buffer that contains a single trajectory per initial state; the dual objective in \Cref{alg:tr-sbs-detail}, in contrast, uses $N_\mathrm{path}$ trajectories per initial state and estimates the conditional expectation directly. In addition, our implementation multiplies the log-weights $\log w$ of each batch of $N_{\mathrm{new}}$ new trajectories by $r = \min\{|\widehat{V}_0| / |\overline{\log w}|, 1\} \in (0, 1]$ before they enter the buffer, where $\overline{\log w}$ is their mean and $\widehat{V}_0$ is the Monte Carlo estimate of $\mathcal{L}^{\mathrm{Dual}}_i(0)$ in \Cref{alg:tr-sbs-detail}, both computed with the unnormalized target. This heuristic additionally tempers the weights in the loss to $w^{r/(1+\lambda_i)}$. On alanine dipeptide, $r < 0.1$ in six of the first seven of the $40$ outer iterations, so that the weights are almost uniform there, and $r \approx 0.99$ afterwards.

\subsubsection{Network architecture}
We parameterize both $u_\theta(t, x)$ and $v_\phi(t, x)$ using a \textbf{Time-Conditional MLP} designed to handle high-dimensional inputs.
\begin{itemize}
    \item \textbf{Embedding:} The time $t$ is mapped via \textit{sinusoidal embeddings} with frequency components, while the input state $x$ is projected using a learnable linear layer. These embeddings are summed to form the input to the backbone.
    \item \textbf{Backbone:} A 4-layer Multilayer Perceptron (MLP) with a hidden dimension of 256 and \textbf{GELU} activations ($8$ layers on alanine dipeptide, \Cref{app:aldp_detail}).
    \item \textbf{Output clipping:} To ensure numerical stability, each component of the network output is clipped to $[-100, 100]$.
\end{itemize}

\subsubsection{Replay buffers and optimization}
\label{app:replay_buffer_details}

\paragraph{Buffer management}
In our implementation, we maintain separate replay buffers, $\mathcal{B}_u$ and $\mathcal{B}_v$, dedicated to the Forward Control (FC) and Backward Control (BC) updates, respectively. Additionally, the optimization of the Lagrange multiplier $\lambda$ (the dual problem) is performed using a fresh set of $N_{\mathrm{state}} \times N_{\mathrm{path}}$ trajectories, starting from $N_{\mathrm{state}}$ initial states sampled from the prior, at each outer iteration.
We adopted a First-In-First-Out (FIFO) update rule where only a portion of the buffer ($N_{\mathrm{new}}$ samples) is refreshed at each iteration. As noted in \cite{liu2025adjoint}, this mechanism renders the algorithm a \textit{hybrid method between on-policy and off-policy} training.

This design choice allows for a fair comparison with the Adjoint Schrödinger Bridge Sampler (ASBS) \citep{liu2025adjoint}. However, we emphasize that our TR-SBS formulation is inherently more amenable to off-policy configurations due to the explicit incorporation of importance sampling weights and the trust region constraint, which theoretically allows for reduced bias when reusing samples from previous policies.

\paragraph{Optimization settings}
We use the Adam optimizer with a constant learning rate of $1 \times 10^{-4}$ for training both control networks across all tasks.

\Cref{tab:all_hyperparams} provides a unified summary of the hyperparameters across all five experimental settings. To ensure a fair comparison, we evaluate the ASBS-family baselines under the same settings, except for the diffusion scale $\sigma_{\max}$ on GMM, which was chosen per method.
\begin{table}[h]
    \centering
    \begin{threeparttable}
        \caption{Unified hyperparameters for the GMM, MW, and GL experiments (alanine dipeptide: \Cref{app:aldp_detail}; LJ-13: \Cref{app: lj13}). For MW, values separated by a slash (e.g., $0.1/0.2$) indicate that both values were evaluated for each specific target configuration, and the one yielding better performance was selected.}
        \label{tab:all_hyperparams}
        
        \small 
        \begin{tabularx}{\textwidth}{l *{5}{>{\centering\arraybackslash}X}}
            \toprule
            \textbf{Parameter} & \makecell{\textbf{GMM} \\ \textbf{(Scale $d$)}} & \makecell{\textbf{GMM} \\ \textbf{(Robust $K$)}} & \makecell{\textbf{MW} \\ \textbf{(Scale $m$)}} & \makecell{\textbf{MW} \\ \textbf{(Robust $\delta$)}} & \makecell{\textbf{GL} \\ \textbf{(Periodic)}} \\
            \midrule
            $\sigma_{\text{prior}}$ & $5.0$ & $30.0$ & $1.0$ & $1.0$ & $1.0$ \\
            $\sigma_{\mathrm{max}}$ & $1/2/1/2^{\ddagger}$ & $1/2/2/4^{\ddagger}$ & $0.2$ & $0.1/0.2$ & $0.2$ \\
            $\sigma_{\mathrm{min}}$ & $0.1$ & $0.1$ & $0.01$ & $0.01$ & $0.01$ \\
            $\varepsilon$ & $0.1$ & $0.3$ & $0.3$ & $0.3$ & $0.3$ \\
            $c_{\mathrm{IW}}$ & $10$ & $10$ & $1$ & $1$ & $1$ \\
            \midrule
            $I$ & $\le 150^{\dagger}$ & $\le 150^{\dagger}$ & $50$ & $35$ & $80$ \\
            $N_{\mathrm{FC}}$ & $150$ & $150$ & $200$ & $100$ & $150$ \\
            $N_{\mathrm{BC}}$ & $50$ & $50$ & $50$ & $100$ & $150$ \\
            $L$ & $100$ & $100$ & $100$ & $100$ & $100$ \\
            $N_t$ & $200$ & $200$ & $200$ & $200$ & $200$ \\
            \midrule
            $B$ & $1000$ & $1000$ & $1000$ & $1000$ & $1000$ \\
            $N_{\text{buf}}$ & $50000$ & $50000$ & $50000$ & $50000$ & $50000$ \\
            $N_{\mathrm{new}}$ & $10000$ & $10000$ & $10000$ & $10000$ & $10000$ \\
            $N_{\mathrm{state}}$ & $5000$ & $5000$ & $5000$ & $5000$ & $5000$ \\
            $N_{\mathrm{path}}$ & $40$ & $40$ & $40$ & $40$ & $40$ \\
            \bottomrule
        \end{tabularx}
        
        \begin{tablenotes}
            \footnotesize
            \item[$\dagger$] Maximum number of outer iterations; the reported checkpoint is selected as described in \Cref{app:gmm_detail}.
            \item[$\ddagger$] For ASBS / TR-ASBS / ASBS-IW / TR-SBS, respectively.
        \end{tablenotes}
    \end{threeparttable}
\end{table}

\subsection{PIS and DIS baselines}
\label{app: pis dis}
We compare with the Path Integral Sampler (PIS) \cite{zhang2022path} and the time-reversed diffusion sampler (DIS) \cite{berner2022optimal} on GMM. Both are trained by minimizing the reverse KL divergence (ELBO) on path space. Their control network uses the same architecture as the forward control of the ASBS-family methods (4 layers, width 256, output norm clipped at $100$) and does not take the score $\nabla \log p_\mathrm{target}$ as an input, so it has the same number of parameters as the ASBS controller at a given dimension. PIS starts from the zero initial state and DIS from the Gaussian prior of the respective setting. Both use $128$ time steps with a fixed cosine noise schedule with minimum value $0.1$ and maximum value $\sigma_{\max}$, where $\sigma_{\max} = 128$ (PIS) and $2$ (DIS) for the mode suite and $\sigma_{\max} = 32$ (PIS) and $4$ (DIS) for the dimension suite. We train for $40{,}000$ steps with Adam (learning rate $5 \times 10^{-4}$, batch size $1{,}000$, gradient clipping at $1.0$) and use the same model selection and evaluation protocol as for the other methods (\Cref{app:gmm_detail}).

\textbf{Differences in budget.} The ASBS-family methods use $200$ time steps, whereas PIS and DIS use $128$; the numbers of optimizer updates and target evaluations also differ. The comparison in \Cref{fig:gmm_results} therefore compares the completed experiments and does not claim equal total compute. A comparison with matched discretization and budget, as well as PIS and DIS results on MW, is left for future work.

\subsection{Computational cost}
\label{app: computational cost}
\textbf{Evaluations per outer iteration.} In each outer iteration, the forward update populates the buffer with $N_{\mathrm{new}}$ new trajectories in each of its $N_{\mathrm{FC}}$ epochs and evaluates the target gradient $\nabla \mathcal{E}(X_T)$ at their end points. This gives $N_{\mathrm{FC}} N_{\mathrm{new}}$ target-gradient evaluations for both ASBS and TR-SBS. TR-SBS additionally evaluates $\mathcal{E}(X_T)$ for the same trajectories to compute the importance weights, and $N_{\mathrm{state}} N_{\mathrm{path}}$ times for the dual objective, but requires no further gradient evaluations. The backward update does not evaluate the target. With the values in \Cref{tab:all_hyperparams}, e.g., MW with varying $\delta$ uses $N_{\mathrm{FC}} N_{\mathrm{new}} = 100 \times 10{,}000 = 10^6$ target-gradient evaluations and $N_{\mathrm{state}} N_{\mathrm{path}} = 2 \times 10^5$ additional energy evaluations per outer iteration.

\textbf{Number of outer iterations.} The total cost scales as $\mathcal{O}(I N_{\mathrm{FC}} N_{\mathrm{new}})$, and the outer iterations needed to converge differ between the methods. On MW with $d=8$ and $\delta=8$ (\Cref{fig:sigma_energy}), the mean energy indicates that ASBS has essentially converged after about $I=6$ outer iterations, although to a mode-collapsed solution, whereas TR-SBS with $\varepsilon=0.3$ reaches convergence after about $I=32$ outer iterations. TR-SBS thus requires approximately $32/6 \approx 5.3$ times more target-gradient evaluations than ASBS in this setting. \Cref{fig:ablation_gamma_mw} provides a complementary view: larger fixed step sizes converge faster in energy but to a worse solution, while a conservative fixed step size is stable but slow.

\textbf{Hardware.} All GMM runs, including PIS and DIS, use a single NVIDIA A40 GPU; the hardware of the other experiments is described in \Cref{app:implementation}. We do not report wall-clock times, since the runs were executed on a shared cluster. On alanine dipeptide, with the same hardware for both methods (\Cref{app:aldp_detail}), one epoch of TR-SBS took about $1.5$ times as long as one epoch of ASBS.

\section{Additional experimental results}
\label{app: additional experiments}
\subsection{Additional metrics on GMM}
\label{app: gmm extra metrics}
\begin{figure*}[t]
    \centering
    \includegraphics[width=0.9\textwidth]{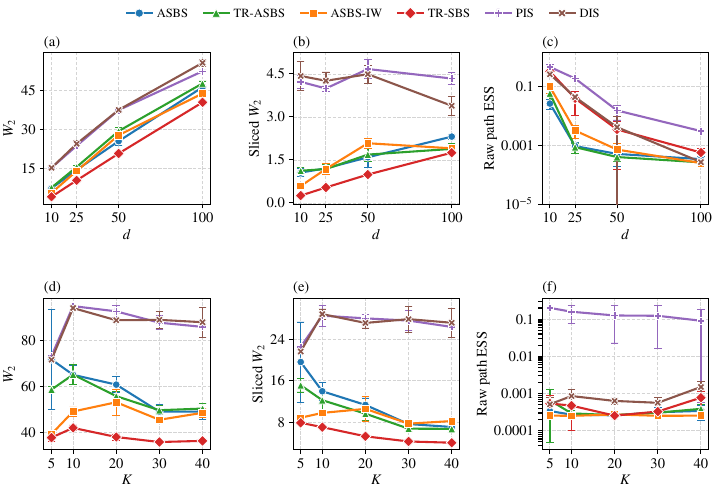}
    \caption{\textbf{Additional metrics on GMM} (mean $\pm$ standard deviation over three seeds). (a)-(c): $K=10$, varying $d$. (d)-(f): $d=10$, varying $K$. Lower is better for $W_2$ (a, d) and sliced $W_2$ (b, e); higher is better for the raw path ESS (c, f).}
    \label{fig:gmm_extra_metrics}
\end{figure*}

\Cref{fig:gmm_extra_metrics} reports further metrics for the GMM experiment of \Cref{sec: gmm}, using the same checkpoints and evaluation protocol (\Cref{app:gmm_detail}). The raw path ESS is the normalized effective sample size of the untempered path weights $\frac{\mathrm d \cev{\P}^{v}}{\mathrm d \vec{\P}^{u}}(X)$, $X \sim \vec{\P}^u$, evaluated with the learned controls.

\subsection{Visualization on GMM ($d=10$, $K=40$)}
\label{app: gmm visualization}
\begin{figure*}[t]
    \centering
    \includegraphics[width=\textwidth]{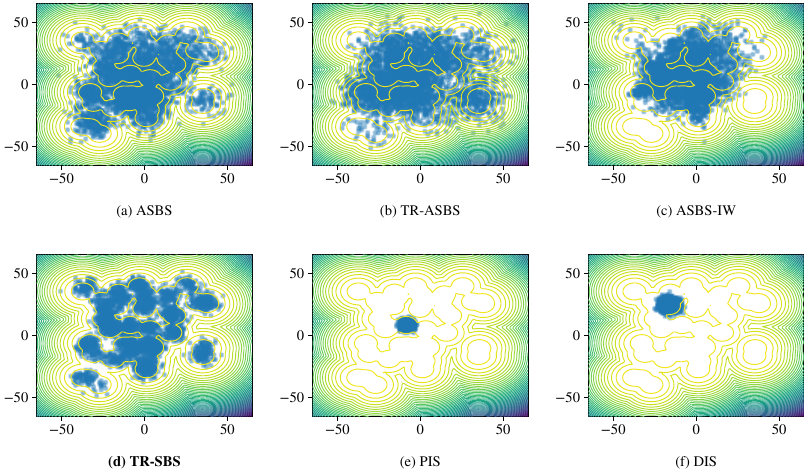}
    \caption{\textbf{Samples on GMM with $d=10$ and $K=40$} (first two coordinates, seed 0). Each panel shows the 4,000 evaluation samples of the checkpoint with the lowest SD during training. Blue points show generated samples; contours show the target marginal log density. All panels use identical contours and coordinate limits and show every sample. PIS and DIS use 128 instead of 200 time steps (\Cref{app: pis dis}).}
    \label{fig:gmm_vis_k40}
\end{figure*}

\Cref{fig:gmm_vis_k40} shows the first two coordinates of the 4,000 evaluation samples of the selected checkpoint (seed 0) for each method in the setting of \Cref{sec: gmm} with $d=10$ and $K=40$, together with contours of the target marginal log density.
PIS and DIS collapse onto a single component. ASBS-IW concentrates on the central components and misses the outlying ones in the lower left and on the right. ASBS and TR-ASBS reach most components, including the outlying ones, but place considerable mass between them and scatter samples into low-density regions. TR-SBS concentrates its samples on the high-density regions, including the outlying components, with clearer gaps between them, which is consistent with its lower SD and TVD in \Cref{fig:gmm_results} (c, d). Since the figure shows a two-dimensional projection of a ten-dimensional target, it complements but does not replace this quantitative comparison.

\subsection{Visualization on high-barrier Many-Well ($d=8, m=5, \delta=8$)}

We provide a detailed visual analysis of the Many-Well potential with a high energy barrier ($\delta=8$) in $d=8$ dimensions.
We compare four methods: ASBS, ASBS-IW, TR-ASBS (Ablation), and TR-SBS (Proposed) under two noise regimes.

\begin{figure*}[t]
    \centering
    \includegraphics{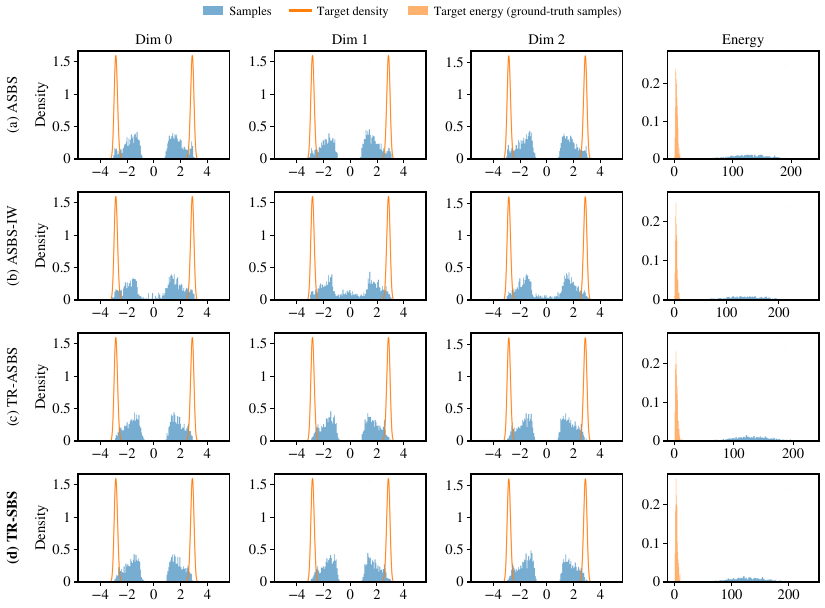}
    \caption{\textbf{Visualization at small noise ($\sigma_{\max}=0.1$).} 
    Rows (a)--(d) show the four methods; columns show the marginal histograms of dimensions 0--2 and the energy distribution.
    Due to insufficient noise, \textbf{all methods} (a-d) fail to reach the wells (underfitting).}
    \label{fig:mw_vis_sigma01}
\end{figure*}

\begin{figure*}[t]
    \centering
    \includegraphics{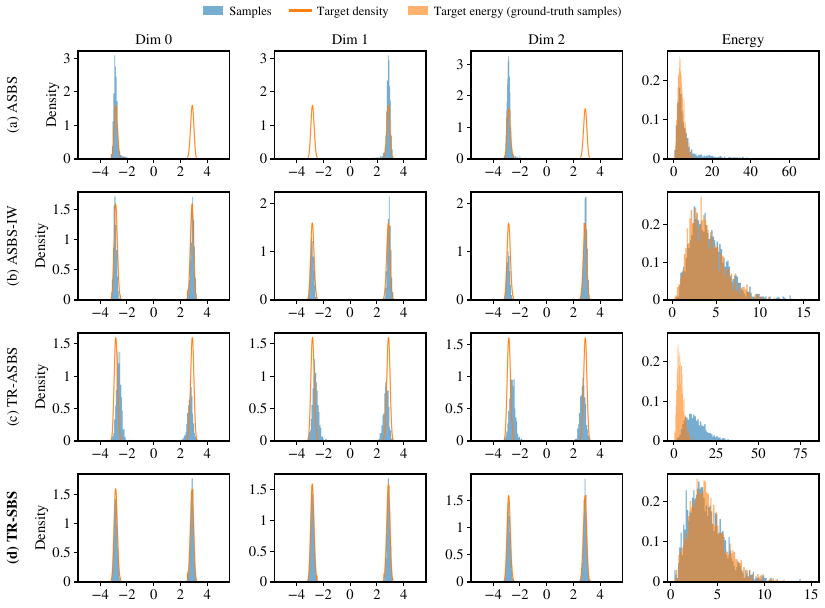}
    \caption{\textbf{Visualization at large noise ($\sigma_{\max}=0.2$).} 
    Rows and columns as in \Cref{fig:mw_vis_sigma01}.
    Aggressive noise facilitates exploration but causes instability.
    \textbf{(a)} ASBS suffers from mode collapse, and \textbf{(b)} ASBS-IW misbalances the modes.
    \textbf{(c)} TR-ASBS prevents total collapse but lacks fidelity.
    \textbf{(d)} Only \textbf{TR-SBS} successfully captures the global structure and energy distribution.}
    \label{fig:mw_vis_sigma02}
\end{figure*}

\begin{itemize}
    \item \textbf{Failure at small noise ($\sigma_{\max}=0.1$):}
    Figure~\ref{fig:mw_vis_sigma01} illustrates the results with a conservative noise scale. 
    Across all methods (a-d), the sampler fails to cross the high energy barrier. The marginal histograms (dim 0-2) show that the samples do not reach the wells at $\pm\sqrt{\delta}$ but remain between them and the barrier, resulting in \textit{underfitting}. The energy distributions (rightmost column) are consistently shifted towards higher values compared to the ground truth, confirming that no method can reach the optimal solution.

    \item \textbf{Success with TR-SBS at Large Noise ($\sigma_{\max}=0.2$):}
    Figure~\ref{fig:mw_vis_sigma02} presents the results under an aggressive noise scale designed to facilitate exploration.
    \begin{itemize}
        \item \textbf{ASBS} (a) suffers from severe mode collapse across almost all dimensions. 
        On the other hand, 
        \item \textbf{ASBS-IW} (b) exhibits \textit{inconsistent recovery across dimensions}. While importance weighting enables it to capture both modes, it misbalances their mass in some dimensions (dim 1 and 2). This suggests that relying solely on importance weighting is insufficient to consistently correct the proposal bias.
        \item \textbf{TR-ASBS} (c) manages to prevent total collapse but lacks fidelity. It fails to fit the distribution sharply. 
        \item \textbf{TR-SBS:} In stark contrast, \textbf{TR-SBS} (d) demonstrates superior performance. The marginal histograms accurately reproduce the symmetric bimodal structure of the ground truth, and the energy distribution overlaps closely with the target. This demonstrates that the combination of the trust region (for stability) and Importance Weighting (for precision) is essential for consistently navigating this challenging landscape.
    \end{itemize}
\end{itemize}

\subsection{Ablation study}
\label{app: sigma ablation}
\textbf{Noise.}\label{sec: step size and cost} Whereas $\varepsilon$ controls the magnitude of each path space update (\Cref{sec: convergence}), the diffusion scale $\sigma_{\max}$ primarily controls exploration and barrier crossing: too little diffusion leads to stagnation, while larger diffusion facilitates barrier crossing but may increase finite-sample instability. The interaction of the two parameters is therefore indirect and problem-dependent.

\begin{figure*}[t]
    \centering
    \setlength{\capshift}{0.175in}
    \includegraphics{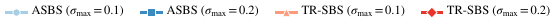}\\[-0.3em]
    \begin{subfigure}{0.48\textwidth}
        \includegraphics[width=\linewidth]{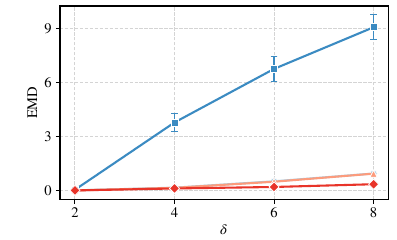}
        \shiftedcaption{EMD (vs. $\delta$)\label{fig:delta_sigma_emd}}
    \end{subfigure}
    \hfill
    \begin{subfigure}{0.48\textwidth}
        \includegraphics[width=\linewidth]{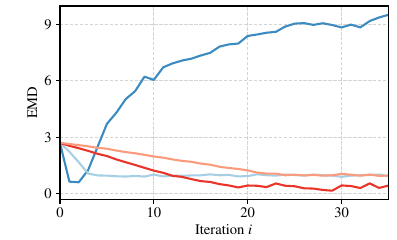}
        \shiftedcaption{EMD (vs. $i$)\label{fig:sigma_emd_iter}}
    \end{subfigure}

    \vspace{0.6em}
    \begin{subfigure}[b]{0.48\textwidth}
        \includegraphics[width=\linewidth]{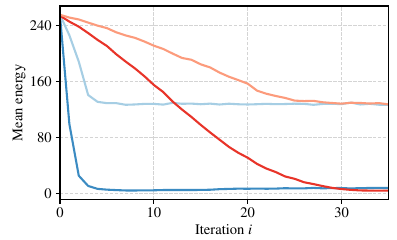}
        \shiftedcaption{Mean energy (vs. $i$)\label{fig:sigma_energy}}
    \end{subfigure}
    \hfill
    \begin{subfigure}[b]{0.48\textwidth}
        \includegraphics[width=\linewidth]{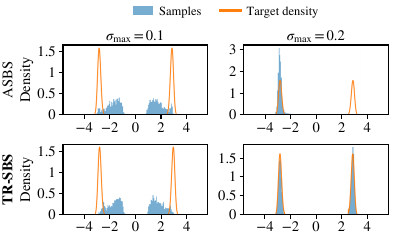}
        \shiftedcaption{Visualization (dim 0)\label{fig:sigma_vis}}
    \end{subfigure}
    
    \caption{\textbf{Ablation study on the maximum noise scale $\sigma_{\max}$.} 
    Legend is shared across subplots (a-c).
    \textbf{(a)} Large $\sigma_{\max}=0.2$ causes mode collapse in ASBS (blue) as barrier distance $\delta$ increases, whereas \textbf{TR-SBS} (red) remains robust.
    \textbf{(b)} Training dynamics confirm that ASBS diverges in EMD while TR-SBS remains stable.
    \textbf{(c)} A larger $\sigma_{\max}$ leads to faster convergence.
    \textbf{(d)} Histograms confirm that ASBS collapses to a single mode under large noise, while TR-SBS maintains the global structure.}
    \label{fig:ablation_sigma_max}
\end{figure*}

We perform ablation studies to validate the impact of key hyperparameters and demonstrate the efficacy of the proposed trust region mechanism in managing the trade-off between efficiency and stability. In addition to the sample quality metrics utilized in the preceding subsection, we introduce the sample mean of the energy, $\mathbb{E}[\mathcal{E}(X_T)]$, to measure the convergence speed (specifically, the required number of iterations $I$) of the IPF update.

\paragraph{Impact of the maximum noise scale ($\sigma_{\max}$)}

A larger $\sigma_{\max}$ yields a smoother optimization landscape, making the problem more tractable and significantly accelerating convergence.
However, in the context of IPF which approximates the objective using finite samples, an excessively large $\sigma_{\max}$ induces mode collapse.
We demonstrate that \textbf{TR-SBS} allows for aggressive smoothing (large $\sigma_{\max}$) without sacrificing mode diversity. We analyze this trade-off using the MW potential ($d=8$), analyzing both robustness to mode distance ($\delta$) and training dynamics ($\delta=8$). \Cref{fig:ablation_sigma_max} presents the results.

\emph{Stability:} While a large $\sigma_{\max}$ aids optimization speed, it introduces instability in the finite-sample regime of standard ASBS, triggering severe mode collapse.
\Cref{fig:delta_sigma_emd} shows that for ASBS with $\sigma_{\max}=0.2$ (blue line), the Marginal EMD explodes as the task difficulty increases ($\delta \to 8$).
This collapse is further evidenced in the training dynamics in \Cref{fig:sigma_emd_iter}, where the error for ASBS rises uncontrollably.
In contrast, \textbf{TR-SBS} with $\sigma_{\max}=0.2$ (red line) maintains a low EMD comparable to conservative settings. \Cref{fig:sigma_vis} shows a histogram of the first coordinate slice of MW. Under large noise ($\sigma_{\max}=0.2$), standard ASBS (top right) collapses to a single mode. In contrast, TR-SBS (bottom right) successfully captures the bimodal structure even with the aggressive noise scale, thanks to the trust region constraint.

\emph{Efficiency vs. Stability:} \Cref{fig:sigma_energy} tracks the averaged energy during training. When $\sigma_{\max}$ is small ($0.1$), both methods stagnate at a high-energy plateau, trapped in local optima. Increasing $\sigma_{\max}$ accelerates the convergence of both, reaching lower energy states by smoothing the landscape.

\paragraph{Sensitivity to the numbers of inner epochs ($N_{\mathrm{FC}}$, $N_{\mathrm{BC}}$)}
As inherited from ASBS, the numbers of forward and backward matching epochs per outer iteration are important hyperparameters. \Cref{tab:nfc_nbc} reports TR-SBS on MW ($d=8$, $m=5$, $\delta=4$) for different values. Too few epochs in either update degrade the result, while too many epochs increase the cost and can overfit; in particular, the error of the log-normalizing constant deteriorates markedly away from $N_{\mathrm{FC}} = N_{\mathrm{BC}} = 100$.

\begin{table}[h]
    \centering
    \small
    \caption{Sensitivity of TR-SBS to $N_{\mathrm{FC}}$ and $N_{\mathrm{BC}}$ on MW ($d=8$, $m=5$, $\delta=4$). The setting $100/100$ is the main result averaged over three seeds; the other settings use a single seed.}
    \label{tab:nfc_nbc}
    \begin{tabular}{cccc}
        \toprule
        $N_{\mathrm{FC}}$ & $N_{\mathrm{BC}}$ & Marginal EMD $\downarrow$ & $|\Delta \log Z|$ $\downarrow$ \\
        \midrule
        $10$ & $100$ & $0.294$ & $200.694$ \\
        $100$ & $10$ & $0.215$ & $2.455$ \\
        $100$ & $100$ & $\mathbf{0.131}$ & $\mathbf{0.006}$ \\
        $100$ & $300$ & $0.191$ & $546.959$ \\
        $300$ & $100$ & $0.212$ & $99.551$ \\
        \bottomrule
    \end{tabular}
\end{table}

\subsection{$\phi^4$ scalar field theory}
\label{sec: Ginzburg-Landau}
We evaluate our approach on the discretized Ginzburg-Landau model ($\phi^4$ scalar field theory), a standard model of phase transitions and symmetry breaking \citep{ginzburg2009theory, rosenstein2021ginzburg}, with energy
\begin{equation}
\label{eq: Ginzburg-Landau density}
\mathcal{E}(x) =
\frac{1}{2}\sum_{i=1}^{d} (x_i - x_{i+1})^2
+\frac{1}{2}\sum_{i=1}^m (x_i^2 - \delta)^2
+\frac{1}{2}\sum_{i=m+1}^d x_i^2 ,
\end{equation}
with periodic boundary conditions $x_{d+1} = x_1$.

\begin{table}[h]
\centering
\small
\setlength{\tabcolsep}{4pt}
\caption{Results on the discretized Ginzburg-Landau model.}
\label{tab:gl_results}
\begin{tabular}{lcc}
\hline
Method & ESS $\uparrow$ & $\widebar{W}_2 \downarrow$ \\
\hline
ASBS   & 0.005        & 1.78  \\
TR-SBS & \textbf{0.445} & \textbf{0.15} \\
\hline
\end{tabular}
\end{table}
We use $d=16$, $m=5$, and $\delta=6$; all other settings are in \Cref{app:gl_detail}. 
The bistable coordinates induce multiple metastable configurations, while the nearest-neighbor term creates
strong spatial correlations across lattice sites. Since the energy satisfies $\mathcal{E}(x)=\mathcal{E}(-x)$, the target distribution obeys $p(x)=p(-x)$. We therefore evaluate global mode balance using $\widebar{W}_2(X_T,-X_T)$ together with the normalized effective sample size (ESS). Table~\ref{tab:gl_results} shows that TR-SBS substantially improves both weight stability and symmetry recovery over ASBS.

\begingroup
\subsection{Alanine dipeptide}
\label{app: aldp results}
\Cref{tab:aldp_full} reports all metrics of \Cref{app:aldp_detail} for TR-SBS and ASBS at similar stages of training, and \Cref{fig:aldp_populations} the populations of the rare conformations over training. Early in training, ASBS strongly overpopulates the region $\phi > 0$, which contains C7ax and $\alpha_L$. It then drains the region $\phi > 0$: at $48{,}500$ epochs, it attains a lower KL$_\phi$ than TR-SBS but already underpopulates C7ax by a factor of two. The training of ASBS diverged between $54{,}000$ and $54{,}500$ epochs. To evaluate ASBS closer to the divergence, we resumed it from its checkpoint at $48{,}500$ epochs with the same settings and newly filled buffers, saving a checkpoint every $500$ epochs. The resumed run drains C7ax further, to $0.034\%$ at $54{,}000$ epochs, about a fifth of the reference, and diverges in the same interval, between $54{,}000$ and $54{,}500$ epochs, in both runs at the beginning of a stage of adjoint matching. After the divergence, the samples are far from the target: at $58{,}500$ epochs of the original run, $19\%$ have the wrong chirality, and all KL divergences and $W_2$ are larger than at any earlier checkpoint. Between $48{,}500$ and $60{,}000$ epochs, $15\%$ of the optimization steps of the original run were skipped because the loss or its gradient was non-finite. In a third run, also resumed from $48{,}500$ epochs, we additionally clip the norm of the regression target of corrector matching per sample at $100$, the bound at which the outputs of the networks are clipped componentwise and the gradient of the energy in the terminal cost is clipped in norm, and discard samples with an energy above $1{,}000$ before they enter the buffers, which affects fewer than one in $10^5$ samples. This run does not diverge within $70{,}000$ epochs, but it drains C7ax further, to $0.0001\%$ at $70{,}000$ epochs, about $1/1{,}400$ of the reference (\Cref{tab:aldp_full,fig:aldp_populations}); between $58{,}000$ and $70{,}000$ epochs, its KL$_\phi$ decreases, but its KL$_\psi$ and $W_2$ increase. Its Ramachandran plot at $70{,}000$ epochs is shown in \Cref{fig:aldp}. TR-SBS overpopulates C7ax by a factor between $1.6$ and $2.1$ at all checkpoints evaluated with $10^7$ samples, while its overpopulation of $\alpha_L$ decreases from a factor of about seven to three, and KL$_\phi$ and KL$_\psi$ decrease by more than an order of magnitude over training. Apart from the diverged ASBS checkpoint and TR-SBS at $33{,}000$ epochs, the KL divergences of $\gamma_2$ and $\gamma_3$ are nearly identical for all checkpoints.

\begin{table}[h]
    \centering
    \footnotesize
    
    \caption{Alanine dipeptide: metrics ($10^7$ samples) of TR-SBS and ASBS with the same prior at similar stages of training and at the end of training. Only some checkpoints of each run were saved and evaluated with $10^7$ samples, so paired rows show the closest such checkpoints, whose epochs differ by up to $6{,}500$. ASBS diverges between $54{,}000$ and $54{,}500$ epochs; the row at $54{,}000$ epochs is from the run resumed at $48{,}500$ epochs, and the row at $58{,}500$ epochs is after the divergence. The rows ``ASBS, clipped'' are from the run resumed at $48{,}500$ epochs with clipped corrector targets (\Cref{app: aldp results}). Reference populations: C7ax $0.169\%$, $\alpha_L$ $0.049\%$.}
    \label{tab:aldp_full}
    \setlength{\tabcolsep}{3.5pt}
    \begin{tabular}{llcccccccc}
        \toprule
        Method & Epoch & KL$_\phi$ & KL$_\psi$ & KL$_{\gamma_1}$ & KL$_{\gamma_2}$ & KL$_{\gamma_3}$ & $W_2$ & C7ax (\%) & $\alpha_L$ (\%) \\
        \midrule
        ASBS   & 31k    & 0.2784 & 0.0243 & 0.0032 & 0.0158 & 0.0263 & 0.6659 & 1.5352 & 1.2509 \\
        TR-SBS & 33k    & 0.1862 & 0.0306 & 0.0193 & 0.0210 & 0.0300 & 0.3205 & 0.3342 & 0.3646 \\
        \midrule
        ASBS   & 48.5k  & 0.0256 & 0.0132 & 0.0036 & 0.0159 & 0.0262 & 0.4922 & 0.0788 & 0.0837 \\
        TR-SBS & 48.5k  & 0.0570 & 0.0093 & 0.0074 & 0.0162 & 0.0264 & 0.2700 & 0.3500 & 0.2259 \\
        \midrule
        ASBS, resumed   & 54k   & 0.0457 & 0.0276 & 0.0032 & 0.0157 & 0.0262 & 0.5744 & 0.0338 & 0.0746 \\
        ASBS (diverged) & 58.5k & 0.2967 & 0.3786 & 0.8315 & 0.0333 & 0.0423 & 1.7838 & 0.1927 & 0.3708 \\
        TR-SBS & 120k   & 0.0144 & 0.0037 & 0.0016 & 0.0159 & 0.0263 & 0.2460 & 0.2724 & 0.1360 \\
        \midrule
        ASBS, clipped   & 58k   & 0.0470 & 0.0205 & 0.0037 & 0.0160 & 0.0262 & 0.5793 & 0.0115 & 0.0156 \\
        TR-SBS & 60.5k & 0.0326 & 0.0069 & 0.0029 & 0.0159 & 0.0261 & 0.2956 & 0.3467 & 0.2001 \\
        \midrule
        ASBS, clipped   & 70k   & 0.0252 & 0.0406 & 0.0025 & 0.0161 & 0.0263 & 0.8319 & 0.0001 & 0.0001 \\
        TR-SBS & 76.5k & 0.0207 & 0.0055 & 0.0018 & 0.0160 & 0.0261 & 0.2595 & 0.3458 & 0.1463 \\
        \bottomrule
    \end{tabular}
\end{table}

\begin{figure}[h]
    \centering
    \includegraphics[width=\linewidth]{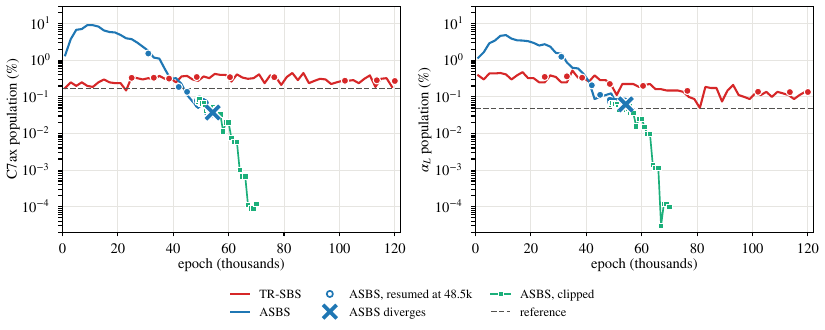}
    \caption{Alanine dipeptide: populations of C7ax (left) and $\alpha_L$ (right) over training for TR-SBS and ASBS with the same prior. Lines: evaluations of $2{,}000$ samples, averaged over $2{,}000$ epochs; dots: $10^7$ samples, open dots from ASBS resumed at $48{,}500$ epochs; squares: ASBS with clipped corrector targets, resumed at $48{,}500$ epochs, $10^7$ samples, up to $70{,}000$ epochs; dashed: reference; cross: divergence of ASBS without clipping, after which its populations are not shown (see \Cref{tab:aldp_full}).}
    \label{fig:aldp_populations}
\end{figure}

\endgroup

\begingroup
\subsection{Lennard-Jones cluster (LJ-13)}
\label{app: lj13}
We compare TR-SBS with ASBS on the $13$-particle Lennard-Jones cluster LJ-13 in three dimensions ($d=39$) with the harmonic confinement of \cite{liu2025adjoint}, using $10^4$ reference samples. Both methods use the same settings: EGNN controls \cite{satorras2021en} with $3$ layers and $64$ hidden units, a mean-free Gaussian prior with scale $2$, $\sigma_{\min} = 10^{-3}$, $\sigma_{\max} = 0.5$, $N_t = 200$, $N_{\mathrm{FC}} = 200$, $N_{\mathrm{BC}} = 50$, $L = 100$, $B = 200$, $N_{\mathrm{buf}} = 50{,}000$, $N_{\mathrm{new}} = 1{,}000$, and Adam with learning rate $10^{-4}$ on one NVIDIA A40 GPU. These are fewer integration steps and a smaller network than in \cite{liu2025adjoint}. TR-SBS uses $\varepsilon = 0.2$, $c_{\mathrm{IW}} = 5$, $N_{\mathrm{state}} = 500$, $N_{\mathrm{path}} = 20$, and the two modifications of \Cref{app:aldp_detail}. ASBS is trained for $12{,}500$ epochs and TR-SBS for $19{,}125$ epochs; for both, we evaluate the checkpoint with the lowest mean energy of the generated samples during training, which requires no reference samples. Following \cite{liu2025adjoint}, we report $W_2$ between $2{,}000$ model and $2{,}000$ reference samples, where the cost between two configurations is computed after optimal relabeling of the particles and rigid alignment, and $W_2$ between their energies, $\mathcal{E}(\cdot)W_2$, averaged over three trials.

\Cref{tab:lj13} shows that TR-SBS performs on par with ASBS: ASBS attains a slightly lower $W_2$ and TR-SBS a slightly lower $\mathcal{E}(\cdot)W_2$, and both attain a $W_2$ close to and an $\mathcal{E}(\cdot)W_2$ below the values reported in \cite{liu2025adjoint}. Two independent sets of reference samples have $W_2 = 1.32$, which is the value expected for exact samples of this size. Without the two modifications of \Cref{app:aldp_detail}, TR-SBS fails on LJ-13 ($W_2 = 5.24$, with a positive median energy).

\begin{table}[h]
    \centering
    \small
    
    \caption{LJ-13: mean $\pm$ standard deviation over three evaluation trials of $2{,}000$ samples.}
    \label{tab:lj13}
    \begin{tabular}{lcc}
        \toprule
        Method & $W_2 \downarrow$ & $\mathcal{E}(\cdot)W_2 \downarrow$ \\
        \midrule
        ASBS   & $1.571 \pm 0.002$ & $0.94 \pm 0.07$ \\
        TR-SBS & $1.585 \pm 0.001$ & $0.82 \pm 0.11$ \\
        \midrule
        ASBS, reported in \cite{liu2025adjoint} & $1.59 \pm 0.03$ & $1.99 \pm 1.01$ \\
        Reference vs.\ reference samples & $1.32 \pm 0.01$ & $0.39 \pm 0.04$ \\
        \bottomrule
    \end{tabular}
\end{table}
\endgroup

\end{document}